\documentclass[a4paper,fleqn]{cas-sc}

\usepackage{graphicx}%
\usepackage{multirow}%
\usepackage{amsmath,amssymb,amsfonts}%
\usepackage{amsthm}%
\usepackage{mathrsfs}%
\usepackage[title]{appendix}%
\usepackage{xcolor}%
\usepackage{textcomp}%
\usepackage{manyfoot}%
\usepackage{booktabs}%
\usepackage{algorithm}%
\usepackage{algorithmicx}%
\usepackage{algpseudocode}%
\usepackage{listings}%
\usepackage{xfrac}
\usepackage{nicefrac}
\usepackage{stmaryrd}
\usepackage{footmisc}
\usepackage{subcaption}

\newcommand{\argmin}{\operatornamewithlimits{argmin}}
\newcommand{\argmax}{\operatornamewithlimits{argmax}}

\usepackage[authoryear]{natbib}

\def\tsc#1{\csdef{#1}{\textsc{\lowercase{#1}}\xspace}}
\tsc{WGM}
\tsc{QE}

\begin{document}
	\let\WriteBookmarks\relax
	\def\floatpagepagefraction{1}
	\def\textpagefraction{.001}
	
	\shorttitle{Image Matching Filtering and Refinement
		by Planes and Beyond }    
	
	\shortauthors{F. Bellavia et al.}  
	
	\title [mode = title]{Image Matching Filtering and Refinement
		by Planes and Beyond }  
			
	\author[1]{Fabio Bellavia}[orcid=0000-0002-1688-8476]
	
	\cormark[1]
		
	\ead{fabio.bellavia@unipa.it}
		
	\credit{Conceptualization, Methodology, Software, Validation, Investigation, Resources, Data Curation, Writing - Original Draft, Writing - Review \& Editing, Visualization, Supervision}
	
	\affiliation[1]{organization={University of Palermo},
		country={Italy}}
	
	\author[2]{Zhenjun Zhao}[orcid=0000-0002-0150-4601]
		
	\ead{ericzzj89@gmail.com}
		
	\credit{Software, Investigation, Resources, Writing - Review \& Editing, Visualization}
	
	\affiliation[2]{organization={University of Zaragoza},
		country={Spain}}

	\author[3]{Luca Morelli}[orcid=0000-0001-7180-2279]

	\ead{lmorelli@fbk.eu}
	
	\credit{Investigation, Resources}

	\author[3]{Fabio Remondino}[orcid=0000-0001-6097-5342]

	\ead{remondino@fbk.eu}
	
	\credit{Investigation, Resources}
	
	\affiliation[3]{organization={Bruno Kessler Foundation},
		country={Italy}}

	\begin{abstract}
	Sparse image matching remains a valid option to more recent dense end-to-end deep approaches due to a better computational efficiency and scalability. The sparse approaches rely on an accurate and broad selection of correspondences for which the choice of appropriate filtering and refinement steps is crucial. This paper provides a consistent and extensive evaluation of state-of-the-art filtering and refinement methods on common image matching pipelines. Unlike previous comparisons, the designed benchmark also takes into account the more general, real, and practical cases where camera intrinsics are unavailable. Moreover, a novel and effective strategy combining non-deep traditional computer vision approaches based on planar constraints and cross correlation is presented.
	
	Experimental analysis provides several insights for current application design and future research directions. In particular, the choice of a proper evaluation protocol discloses the effective differences within the compared solutions which otherwise would tend to flatten. Moreover, the proposed classical algorithmic approach is competitive with recent deep methods. Besides providing robust baseline using traditional computer vision for the evaluation of deep-based methods, this knowledge is useful to improve and better understand the deep image matching architectures. On one hand, geometry-based filtering is effective in presence of outliers without degrading already robust deep pipelines; on the other hand cross-correlation refinement is valid in the case of corner-like keypoints and allows to not directly discard inaccurate matches by default in deep pipelines but to retain and refine them for achieving a better coverage of the scene.  
	
	All the resources used within this paper, including code and data, further experimental evaluations, high-resolution versions of the images, plots, and tables presented in this paper are freely available at \url{https://github.com/fb82/MiHo}.
	\end{abstract}
	
	
	
	
	
	\maketitle
	
\section{Introduction}\label{sec1}
\subsection{Background}
Image matching is the backbone of most higher-level computer vision applications that require image registration to recover the 3D scene structure, including the camera poses. Image stitching~(\cite{stitching}), Structure-from-Motion (SfM, \cite{colmap}), Simultaneous Localization and Mapping (SLAM, \cite{orbslam}) and the more recently Neural Radiance Fields (NeRF, \cite{nerf}) and Gaussian Splatting (GS, \cite{gaussian_splatting}) are nowadays common and major tasks critically relying on image matching.

Image matching traditional paradigm can be organized into a modular pipeline concerning keypoint extraction, feature description, and the proper correspondence matching~(\cite{imw2020}). This representation is being neglected in modern deep end-to-end image matching methods due to their intrinsic design which guarantees a better global optimization at the expense of the interpretability of the system. Notwithstanding that dense end-to-end deep networks represent for many aspects the State-Of-The-Art (SOTA) in image matching, such in terms of the robustness and accuracy of the estimated poses in complex scenes~(\cite{roma2, vggt}), sparse and even handcrafted image matching pipelines such as the popular Scale Invariant Feature Transform (SIFT,~\cite{sift}) are employed still today in practical applications~(\cite{colmap}) due to their scalability, adaptability, and understandability. Furthermore, handcrafted or deep modules remain present as match post-processing, in order to further filter the final output matches as done notably by the RAndom SAmpling Consensus (RANSAC,~\cite{ransac}), based on geometric constraints.

Image matching filters to prune correspondences are not limited to RANSAC, which in any of its form~(\cite{superransac}) still remains mandatory nowadays as final step, and have been evolved from handcrafted~(\cite{gms}) methods to deep ones~(\cite{learning_correspondences}). In addition, more recently methods inspired by coarse-to-fine paradigm, such as the Local Feature TRansformer (LoFTR,~\cite{loftr}), have been developed to refine correspondences~(\cite{fcgnn}), which can be considered as the introduction of a feedback-system in the pipelined interpretation of the matching process. 

\subsection{Contributions}
\begin{itemize}
\item As contribution, this paper presents an extensive comparative evaluation against more than ten recent SOTA deep and handcrafted image matching filtering and refinement methods. The evaluation includes both non-planar indoor and outdoor standard benchmark datasets~(\cite{megadepth,scannet,imw2020}), as well as planar scenes~(\cite{hpatches,mods}). Each filtering approach has been tested to post-process matches obtained with nine different image matching pipelines and end-to-end networks, also considering distinct RANSAC configurations.

\item Unlike the mainstream evaluation which has been taken as current standard~(\cite{loftr}), the proposed benchmark also considers the case that no camera intrinsics are available, reflecting a more general, realistic, and practical scenario. On this premise, the specific complexity of the pose estimation working with fundamental and essential matrices~(\cite{multiview}) is highlighted. In particular, the evaluation protocol relying on the fundamental matrix discloses behavior differences otherwise disguised by using instead error metrics based on the essential matrix. This analysis reveals the many-sided nature of the image matching problem and of its solutions.

\item As transversal contribution, this paper also introduces a modular approach to filter and refine matches as post-process before RANSAC, which was sketched in~\cite{miho_base}. The key idea is that the motion flow within the images, i.e. the correspondences, can be approximated by local overlapping homography transformations~(\cite{multiview}), a concept already exploited in a different form by~\cite{slime}. This is the core concept of traditional image matching pipelines, where local patches are approximated by less constrained transformations going from a fine to a coarse scale level. Specifically, besides the similarity and affine transformations, the planar homography is introduced at a further level. The extrapolated local planar approximation clues can be used to filter out inconsistent matches, but also achieve a more reliable local patch normalization between matched regions so as to refine keypoints by traditional template matching through Normalized Cross-Correlation (NCC,~\cite{gonzales_wood}). Different strategies in this sense are proposed and analyzed. 

\item According to the evaluation, this modular approach is competitive with deep SOTA equivalent architectures. RANSAC takes noticeable benefits from the above approach as pre-filtering in most configurations and scenes when the outlier contamination of the base pipeline is relevant, and in any case does not deteriorate the matches. Concerning the NCC-based keypoint refinement step, localization accuracy improvements seem to strongly depend on the kind of extracted keypoint. In particular, for corner-like keypoints, which are predominant in detectors such as the Keypoint Network (Key.Net,~\cite{keynet}) or SuperPoint~(\cite{superpoint}), the proposed NCC-based solution greatly improves the keypoint position. For blob-like keypoints, predominant for instance in SIFT, or flat patches, the NCC refinement can degrade the keypoint localization, due probably to the different nature of their surrounding areas.

\item The proposed solution for the filtering step is purely geometric, requiring only keypoint coordinates, while the NCC refinement relies on the image intensity in the local area of the patches. The whole approach is handcrafted and does not require re-training to adapt to the particular kind of scene. Furthermore, its behavior is interpretable as not hidden by deep architectures. As side yet central effect, the above filter and refinement strategies can be employed as a robust baseline using traditional computer vision for the evaluation of deep-based methods and to improve and better understand the deep image matching architectures.
\end{itemize}

\subsection{Paper organization}
The rest of the paper is organized as follows. The related work is presented in Sec.~\ref{related_work}, the design of the proposed match filtering and refinement strategy is discussed in Sec.~\ref{mop_miho_ncc}, and the evaluation setup and the results are analyzed in Sec.\ref{eval}. Finally, conclusions and future work are outlined in Sec.~\ref{conclusion}. 


\section{Related work}\label{related_work}

\subsection{Non-deep image matching}
Traditionally, sparse image matching has three main steps: keypoint detection, description, and proper matching.

Keypoint detection is aimed at finding repeatable yet discriminant image regions, usually recognized as corner-like and blob-like ones, even if there is no clear separation within them outside ad-hoc ideal images. Nowadays corners are predominantly extracted by the Harris detector~(\cite{harris}), while blobs by SIFT~(\cite{sift}). More recently, the joint extraction and refinement of robust keypoints has been investigated by the Gaussian Mixture Models for Interpretable Keypoint Refinement and Scoring (GMM-IKRS,~\cite{gmmikrs}). The idea is to extract image keypoints after several homography warps and cluster them after back-projection into the input images so as to select the best clusters in terms of robustness and deviation, as another perspective of the Affine SIFT (ASIFT,~\cite{asift}) strategy. 

Feature descriptors aim at embedding the keypoint characterization into numerical vectors. Among non-deep keypoint descriptors, nowadays only the SIFT descriptor is practically still used and widespread due to its balance within computational efficiency, scaling, and matching robustness. Many high-level computer vision applications, like COLMAP~(\cite{colmap}) in the case of SfM, are based on SIFT. 

Next, patch normalization takes place in between the detection and description of the keypoints, usually implicitly, and roughly consists of aligning patches before the descriptor computation~(\cite{sift}). The key idea is that complex  deformations can be locally approximated by more simple transformations as long as the considered neighborhood is small. SIFT normalizes the patch by a similarity transformation, where the scale is inferred by the detector and the dominant rotation is computed from the gradient orientation histogram. Affine covariant patch normalization~(\cite{hess_lapl_affine}) was indeed the natural evolution, which can improve the normalization in case of hard perspective distortions. For small scene rotation, up-right patches can lead to better matches since a strong geometric constraint limiting the solution space is imposed~(\cite{sgloh2}). 

The last step is the proper matching, which mainly pairs keypoints by considering the distance similarity on the associated descriptors. Nearest Neighbor (NN) and Nearest Neighbor Ratio (NNR) are the common and simple approaches even if additional extensions have been developed~(\cite{fginn,dtm}). Similarity alone does not generally provide satisfactory results, so matches are further filtered by geometric constraints.

\subsection{Non-deep image matching filtering}
RANSAC provides an effective method to discard outliers under the more general epipolar geometry constraints of the scene or in the case of planar homographies. RANSAC continues to be extended in several ways~(\cite{superransac}). Worth mentioning are the Degenerate SAmpling Consensus (DegenSAC,~\cite{degensac}) to avoid degenerate configurations, the MArGinalized SAmpling Consensus (MAGSAC,~\cite{magsac}) for defining a robust inlier threshold, which remains the most critical parameter. 

A more general and relaxed geometric assumption considers only local neighborhood consistency across the images, as in Grid-based Motion Statistics (GMS,~\cite{gms}) using square neighborhood. Circular~(\cite{lpm}) and affine-based~(\cite{pgm,pfm}) neighborhoods, or Delaunay triangulation~(\cite{dtm,pfm}), have been employed too. In most cases, these approaches require an initial set of robust seed matches for initialization, whose selection may become critical~(\cite{glpm}), while other approaches explicitly exploit descriptor similarities~(\cite{dtm}). In this respect, the vanilla RANSAC and GMS are pure geometric image filters. Another class of methods estimates the motion field for checking consistency~(\cite{vfc,bm,code}) or frames local neighborhood consistency into RANSAC as in the Adaptive Locally-Affine Matching (AdaLAM,~\cite{adalam}). AdaLAM design can in turn be related with GroupSAC~(\cite{groupsac}), which draws hypothesis model sample correspondences from different clusters, and with the more general multi-model fitting problem for which several solutions have been investigated~(\cite{jlinkage}). Among these approaches Consensus Clustering (CC,~\cite{cc}) and Progressive-X~(\cite{progx}) have been developed for the specific case of multiple planar homographies.   

\subsection{Deep image matching}
Deep learning evolution has provided a boost to image matching as well as other computer vision research fields thanks to the evolution of the deep architectures, hardware and software computational capability and design, and the huge amount of datasets now available for the training. Hybrid pipelines raised first, replacing handcrafted descriptors with deep ones. SOTA in this sense is represented by the Hard Network (HardNet,~\cite{hardnet}) and similar networks~(\cite{sosnet,hynet}). For what concerns the keypoint extraction, better results with respect to the handcrafted counterparts were achieved by mimic handcrafted detector design as for Key.Net. Deep learning has been successfully employed to learn patch orientation~(\cite{learning_orientation}) or more complete affine transformations~(\cite{affnet}). The mentioned deep modules have been employed successfully in many hybrid matching pipelines, especially in conjunction with SIFT~(\cite{imw2020}) or Harris corners~(\cite{hz_pipeline}).

Deep image matching pipelines began exceeding handcrafted as with the introduction of the joint optimization of both the detector \& descriptor. SuperPoint integrates both in a single network and does not require a separate yet consecutive training of these as in~\cite{lift}. A further key feature introduced by SuperPoint was the use of homographic adaptation, which consists of the use of planar homographies to generate corresponding images during the training, to allow self-supervised training. Further solutions were proposed by similar architectures~(\cite{r2d2}). Moreover, the Accurate and Lightweight Keypoint Detection and Descriptor Extraction (ALIKE,~\cite{alike}) implements a differentiable Non-Maximum Suppression (NMS) for the keypoint selection on the score map, and the DIScrete Keypoints (DISK,~\cite{disk}) employs reinforcement learning. ALIKE design is further improved by A LIghter Keypoint and descriptor Extraction network with Deformable transformation (ALIKED,~\cite{aliked}), which exploits deformable local features for each sparse keypoint instead of dense descriptor maps, saving computational budget. Alternative rearrangements in the pipeline structure to use the same sub-network for both detector \& descriptor (detect-and-describe, ~\cite{d2net}), or to extract descriptors and then keypoints (describe-then-detect,~\cite{d2d}) instead of the common approach (detect-then-describe), have been also investigated. More recently, the idea of decoupling the two blocks by formulating the matching problems in terms of 3D tracks in large-scale SfM has been applied in Detect, Don’t Describe--Describe, Don’t Detect (DeDoDe,~\cite{dedode}) in turn employed on the recent Local Matching (LoMa,\cite{loma}).

SuperGlue can represent a breakthrough in the areas providing the first end-to-end deep matching pipeline and has been extended by the recent LightGlue~(\cite{lightglue}) improving efficiency, accuracy and the training process. Detector-free end-to-end image matching architectures have been also proposed such LoFTR, which avoid explicitly computing a sparse keypoint map, considering instead a semi-dense strategy. Coarse-to-fine strategy following LoFTR design has been also applied to hybrid canonical pipelines, leading to matching results comparable in challenging scenarios at an increased computation cost~(\cite{slime}). More recently, dense image matching networks that directly estimate the scene as dense point maps, as Matching And Stereo
3D Reconstruction (MASt3R,~\cite{mast3r}) and others~(\cite{vggt,mapany}), or employing Gaussian process as the Robust Matching (RoMa,~\cite{roma}), achieve SOTA results in many benchmarks but are still more computationally expensive than sparse or semi-dense methods and less scalable in many common context.

\subsection{Deep image matching filtering and refinement}
The first effective deep matching filter module was achieved with the introduction of the context normalization~(\cite{learning_correspondences}) that guarantees to preserve permutation equivalence. The Attentive Context Network (ACNe,~\cite{acne}) further includes local and global attention and the Order-Aware Network (OANet, ~\cite{oanet}) adds additional layers to learn how to cluster unordered sets of correspondences. The Consensus Learning Network (CLNet,~\cite{clnet}) prunes matches by filtering data according a local-to-global dynamic neighborhood consensus graphs and the Neighbor Consistency Mining Network (NCMNet,~\cite{ncmnet}) improves the CLNet by considering different neighborhoods in both the feature space and the keypoint coordinate space. More recently, the Multiple Sparse Semantics Dynamic Graph Network (MS$^2$DG-Net,~\cite{ms2dgnet}) designs the filtering in terms of a neighborhood graph through transformers~(\cite{transformer}). Unlike previous deep approaches, ConvMatch~(\cite{convmatch}) builds a smooth motion field by making use of convolutional layers to verify the consistency of the matches. DeMatch~(\cite{dematch}) refines ConvMatch motion field estimation for a better accommodation of scene disparities by decomposing the rough motion field into several sub-fields. Deep differentiable RANSAC modules have been investigated as well~(\cite{ngransac,generalizedransac}).

The coarse-to-fine strategy trend has pushed the research of solutions focusing on the refinement of keypoint localization as a standalone module after the coarse assignment of the matches. Worth mentioning is Pixel-Perfect SfM~(\cite{pixelperfectsfm}) that learns to refine multi-view tracks on their photometric appearance and the SfM structure, while Patch2Pix~(\cite{patch2pix}) refines and filters patches exploiting image dense features extracted by the network backbone. Similar to Patch2Pix, Keypt2Subpx~(\cite{keypt2subpx}) is a recent lightweight module to refine matches which requires as input, beside keypoint positions, the descriptors of the local heat map of the correspondences.

Another approach related to ASIFT was proposed by~\cite{monoref}, where the planes on the scene are extracted by a single image deep-estimation network independently for each image, so as to rectify the corresponding image regions accordingly before extracting keypoints and patches. This approach requires real image planes and to extract the 3D structure of the scene, which makes it unable to resolve ambiguities in actual planar scenes leading to failure cases not present in both standard pipelines and our approach.

Finally, the Filtering and Calibrating Graph Neural Network (FC-GNN,~\cite{fcgnn}) is an attention-based Graph Neural Network (GNN) jointly leveraging contextual and local information to both filter and refine matches. Unlike Patch2Pix and Keypt2Subpx, both FC-GNN and the proposed strategy do not require feature maps or descriptor confidences as input. 

\section{Proposed modular non-deep image matching filtering and refinement}\label{mop_miho_ncc}
\subsection{Multiple Overlapping Planes (MOP)}\label{mop}
MOP is the main core module of the proposed filtering strategies. Assuming that motion flows within the images can be approximated piecewise by local planar homography, the base idea of MOP to collect several relaxed planar hypotheses through RANSAC estimation.

Matches are denoted as relaxed or strict inliers according to two different error thresholds on the maximum reprojection homography error $t_l=15$ px and $t_h=t_l/2$, respectively. For a better estimation, the implemented RANSAC includes requires a minimum number of tested hypothesis and introduces checks for the compactness of minimum model sample matches, the degenerate cases and the cheilarity. Note that MOP relies on loose thresholds in order of being able to work with approximated virtual and real planes of the scene, unlike of other methods designed for extracting strict planes relying on tight reprojection error thresholds such as CC and Progressive-X. 

Starting with the initial set of candidate matches, the $k$ MOP iteration retrieves the best planar hypothesis $\mathrm{H}^k$ by RANSAC according to $t_l$ and accepts this if the number of inliers provided is less than the required amount $n=12$. Otherwise, a failure counter $c_f$ is incremented. In case of success all inliers according to an adaptive threshold $t^k$ are removed and a further $\mathrm{H}^{k+1}$ is sought. The adaptive threshold $t^k$ is equal to a strict threshold $t_h$ in case the cardinality of the inlier set according to $t_h$ is greater than half of $n$, i.e. in the worst case no more than half of the added matches should belong to previous overlapping planes. If this condition does not hold $t^k$ turns into the relaxed threshold $t_l$ and a failure counter $c_f$ is incremented. 

A loose threshold $t_l$ is able to select a real motion flow which is only locally approximated by planes, while removing only strong inliers by $t_h$ for the next iteration $k+1$ guarantees smooth changes within overlapping planes and more robustness to noise. Nevertheless, in case of slow convergence or when the homography search gets stuck around a wide overlapping or noisy planar configuration, limiting the search to matches excluded by previous planes can provide a way out. The final set of homography returned is
$\mathcal{H}^\star=\{\mathrm{H}^1,\mathrm{H}^2,\cdots,\mathrm{H}^{k^\star}\}$ where $k^\star$ is last iteration obtained when the failure counter reaches $c_f^\star=3$. 

In order to speed up the RANSAC search, a buffer is globally maintained in order to contain the top $z=5$ discarded sub-optimal homographies encountered within a RANSAC run. At the beginning of a RANSAC run, homographies in the buffer are evaluated before proceeding with true random samples to provide a bootstrap for the solution search. This also guarantees a global sequential inspection of the overall solution search space within MOP.

\begin{figure}
	\centering
	\includegraphics[width=0.47\textwidth]{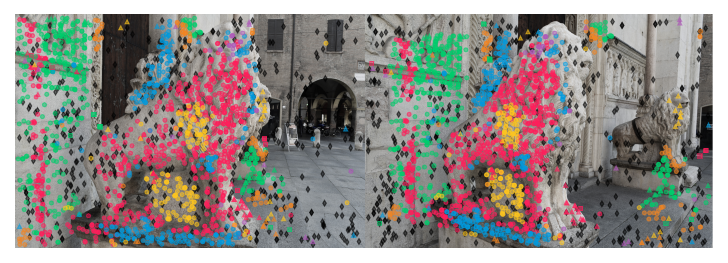} \\
	\includegraphics[width=0.47\textwidth]{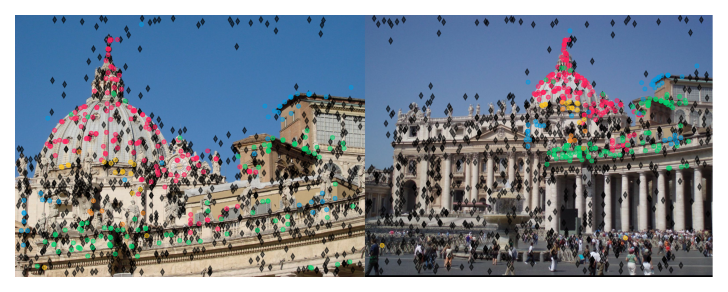} \\
	\includegraphics[width=0.47\textwidth]{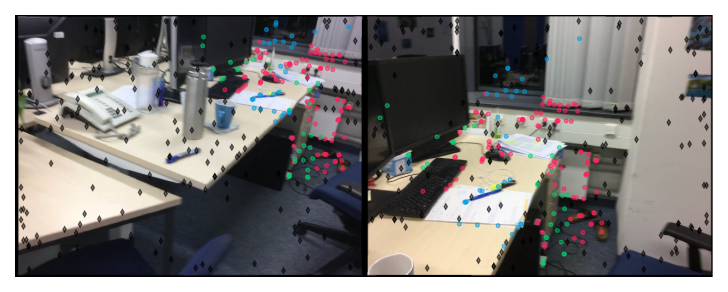}
	\caption{MOP+MiHo clustering and filtered matches for some image pair examples. Each combination of makers and colors is associated to a unique virtual planar homography as described in Secs.~\ref{mop}-\ref{miho}, while discarded matches are indicated by black diamonds. The matching pipeline employed is based on Key.Net; note that in terms of cluster representation MOP and MOP+MiHo do not present visually appreciable differences. Best viewed in color and zoomed in.\label{miho_img}}
\end{figure}

At the end of the process, matches not compatible with any homography in $\mathcal{H}^\star$, i.e. that are not relaxed inliers, are discarded as outliers. For the remaining matches, among the top compatible homographies in terms on the inlier set cardinality, the one with the minimum reprojection error is assigned. Figure~\ref{miho_img} shows an example of the achieved solution. Keypoint belonging to discarded matches are marked with black diamonds, while the clusters highlighted by other combinations of markers and colors indicate the resulting filtered matches belonging to the same planar homography. MOP implementation details can be found in Appendix~\ref{mop_appendix}.

\subsection{Middle and half homographies}\label{miho}
The homography $\mathrm{H}_m\in\mathcal{H}^\star$ for the match $m=(\mathbf{x}_1,\mathbf{x}_2)$ allows to reproject the local patch neighborhood of $\mathbf{x}_1\in I_1$ onto the corresponding one centered in $\mathbf{x}_2\in I_2$. Nevertheless, $\mathrm{H}_m$ approximates the true transformation within the images which becomes less accurate as long as the distance from the keypoint center increases. This implies that patch alignment could be invalid for wider, and theoretically more discriminant, patches. Solutions aimed at improving the patch normalization for a better keypoint refinement are investigated.

The Middle Homography (MiHo) is aimed to alleviate the above issue by providing symmetric patch rectification for matches and error threshold consistency. The idea behind MiHo is to break the homography $\mathrm{H}_m$ in the middle by two homographies $\mathrm{H}_{m_1}$ and $\mathrm{H}_{m_2}$ such that $\mathrm{H}_m=\mathrm{H}_{m_2}\mathrm{H}_{m_1}$. The assumption is that the patch on $\mathbf{x}_1$ gets deformed by $\mathrm{H}_{m_1}$ less than by $\mathrm{H}_m$, and likewise the patch on $\mathbf{x}_2$ gets deformed by $\mathrm{H}_{m_2}^{-1}$ less than by $\mathrm{H}_m^{-1}$. This means that visually on the reprojected patch a unit area square should remain almost similar to the original one. As this must hold from both the images, the deformation error must be distributed almost equally between the two homographies $\mathrm{H}_{m_1}$ and $\mathrm{H}_{m_2}$. Moreover, since interpolation degrades with up-sampling, MiHo aims to balance the down-sampling of the patch at ﬁner resolution and the up-sampling of the patch at the coarser resolution.

MiHo computation exploits the heuristic of~\cite{average_homography} to modify the base RANSAC in MOP in order to account for the required constraints. Each match $m$ is replaced by two corresponding matches $m_1=(\mathbf{x}_1,\mathbf{m})$ and $m_2=(\mathbf{m},\mathbf{x}_2)$ where $\mathbf{m}$ is the midpoint within the two keypoints. Two concurrent RANSAC are run at each MOP iteration, considering corresponding matches $m_1$ and $m_2$ and leading to the homography pair $(\mathrm{H}_{m_1}, \mathrm{H}_{m_2})$ whose compatibility have to be verified simultaneously for the original match being inlier. Full details are reported in Appendix~\ref{miho_appendix}.

MiHo shares some similarities with the half homography concept described in VSAC~(\cite{VSAC}) with the purpose of adjust the matched keypoints after the final homography has been estimated. Specifically, the decomposition $\mathrm{H}_m=\mathrm{H}_v\mathrm{H}_v$ of the original homography matrix $\mathrm{H}_m$ into the half homography matrix $\mathrm{H}_v$ provided by the square root of $\mathrm{H}_m$ is employed as replacement of both $\mathrm{H}_{m_1}$ and $\mathrm{H}_{m_2}$. The VSAC half homography can be used to directly split the homographies computed by the base MOP into pairs of homographies instead of MiHo.

\begin{figure}
	\centering
	\subcaptionbox{\label{ma}}{\includegraphics[width=0.33\textwidth]{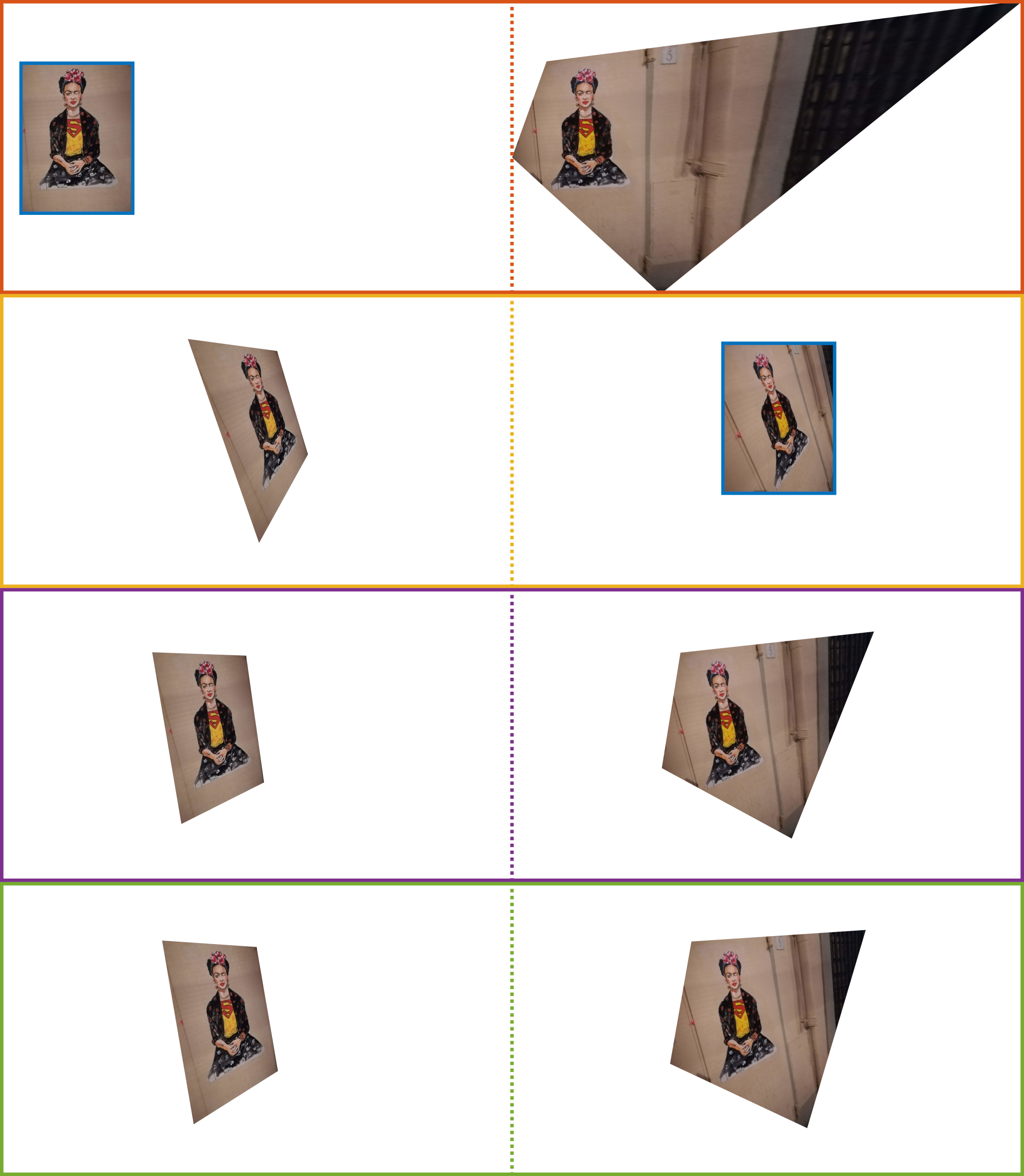}}
	\subcaptionbox{\label{mb}}{\includegraphics[width=0.33\textwidth]{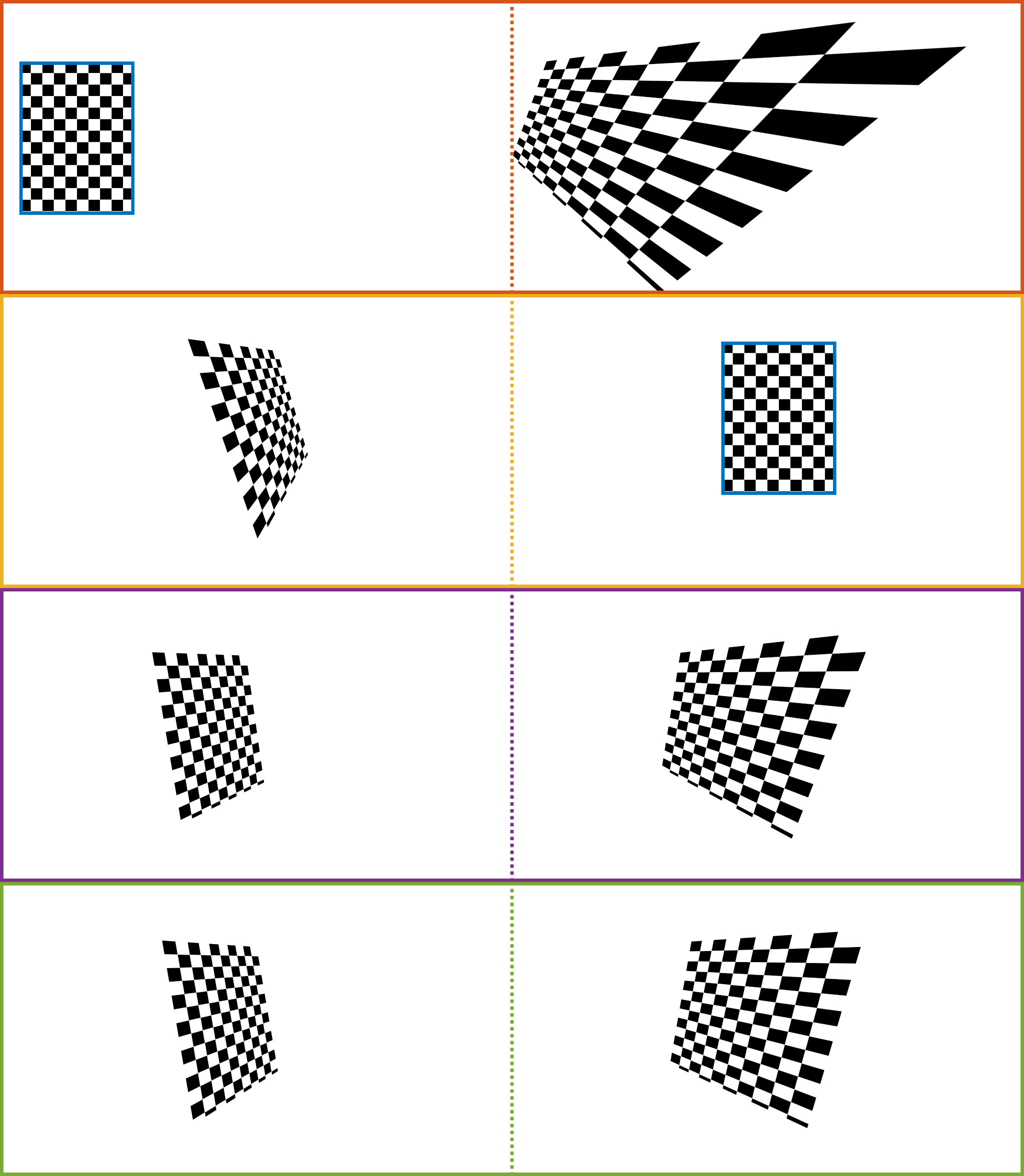}}
	\caption{Visual comparisons between the standard planar homography, MiHo paired homographies and VSAC half homography, both detailed in Sec.~\ref{miho}, when the images share almost the same common area. (\subref{ma}) Two corresponding images, framed in blue, are used in turn as reference in the two top rows while the other one is reprojected by a planar homography, show in the same row. In the case of MiHo, both images are warped as shown in the third row, yet the overall distortion is reduced. VSAC half homography, shown in the bottom row, works similar to MiHo. The local transformation is better highlighted in (\subref{mb}) with the related warping of the unit-square representation of the original images. Best viewed in color and zoomed in.\vspace{-1em}}\label{miho_figa}
\end{figure}

\begin{figure}
	\centering
	\subcaptionbox{\label{mc}}{\includegraphics[width=0.33\textwidth]{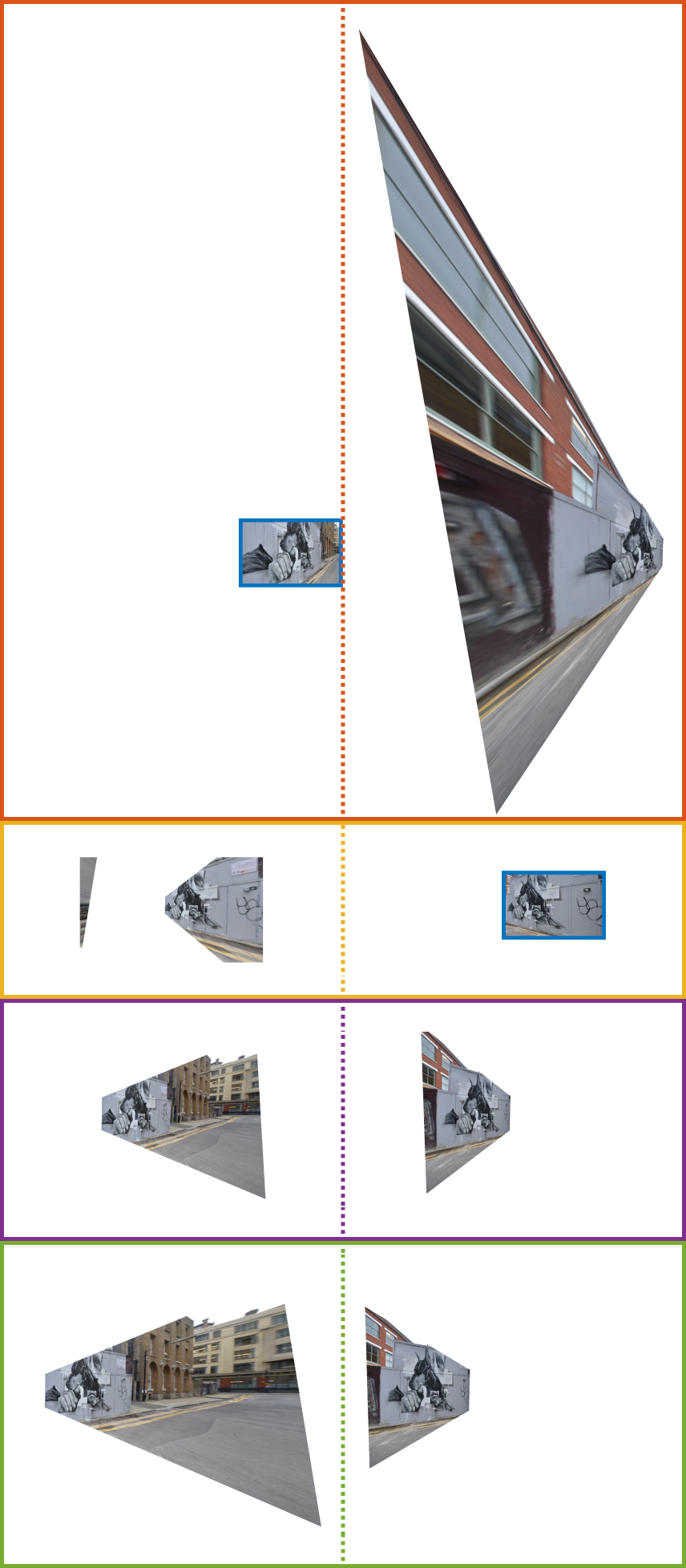}}
	\subcaptionbox{\label{md}}{\includegraphics[width=0.33\textwidth]{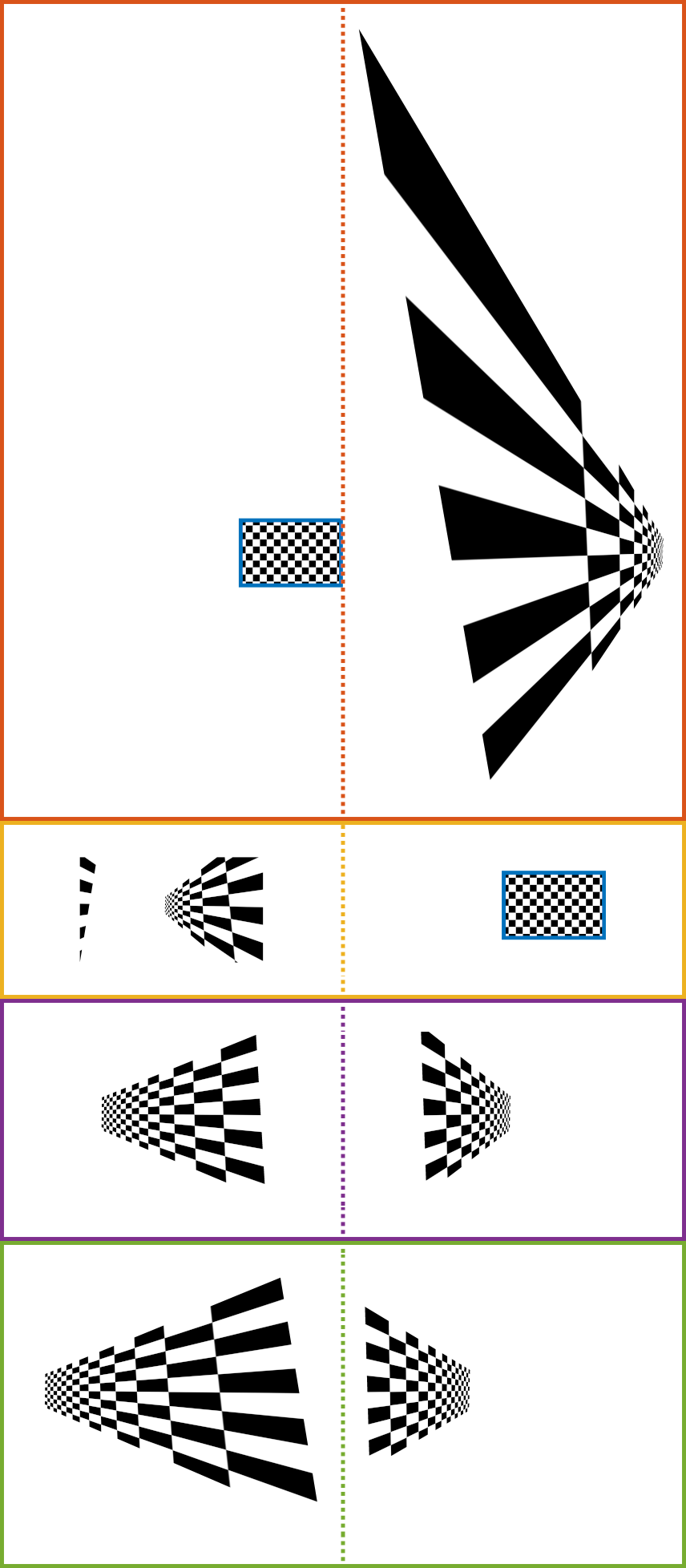}}
	\caption{Visual comparisons between the standard planar homography, MiHo paired homographies and VSAC half homography, both detailed in Sec.~\ref{miho}, when the images share only a partial common area. The same notation employed in Fig.~\ref{miho_figa} is adopted. As shown in the second row, the homography distortion can break the plane convex hull. MiHo, depicted in the third row, deforms less in the unshared image areas than VSAC half homography, shown in the bottom row. Best viewed in color and zoomed in.\vspace{-1em}}\label{miho_figb}
\end{figure}

Figure~\ref{miho_figa}-\ref{miho_figb} show the differences in applying MiHo to align to planar scenes with respect to directly reproject in any of the two images considered as reference, or using VSAC half homography. More examples can be found in the supplemental material. With respect to the direct reprojections, it can be noted that MiHo distortions are overall reduced, as highlighted by the corresponding grid deformation. With respect to VSAC half homography, on one hand the differences are almost undetectable when the images mostly overlap, as shown in Fig.\ref{miho_figa}. On the other hand, MiHo reduces the distortions on the non-overlapping regions as the common scene area between the images is reduced, as shown in \ref{miho_figb}. By inspection, the differences between MiHo and VSAC half homograhy can mainly abducted to different camera center assumptions.

As discussed by~\cite{multiview}, VSAC half homography corresponds to use conjugate rotations to generate synthetic views through fractional angles, implicitly assuming the same and general intrinsic camera matrix for both images. MiHo does not impose this constraint since in general $\mathrm{H}_{m_1}\neq\mathrm{H}_{m_2}$ , but indirectly includes a regularization factor towards affine transformations. Midpoints are preserved by similarity and affine transformation, being linear combinations, but not by general planar homographies. This design is also reflected into different planes and inliers extracted by MOP and MOP+MiHo. To balance this additional restrain with respect to the original MOP, in MOP+MiHo the minimum requirement amount of inliers for increasing the failure counter has been relaxed after experimental validation from $n=12$ to $n=8$.

\begin{figure}
	\hfill
	\subcaptionbox{$0^\circ$\label{ra}}{\includegraphics{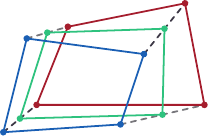}}
	\hfill
	\subcaptionbox{$90^\circ$\label{rb}}{\includegraphics{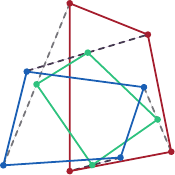}}
	\hfill
	\subcaptionbox{$180^\circ$\label{rc}}{\includegraphics{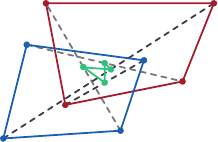}}
	\hfill
	\subcaptionbox{$270^\circ$\label{rd}}{\includegraphics{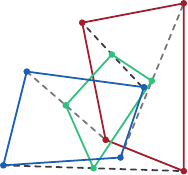}}
	\hfill
	\caption{Midpoints and rotations with MiHo. The blue and red quadrilaterals are linked by a homography which is defined by the four corner correspondences, indicated by dashed gray lines. The midpoints of the corner correspondences identify the reference green quadrilateral and the derived two MiHo planar homographies with the original quadrilaterals. In the optimal MiHo configuration (\subref{ra}) the area of the quadrilateral defined by the midpoints is maximized. Incremental relative rotations within the two original quadrilaterals by $90^\circ$ decrease the above area through (\subref{rb}) to the minimum (\subref{rc}). From the minimum further rotations increase the area through (\subref{rd}) to the maximum again. As heuristic, in the best configuration the distance of any two midpoint corners should be within the distances of the corresponding original corners of both images. Note that for the specific example, in the worst case there is also a violation of the planar orientation constraints. Best viewed in color and zoomed in.}\label{miho_rot_fig}
\end{figure}

Notice that both MiHo and VSAC half homography are translation invariant but not rotation invariant. In particular, rotations about $180^\circ$ as well as reflections could lead to degenerate solutions. As illustrated in Fig.~\ref{miho_rot_fig}, a simple yet effective heuristic can avoid MiHo worst case, which is also valid for VSAC half homography. Further implementation details can be found in Appendix~\ref{miho_rot_fix}. Finally, it is also possible to directly excluding reflections in the RANSAC loop inside MOP. This further variation is denoted by MOP$^\star$ and detailed in Appendix~\ref{mop_star}. 

\subsection{Normalized Cross Correlation (NCC)}\label{ncc}
NCC is employed to effectively refine corresponding keypoint localization. The approach assumes that corresponding keypoint patches have been roughly aligned locally by the transformations provided as a homography pair, for instance, by MOP+MiHo. Indeed, each homography pair provides a consistent canonical frame for the patches associated to a match. In the warped alignment space, template matching~(\cite{gonzales_wood}) is employed to refine translation offsets in order to maximize the NCC peak response when the patch of a keypoint is employed as a filter onto the other image of the corresponding keypoint.

\vspace{2em}
For the match $m=(\mathbf{x}_1,\mathbf{x}_2)$ within images $I_1$ and $I_2$ there can be multiple potential choices of the warping homography pair, which are denoted by the set of associated homography pairs $\mathcal{H}_m=\{(\mathrm{H}_1,\mathrm{H}_2)\}$. In detail, according to the previous steps of the matching pipeline, a candidate is the base homography pair $(\mathrm{H}^b_1,\mathrm{H}^b_2)$ defined as
\begin{enumerate}[a)]
    \item $(\mathrm{I},\mathrm{I})$, where $\mathrm{I}\in\mathbb{R}^{3\times3}$ is the identity matrix, when no local neighborhood data are provided, e.g. for SuperPoint; 
    \item $(\mathrm{A}_1,\mathrm{A}_2)$, where $\mathrm{A}_1,\mathrm{A}_2\in\mathbb{R}^{3\times3}$ are affine or similarity transformations in case of affine covariant detector \& descriptor, as for AffNet or SIFT, respectively;
\end{enumerate}
and the other is the extended pair $(\mathrm{H}^e_1,\mathrm{H}^e_2)$ defined as
\begin{enumerate}[a)]\addtocounter{enumi}{2}
    \item $(\mathrm{I},\mathrm{H}_m)$, where $\mathrm{H}_m$ is defined in Eq.~\ref{estimated_hom}, in case MOP is used without MiHo;
    \item $(\mathrm{H}_{m_1},\mathrm{H}_{m_2})$ in case of MOP+MiHo;
    \item $(\mathrm{H}_v,\mathrm{H}_v)$ in case of MOP+VSAC;
    \item $(\mathrm{H}^b_1,\mathrm{H}^b_2)$ as above otherwise.
\end{enumerate}
Since the found transformations are approximated, the extended homography pair is perturbed in practice by small shear factors $f$ and rotation changes $\rho$ through the affine transformations \begin{equation}
	\mathrm{A}_{\rho,f}=\begin{bmatrix}
		f\cos(\rho)&\text{-}f\sin(\rho)&0\\
		\hfill\sin(\rho)&\hfill\cos(\rho)&0\\
		0&0&1
	\end{bmatrix}
\end{equation}
with
\begin{align}
    &\rho\in\mathcal{R}=\{-\tfrac{\pi}{6},-\tfrac{\pi}{12},0,\tfrac{\pi}{12},\tfrac{\pi}{6}\}\\
    &f\in\mathcal{F}=\{\tfrac{5}{7},\tfrac{5}{6},1,\tfrac{6}{5},\tfrac{7}{5}\}    
\end{align}
so that
\begin{align}\label{local_image_warping}\notag
    \mathcal{H}_m=&\left\{\left(\mathrm{H}^b_1,\mathrm{H}^b_2\right)\right\}\\ \notag
    \cup&\left\{\left(\mathrm{A}_{\rho,f}\mathrm{H}^e_1,\mathrm{H}^e_2\right): (\rho,s)\in \mathcal{R}\times\mathcal{F}\right\}\\
    \cup&\left\{\left(\mathrm{H}^e_1,\mathrm{A}_{\rho,f}\mathrm{H}^e_2\right): (\rho,s)\in \mathcal{R}\times\mathcal{F}\right\}
\end{align}

For each homography pair in $\mathcal{H}_m$, the associated patches are warped accordingly and NCC is used to get the pixel offset which identify the estimate discrete refinement. The keypoint offset corresponding to the homography pair globally maximizing the NCC similarity is finally selected and refined with sub-pixel precision by parabolic interpolation. Please refer to Appendix~\ref{ncc_appendix} for more details.

It is well known that NCC template matches is not suitable in the case of flat, homogeneous patches or in the presence of edges~(\cite{zeliski_book}). NCC can be extended to alleviate this issue, by limiting the offset search through a mask obtained by inscribing a maximum ellipsis with the shape provided by the gradient autocorrelation matrix on the squared patch that defines NCC search space. In this way, for instance, if the patch content is just an edge in a particular direction, NCC refinements are mainly attempted along the orthogonal edge direction, where there is an effective confidence of performing this operation. This extended NCC will be denoted by NCC$^\star$.

\section{Evaluation}\label{eval}
\subsection{Setup}\label{setup}
\subsubsection{Image matching pipelines and datasets}\label{base_pipeline}
Seven base pipelines have been tested to provide the input matches to be filtered. These include \textbf{SIFT+NNR}~(\cite{sift}) as the standard handcrafted reference, \textbf{Key.Net+} \textbf{$\substack{\text{AffNet}\\\text{HardNet}}$+NNR}~(\cite{imw2020}) as a modular almost deep pipeline, \textbf{SuperPoint+LightGlue}, \textbf{DISK+LightGlue} and \textbf{ALIKED+LightGlue}~(\cite{lightglue}) as sparse fully end-to-end deep image matching architectures, \textbf{DeDoDe v2}~(\cite{dedodev2}) which provides an alternative end-to-end deep network different from the previous ones, and \textbf{LoFTR}~(\cite{loftr}), a semi-dense detector-free end-to-end network. Additionally, further specific experiments have been carried out with the recent \textbf{MASt3R}~(\cite{mast3r}) and \textbf{RoMa}~(\cite{roma}) dense image matching architectures, for a total of nine base pipelines.

The maximum number of keypoints was set to 8000 to retain as many as possible matches, with the exception of more recent MASt3R and RoMa. This leads to including more outliers and better assessing the filtering potential. For MASt3R and RoMa the number of maximum output matches was set to 2048, considering computation resource constraints. Further details for each specific pipeline can be found in Appendix~\ref{pipeline_configs}.

Four different datasets have been employed in the evaluation, detailed descriptions can be found in Appendix~\ref{dataset_details}. Non-planar datasets include \textbf{MegaDepth}~(\cite{megadepth}) and \textbf{ScanNet}~(\cite{scannet}), respectively outdoor and indoor, together with the \textbf{IMC PhotoTourism (IMC-PT)} dataset\footnote{\label{imc2022}\url{https://www.kaggle.com/competitions/image-matching-challenge-2022/data}} which provides further insight in the case of outdoor scenes. Lastly, the \textbf{Planar} dataset contains around 140 image pairs collected from HPatches~(\cite{hpatches}), the Extreme View dataset (EVD,~\cite{mods}) and further datasets aggregated by~\cite{slime}.

\subsubsection{Match filtering and post-processing}\label{filtering}
Besides the proposed modular solution, thirteen match filters are applied after the base pipelines and included in the evaluation. Concerning proposed modular solutions, several combinations have been tested to analyze and better acknowledge the different behaviors, also in comparison with known methods. For instance, applying NCC after SIFT+NNR or Key.Net+$\substack{\text{AffNet}\\\text{HardNet}}$+NNR can highlight the improvement introduced by MOP which uses homography-based patch normalization against similarity or affine patches, respectively. Likewise, applying NCC to the remaining end-to-end architectures should remark the importance of patch normalization. The setup parameters indicated in Sec.~\ref{mop_miho_ncc} are employed, which have worked well in preliminary experiments. Notice that MiHo orientation adjustment is applied although not required in the case of upright images. This would indirectly assess the correctness of base assumptions and the overall general method robustness because in the case of wrong orientation estimation the consequential failure can only decrease the match quality.

To provide a fair and general comparison, the analysis was restricted to methods that require as input only the keypoint positions of the matches and the image pair. This excludes approaches requiring in addition descriptor similarities~(\cite{lpm,pfm,pgm}), related patch transformations~(\cite{sift_affine}), specific non-general pipelines~(\cite{patch2pix}) or a SfM framework~(\cite{pixelperfectsfm}). The compared methods are \textbf{GMS}~(\cite{gms}), \textbf{AdaLAM}~(\cite{adalam}), \textbf{CC}~(\cite{cc}) and \textbf{Progressive-X}~(\cite{progx}) as handcrafted filters, and \textbf{ACNe}~(\cite{acne}), \textbf{CLNet}~(\cite{clnet}), \textbf{ConvMatch}~(\cite{convmatch}), \textbf{DeMatch}~(\cite{dematch}), \textbf{FC-GNN}~(\cite{fcgnn}), \textbf{MS$^2$DG-Net}~(\cite{ms2dgnet}), \textbf{NCMNet}~(\cite{ncmnet}), \textbf{OANet}~(\cite{oanet}) and \textbf{Keypt2Subpx}~(\cite{keypt2subpx}) as deep modules. The implementations from the respective authors have been employed for all methods with the default setup parameters if not stated otherwise. Keypt2Subpx is only evaluated with SuperPoint+LightGlue since there are no pretrained models available for the other specific base pipeline employed in this evaluation.

With the exception of FC-GNN, Keypt2Subpx and OANet, deep filters require as input the intrinsic camera matrices of the two images, which is not commonly available in real practical scenarios the proposed evaluation was designed for. To bypass this restriction, the approach presented in ACNe was employed, for which the intrinsic camera matrix
\begin{equation}
	\mathrm{K}=\begin{bmatrix}
		f & 0 & c_x\\
		0 & f & c_y\\
		0 & 0 & 1
	\end{bmatrix}
\end{equation}
for the image $I$ with a resolution of $w\times h$ pixel is estimated by setting the focal length to
\begin{equation}
	f=\max(h,w)
\end{equation}
and the camera centre as
\begin{equation}
	\mathbf{c}=\begin{bmatrix}
		c_x\\c_y
	\end{bmatrix}=\frac{1}{2}
	\begin{bmatrix}
		w\\h
	\end{bmatrix}	
\end{equation}
The above focal length estimation $f$ is quite reasonable according to the statistics reported in Appendix~\ref{dataset_details} that have been extracted from the evaluation data. Notice also that most of the deep filters have been trained on SIFT matches, yet to be robust and generalizable they must be able to depend mainly only on the scene and not on the kind of feature extracted by the pipeline.

For AdaLAM, no initial seeds or local similarity or affine clues were used as additional input as discussed above. Nevertheless, this kind of data could have been used only for SIFT+NNR or Key.Net+$\substack{\text{AffNet}\\\text{HardNet}}$+NNR. For CC and Progressive-X, the inliers threshold was experimental set to $t_h=7$ px, as defined for MOP in Sec.~\ref{mop}, instead of the value of 1.5 px suggested by authors. Otherwise, both filters would not be able to work in general scenes with no planes. Note that AdaLAM, CC and Progressive-X are the closest approaches to the proposed MOP-based filtering pipelines. Moreover, among the compared methods, only FC-GNN and Keypt2Subpx are able to refine matches, and only FC-GNN is able to both filtering and refining matches as the proposed full solution does.

RANSAC is also optionally evaluated as the final post-process step to filter matches according to epipolar or planar constraints. Three different RANSAC implementations were initially considered, i.e. the RANSAC implementation in PoseLib\footnote{\url{https://github.com/PoseLib/PoseLib}}, DegenSAC\footnote{\url{https://github.com/ducha-aiki/pydegensac}} and MAGSAC\footnote{\url{https://github.com/danini/magsac}}. The maximum number of iterations for all RANSACs was set to $10^5$, which is uncommonly high, in order to provide a more robust pose estimation and to make the whole pipeline mostly depending on the previous filtering and refinement stages. Five different inlier threshold values $t_{SAC}$, i.e.
\begin{equation}
	t_{SAC} \in \{0.5\,\text{px},\,0.75\,\text{px},\,1\,\text{px},\,3\,\text{px},\,5\,\text{px}\}
\end{equation}
were considered. After a preliminary RANSAC ablation study\footnote{\label{github_results}\url{https://github.com/fb82/MiHo/tree/main/data/results}} according to proposed benchmark, the best general choice found is \textbf{MAGSAC} with $t_{SAC}=0.75$ px, followed by MAGSAC with $t_{SAC}=1$ px. On one hand, the former and more strict threshold is generally slightly better in terms of accuracy. On the other hand, the latter provides more inliers. MAGSAC results with $t_{SAC}=1$ px are provided as additional material for most of the evaluation configurations. Matching outputs \textbf{without RANSAC} are also reported to complete the analysis.

Finally, as the main focus of the this analysis is to provide a better understanding and interconnection of the matching capability of geometric strategies with respect to deep approaches without focusing on strict engineering optimizations, running time considerations are reported in Appendix~\ref{appendix_time} for completeness.

\subsubsection{Error metrics}\label{error_definitions}
For non-planar datasets, unlike most common benchmarks~(\cite{loftr}), it is assumed that no Ground-Truth (GT) camera intrinsic parameters are unavailable. This is the general setting that occurs in most practical SfM scenarios, especially in outdoor environments. For a given image pair, the fundamental matrix $\mathrm{F}$ can be estimated as
\begin{equation}\label{fund_mat}
	\mathrm{F} = \mathrm{K}_2^{-T} \mathrm{E}\, \mathrm{K}^{-1}_1 = \mathrm{K}_2^{-T} [\mathbf{t}]_\times \mathrm{R}\, \mathrm{K}^{-1}_1
\end{equation}
where $\mathrm{E},\mathrm{R},\mathrm{K}_1,\mathrm{K}_2,[\mathbf{t}]_\times\in\mathbb{R}^{3\times3}$ are respectively the essential matrix, the rotation matrix, the intrinsic camera matrices and the skew-symmetric matrix corresponding to the translation vector $\mathbf{t}\in\mathbb{R}^3$~(\cite{multiview}). The maximum epipolar error for a match $m=(\mathbf{x}_1,\mathbf{x}_2)$ according to $\mathrm{F}$ is
\begin{equation}
	\varepsilon_\mathrm{F}(m)=\max\left(\frac{\mathbf{x}_2^T\mathrm{F}\mathbf{x}_1}{\left\|[\mathrm{F}\mathbf{x}_1]_{[1:2]}\right\|^2},\frac{\mathbf{x}_1^T\mathrm{F}^T\mathbf{x}_2}{\left\|[\mathrm{F}^T\mathbf{x}_2]_{[1:2]}\right\|^2}\right)
\end{equation}
where points are considered in normalized homogeneous coordinates and $[\mathbf{x}]_{[1:2]}$ denotes the vector obtained by the first two elements of $\mathbf{x}$. For planar images, the reprojection error $\varepsilon_\mathrm{H}(m)$ defined in Eq.~\ref{reproj_error} will be used instead.

Given a set of matches $\mathcal{M}_b=\{m\}$, obtained by a base pipeline of those presented in Sec.~\ref{base_pipeline}, and the final refined match set $\mathcal{M}$ according to one of the successive post-processing listed in Sec.~\ref{filtering}, the GT fundamental matrix $\mathrm{\tilde{F}}$ is estimated by Eq.~\ref{fund_mat} from the GT intrinsic and extrinsic camera parameters\footnote{Actually these are quasi or pseudo GT data estimated from wide and reliable SfM models.} in the case of the non-planar scene, or the GT homography $\mathrm{\tilde{H}}$ is provided for planar scenes. The notation $\sfrac{\mathrm{\tilde{F}}}{\mathrm{\tilde{H}}}$ indicates the usage of the GT matrix according to the scene kind. The \textbf{recall} is computed as
\begin{equation}\label{recall}
	\text{recall}^{\mathcal{M}_b}_{\sfrac{\mathrm{\tilde{F}}}{\mathrm{\tilde{H}}}}(\mathcal{M})=\frac{\sum_{\substack{m\in\mathcal{M}\\t\in\mathcal{T}}} \left\llbracket\varepsilon_{\sfrac{\mathrm{\tilde{F}}}{\mathrm{\tilde{H}}}}(m)<t\right\rrbracket}{\sum_{\substack{m'\in\mathcal{M}_b\\t'\in\mathcal{T}}} \left\llbracket\varepsilon_{\sfrac{\mathrm{\tilde{F}}}{\mathrm{\tilde{H}}}}(m')<t'\right\rrbracket}
\end{equation}
and the \textbf{precision} as
\begin{equation}
	\text{precision}_{\sfrac{\mathrm{\tilde{F}}}{\mathrm{\tilde{H}}}}(\mathcal{M})=\frac{1}{|\mathcal{M}|\,|\mathcal{T}|}\sum_{\substack{m\in\mathcal{M}\\t\in\mathcal{T}}} \left\llbracket\varepsilon_{\sfrac{\mathrm{\tilde{F}}}{\mathrm{\tilde{H}}}}(m)<t\right\rrbracket
\end{equation}
where the common definitions have been modified following the MAGSAC strategy to consider a set of threshold values 
\begin{equation}
	\mathcal{T}=\left\{t\in[1, t_\varepsilon]\cap\mathbb{Z}\right\}	
\end{equation}
limited by $t_\varepsilon=16$. As additional statistics, the number of \textbf{filtered} matches is computed as
\begin{equation}
	\text{filtered}_{\mathcal{M}_b}(\mathcal{M}) = 1 - \frac{|\mathcal{M}|}{|\mathcal{M}_b|}
\end{equation}
For a zero denominator in Eq.~\ref{recall} the recall is forced to 0. Notice that in the case of a non-zero denominator, the recall is less than or equal to 1 only when $\mathcal{M}\subseteq\mathcal{M}_b$ which does not hold for pipelines where the keypoint refinement step effectively improves the matches. The number of filtered matches and the precision are less than or equal to 1 in any case by pipeline design. The final aggregated measurement for each error and dataset is provided by averaging on all dataset image pairs. It should be remarked that for non-planar scenes both precision and recall are non-perfect measures since epipolar error constraints can hold also for outliers in specific and common scene configurations~(\cite{dtm}).

To provide more reliable measures also in view of the final aim of the image matching, the relative pose error is employed for the non-planar scenes. Two alternative estimations of the pose are given in terms of rotation and translation matrices. In the first one, the fundamental matrix is estimated from the matches in $\mathcal{M}$ by the 8-point algorithm using the normalized Direct Linear Transformation (DLT) by the OpenCV function \texttt{findFundamentalMat} as $\mathrm{F}_\mathcal{M}$ and then the essential matrix is obtained as
\begin{equation}
	\mathrm{E}_\mathcal{M} = \mathrm{\tilde{K}}_2^{T} \mathrm{F}_\mathcal{M} \mathrm{\tilde{K}}_1
\end{equation} 
given the GT intrinsic cameras $\mathrm{\tilde{K}}_1$ and $\mathrm{\tilde{K}}_2$. The four possible choice of camera rotation and translation pairs are then obtained by SVD decomposition using the OpenCV \texttt{decomposeEssentialMat} function as
\begin{equation}\label{pose_fundamental}
	\mathcal{E}_\mathcal{M}=\left\{\mathrm{R}^1_\mathcal{M},\mathrm{R}^2_\mathcal{M}\right\}\times\left\{\pm\mathbf{t}_\mathcal{M}\right\}
\end{equation}
The second alternative follows the standard benchmarks. After having normalized keypoint matches according to the GT instrinsic cameras $\mathrm{\tilde{K}}_1$ and $\mathrm{\tilde{K}}_2$ so that
\begin{equation}
	\mathcal{\widehat{M}}=\left\{\left(\mathrm{\tilde{K}}_1\mathbf{x}_1,\mathrm{\tilde{K}}_2\mathbf{x}_2\right): \left(\mathbf{x}_1,\mathbf{x}_2\right)\in\mathcal{M}\right\}
\end{equation}
the essential matrix is computed directly as $\mathrm{E}_\mathcal{\widehat{M}}$ by the OpenCV function \texttt{findEssentialMat} using the 5-point algorithm~(\cite{nister}) embedded into RANSAC with a normalized threshold $t_\mathrm{E}=0.5$~\cite{loftr} and the maximum number of iteration equal to $10^4$. The two possible solutions for the camera rotation and translation components are recovered by the OpenCV function \texttt{recoverPose} as
\begin{equation}\label{pose_essential}
	\mathcal{E}_\mathcal{\widehat{M}}=\left\{\mathrm{R}_\mathcal{\widehat{M}}\right\}\times\left\{\pm\mathbf{t}_\mathcal{\widehat{M}}\right\}
\end{equation}
Notice that this second approach makes it easier to find a correct solution since five points are required instead of eight for the minimum model and $\mathrm{\tilde{K}}_1$ and $\mathrm{\tilde{K}}_2$ are used earlier stage of the process. Nevertheless, these conditions are more unlikely to be met in practical real situations, especially in outdoor environments. Note that the translation vectors in both $\mathcal{E}_\mathcal{M}$ and $\mathcal{E}_\mathcal{\widehat{M}}$ are defined up to scale. The final pose accuracy is provided in terms of the Area Under the Curve (AUC) of the best maximum angular error estimation for each dataset image pair and a given angular threshold 
\begin{equation}\label{angle_theta}
	\theta\in\left\{5^\circ,10^\circ,20^\circ\right\}	
\end{equation}
The maximum angular error for an image pair is
\begin{equation}\label{angle_error}
	\rho(\mathcal{M})=\min_{\left(\mathrm{R},\mathbf{t}\right)\in\sfrac{\mathcal{E}_\mathcal{M}}{\mathcal{E}_\mathcal{\widehat{M}}}  }\max\left(\rho\left(\mathbf{\tilde{t}},\mathbf{t}\right),\rho\left(\mathrm{\tilde{R}},\mathrm{R}\right)\right)
\end{equation}
with $\sfrac{\mathcal{E}_\mathcal{M}}{\mathcal{E}_\mathcal{\widehat{M}}}$ indicating the use of one of the two options between Eq.~\ref{pose_fundamental} and Eq.~\ref{pose_essential}, $\mathbf{\tilde{t}}$ and $\mathrm{\tilde{R}}$ being the respective GT translation, defined again up to scale, and rotation. In the above formula, the angular error for the translation and rotation are respectively computed as
\begin{align}\label{translation_angle}
	\rho(\mathbf{\tilde{t}},\mathbf{t})=&\text{acos}\left(\frac{\mathbf{\tilde{t}}^T\mathbf{t}}  {\left\|\mathbf{\tilde{t}}\right\|\,\left\|\mathbf{t}\right\|}\right)\\
	\rho(\mathrm{\tilde{R}},\mathrm{R})=&\text{acos}\left(\frac{1}{2}\left(1-\cos\left(\mathrm{\tilde{R}}^T \mathrm{R}\right)\right)\right)
\end{align}
The two obtained AUC values for the chosen pose estimation between Eq.~\ref{pose_fundamental} and Eq.~\ref{pose_essential} and a fixed threshold $\theta$ will be denoted by AUC$^F_{@\theta}$, AUC$^E_{@\theta}$ respectively, with the superscript indicating the kind of error estimator employed, i.e. `$F$' for fundamental matrix as in Eq.~\ref{pose_fundamental}, `$E$' for essential matrix as in Eq.~\ref{pose_essential}. The final aggregated pose accuracy measures are their average over the $\theta$ values, indicated as \textbf{AUC$^F_\measuredangle$} and \textbf{AUC$^E_\measuredangle$}, where the subscript `$\measuredangle$' refers to the use of an angular translation error according to Eq.~\ref{translation_angle} to differentiate with respect to the metric translation error introduced hereafter.

In the case of IMC-PT dataset, a fully metric estimation of the translation component of the pose is possible since the real GT reconstruction scale is available, i.e. $\left\|\mathbf{\tilde{t}}\right\|$ is the real metric measurement. Following the IMC 2022 benchmark\footref{imc2022}, the angular thresholds $\theta$ of Eq.~\ref{angle_theta} are replaced by pairs of thresholds $(\theta,t_+)$, where $t_+$ is a translation thresholds. The corresponding AUC values for the chosen pose estimation between Eq.~\ref{pose_fundamental} and Eq.~\ref{pose_essential} will be denoted by AUC$^F_{@(\theta,t_+)}$, AUC$^E_{@(\theta,t_+)}$ respectively, for the given threshold pair $(\theta,t_+)$. The final metric accuracy scores are given by the average over the $(\theta,t_+)$ value pairs and denoted as \textbf{AUC$^F_\square$} and \textbf{AUC$^E_\square$}, respectively, where the `$\square$' subscript indicates the metric translation error. 

For the planar dataset, the homography accuracy is measured analogously by AUC, replacing the pose error by the maximum of the average reprojection error in the common area when using one of the two images as a reference, leading to the AUC$^H_{@t_\mathrm{H}}$ value for a given threshold $t_\mathrm{H}$. The final homography accuracy is provided as the mean AUC over the different thresholds and is indicated by \textbf{AUC$^H_\square$} to be consistent with the previous AUC notation adopted in the case of non-planar images, where the superscript `$H$' clearly stands for homography.

Detailed descriptions on both the fully metric and planar AUC estimations can be found in Appendix~\ref{error_metric_details}.

\subsection{Results}
\subsubsection{Prelude}\label{prelude}
In the next discussion, the base pipelines will be referred to for brevity by the first component, e.g. Key.Net will refer to Key.Net+$\substack{\text{AffNet}\\\text{HardNet}}$+NNR. Tables~\ref{sift_res}-\ref{aliked_res} report the results obtained according to the setup presented in Sec.~\ref{setup} for SIFT, Key.Net, SuperPoint, LoFTR and ALIKED. For DISK and DeDoDe v2 results are similar to ALIKED and are reported in Appendix~\ref{more_results}. The left aligned shift in the pipeline column indicates incremental inclusion of the modules as in the folder-tree style visualization. All the values in the tables are expressed in percentages. Extended tables including AUCs values in the non-aggregated form, i.e. for the different thresholds, are provided as online material\footref{github_results}. In any case, non-aggregated AUC ranks are almost stable and well correlated with the corresponding aggregated measure.   

The proposed pipeline modules are in bold. For the other numeric columns, the three highest values are also in bold, with the displayed bars normalized within the minimum and maximum value of the column. Increasing color shades of the bars identify the value range among the four incremental percentage intervals with exponentially decreasing length contained in $[0,50,75,82.5,100] \%$ after the value has been normalized by the minimum and maximum of each column. Darker shades indicate exponentially better ranks.

CC filtering was not evaluated on IMC-PT because it experienced a high number of unreasonable crashes for this dataset, neither Progressive-X and Keypt2Subpx, since they were included later and according to the preliminary analysis further evaluations would have not provided any interesting additions.

There are obvious factors to take into account when analyzing the results:
\begin{enumerate}[a)]
	\item When using a robust outlier estimation (i.e. MAGSAC or the base essential matrix estimation which implicitly runs RANSAC) small differences in AUC if not exhibiting regularly among the different pipelines and datasets, are not effectively relevant since mainly due to randomization.
	\item The estimation of the fundamental and homography matrices without a RANSAC-based robust inlier framework becomes basically a least-square minimization problem so that a few strong outliers may affect the final result very badly.
	\item A low number of very accurate matches may be not sufficient for a precise fundamental matrix estimation, not only due to possible degenerate configurations (i.e. matches only lying on a plane), but also because epipolar line estimations far from the point locations of the matches (i.e. outside the related convex hulls in the images) generally become inaccurate.
	\item That said, having more matches on the images, even if less accurate but better distributed, can lead to a better fundamental matrix estimation. It depends on the specific input images and matches, the pipeline, the RANSAC employed and its parameters.
	\item More constrained problems, i.e. homography and essential matrix estimations, are more robust to pipeline providing less accurate matches.
	\item The more the base pipeline precision is, the less the filter improves the matches. In any case a filter should not degrade the matching performances.
\end{enumerate}

According to above points, referring to the tables, the pose estimation from the fundamental matrix is harder than that from the essential matrix. This can be observed by the value differences in the sided green and yellow columns. Moreover, using MAGSAC before RANSAC-based essential matrix estimation in general does not affect or even slightly decreases the accuracy. The behavior of the different matching filters is stressed when the pose is estimated from the fundamental matrix. For IMC-PT the corresponding angular (green and yellow columns) and metric (purple and olive green columns) AUC scores maintain the same relative order even if the metric scores are lower in terms of absolute values, thus highlighting that real pose is still far to be found. The previous observations are further confirmed by the analysis of the correlation between the employed error measures, reported in Appendix~\ref{corr_appendix}. The average number of input matches for each base method and dataset is provided in Appendix~\ref{num_appendix}.

To summarize, it is better to have more matches for pose registration even with higher levels of outlier contamination when the camera intrinsic parameters are available, the opposite holds when camera parameters are missing. Moreover, RANSAC flattens scores and strongly influences the final matches. Nevertheless, one error measure cannot replace another, although highly correlated, since they highlight different properties and behaviors of the matching pipeline. 

According to the previous observations, the score achieved by filters without MAGSAC will be taken into consideration for the essential matrix estimation and results obtained with MAGSAC in the case of fundamental matrix or homography estimations.

The detailed comparative analysis of the considered filtering modules for the different base matching pipelines in terms of pose accuracy, as well as the recall and precision of the matches, is reported hereafter.

\begin{table}
	\caption{SIFT+NNR pipeline evaluation results. All values are percentages, please refer to Sec.~\ref{sift_text}. Best viewed in color and zoomed in.}\label{sift_res}
	\includegraphics[width=1\textwidth]{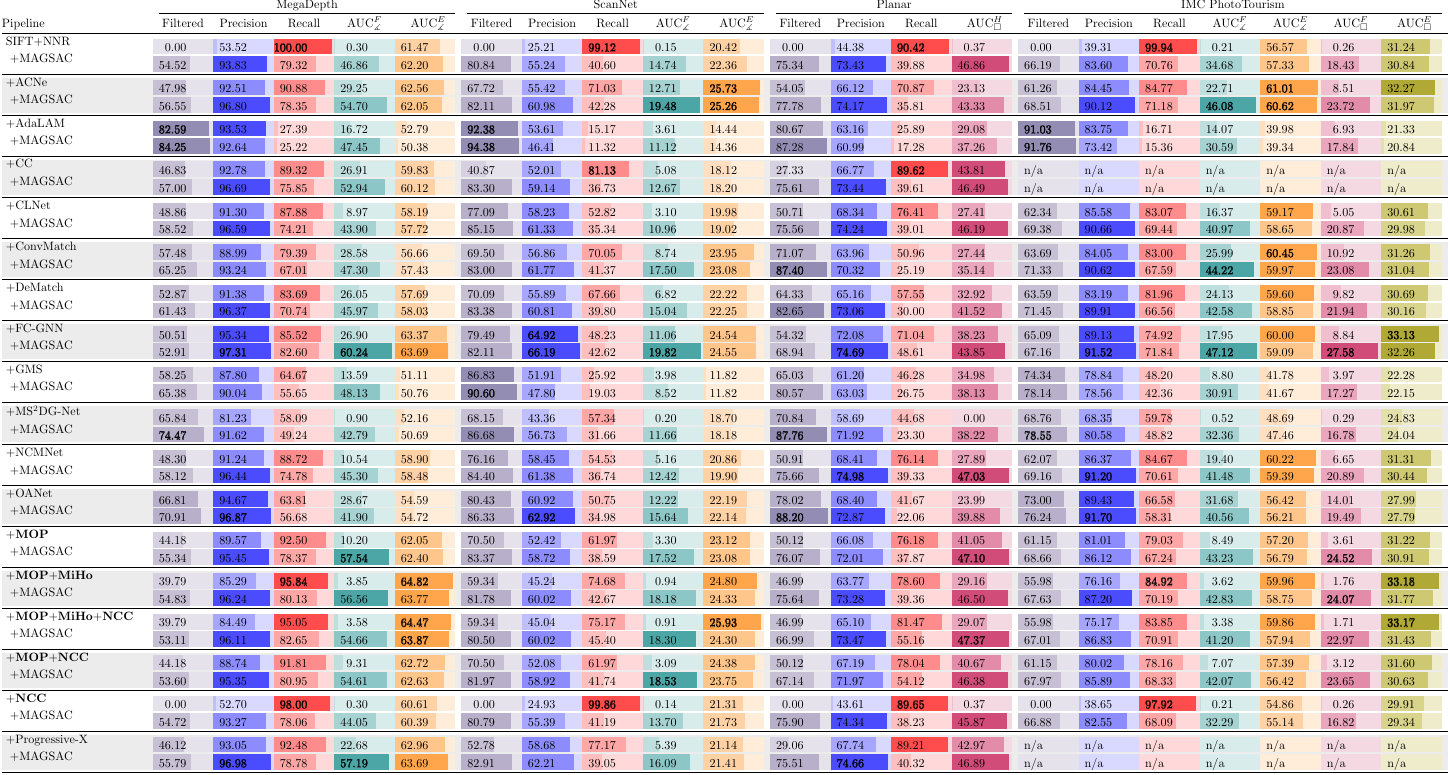}
\end{table}

\subsubsection{SIFT+NNR}\label{sift_text}
Table~\ref{sift_res} reports the results for SIFT. MOP+MiHo works overall quite well, achieving top rank scores or close to them in all dataset and configurations. Adding NCC somewhat degrades the scores for the fundamental matrix estimation with MAGSAC on MegaDepth and IMC-PT, probably due to a prominent presence of flat blob-like keypoint patches which are not suitable for NCC template matching. FC-GNN obtains top rank scores or close to these, while ACNe is quite good for ScanNet and both IMC-PT and Progressive-X on MegaDepth. Most of the other filters do better than the base pipeline for the non-planar datasets, but only for those mentioned above the gain is remarkable.

In the case of planar scenes, the differences of the filtering pipelines with respect to the base pipeline are quite limited. Filters based on planes like the MOP-based ones, CC and Progressive-X are generally more stable, in the sense that they do not degrade the base solution. Notice than FG-CNN is effectively worse than the base pipeline with the planar scenes. 

MiHo reduces the MOP overfit towards planar structures, a behavior that the base MOP has in common with CC and Progressive-X due to their designs. This is highlighted by the higher number of filtered matches of MOP without MiHo that leads in turn to a higher AUC accuracy without MAGSAC for the fundamental or homography estimations. Nevertheless, including robust estimators in the pipeline inverts or cancels out this trend as discussed in Sec.~\ref{prelude}.

As final consideration, FC-GNN and MOP-based filters tend to reach the highest level of precision and recall possible, respectively. This in turn leads to have high accuracy in term of AUC for fundamental, essential matrix estimations for FC-GNN and MOP-based filters, respectively.

As previously mentioned and confirmed by a private discussion with authors, ACNe has an overlap with the training data used for the estimation of the indoor and outdoor model weights with ScanNet and IMC-PT, respectively. Moreover, although training data used to compute FC-GNN weights and the validation data are different for Megadepth, the kinds of scenes are similar. For SIFT as well as other base pipelines discussed next, the gap within AUC accuracy computed with fundamental and essential matrix estimations for FG-CNN is quite limited in the case of MegaDepth and increases in order with IMC-PT and ScanNet. This suggests, together with the analysis of the intrinsic camera parameters reported in Appendix.~\ref{dataset_details}, the possibility that FC-GGN implicitly learned the camera intrinsic priors for the specific MegaDepth dataset, partially overlapping with IMC-PT, that cannot be always exploited in other non-planar datasets. Differently, being handcrafted, MOP is not tuned on the data.

\begin{table}
	\caption{Key.Net+$\protect\substack{\text{AffNet}\\\text{HardNet}}$+NNR pipeline evaluation results. All values are percentages, please refer to Sec.~\ref{keynet_text}. Best viewed in color and zoomed in.}\label{keynet_res}
	\includegraphics[width=1\textwidth]{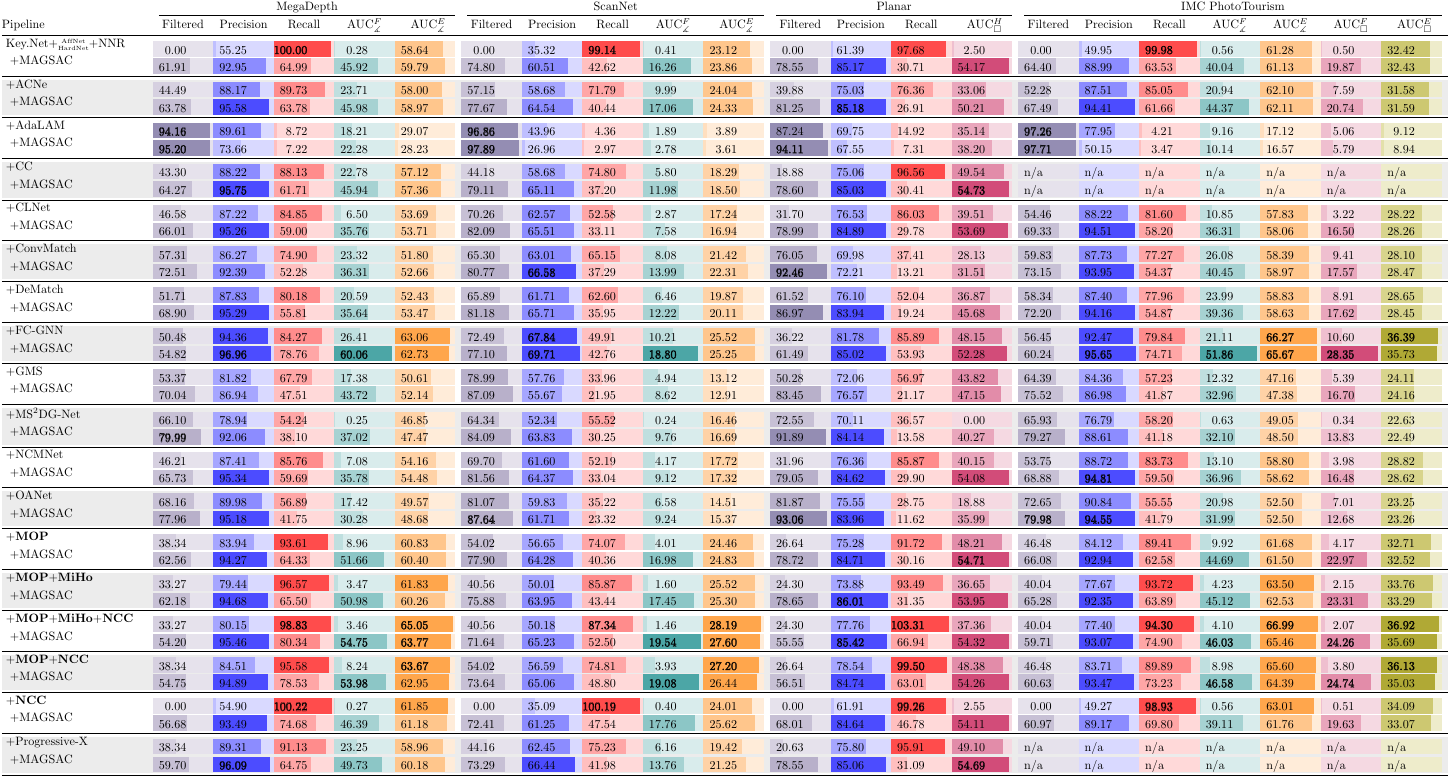}
\end{table}

\subsubsection{Key.Net+AffNet+HardNet+NNR}\label{keynet_text}
Table~\ref{keynet_res} reports the the results for Key.Net. The best scores for all datasets are obtained by FC-GNN and MOP+ MiHo+NCC. The inclusion of NCC in this case is effective, since the kind of keypoints provided by the base Key.Net pipeline are almost corners by design, where NCC template matching is robust. As for SIFT, FC-GNN tends to increase more the precision while MOP-based filters the recall, leading to the previous observations. Again, FC-GNN worsens the performances of the base pipeline with MAGSAC on the planar dataset.

Overall, the best score achieved by any Key.Net pipeline with or without matching filters is lower than the corresponding SIFT one, with the relevant exception of Key.Net with MOP-based filtering and NCC. Within the MOP-based pipelines, Key.Net plus MOP+MiHo+NCC obtains the best absolute scores also among SIFT-based pipelines when the pose accuracy is estimated from the essential matrix. This indirectly supports the different nature of the SIFT blobs and Key.Net corners. From the authors' experience, on one hand, blobs are characterized to be more flat-symmetric and hence less discriminative than corners. However, on the other hand,for the same reasons keypoint localization is more stable for blobs than corners. A proper refinement step as NCC thus improves the matching process for corner-like features. Notice also that the recall improvement obtained by adding NCC with respect to the same pipeline without NCC, and in minor part also the precision, give indications of previous matches that were outliers due to a low keypoint accuracy and have been turned into inliers.  

All the filters yield worse scores than the base pipeline plus MAGSAC for Key.Net, with the exception of FC-GNN and MOP-based ones. This highlights a critical aspect of learning-based methods, that need an appropriate tuning and training according to the specific pipeline components. Notice also that the difference between the corresponding AUC$^F_\measuredangle$ and AUC$^E_\measuredangle$ for FC-GNN on MegaDepth is more limited than with other datasets, and in particular with respect to IMC-PT. A possible justification is that the network is able to learn camera intrinsic priors of the MegaDepth images, which have less variability as outdoor scenes than IMC-PT. As discussed in the evaluation setup, all deep filters have been trained on SIFT, including FC-GNN, suggesting a less generalization to different image features with respect to handcrafted approaches based on geometric constraints like MOP.

Finally, unlike SIFT, directly applying NCC after the base pipeline without MOP provides some improvements, even if these are inferior to when NCC is preceded by MOP-base filtering. This confirming the original assumption that affine patch normalization is better than normalization by similarity for the refinement, but not as good as planar patch normalization.

\begin{table}
	\caption{SuperPoint+LightGlue pipeline evaluation results. All values are percentages, please refer to Sec.~\ref{sp_text}. Best viewed in color and zoomed in.}\label{superpoint_res}
	\includegraphics[width=1\textwidth]{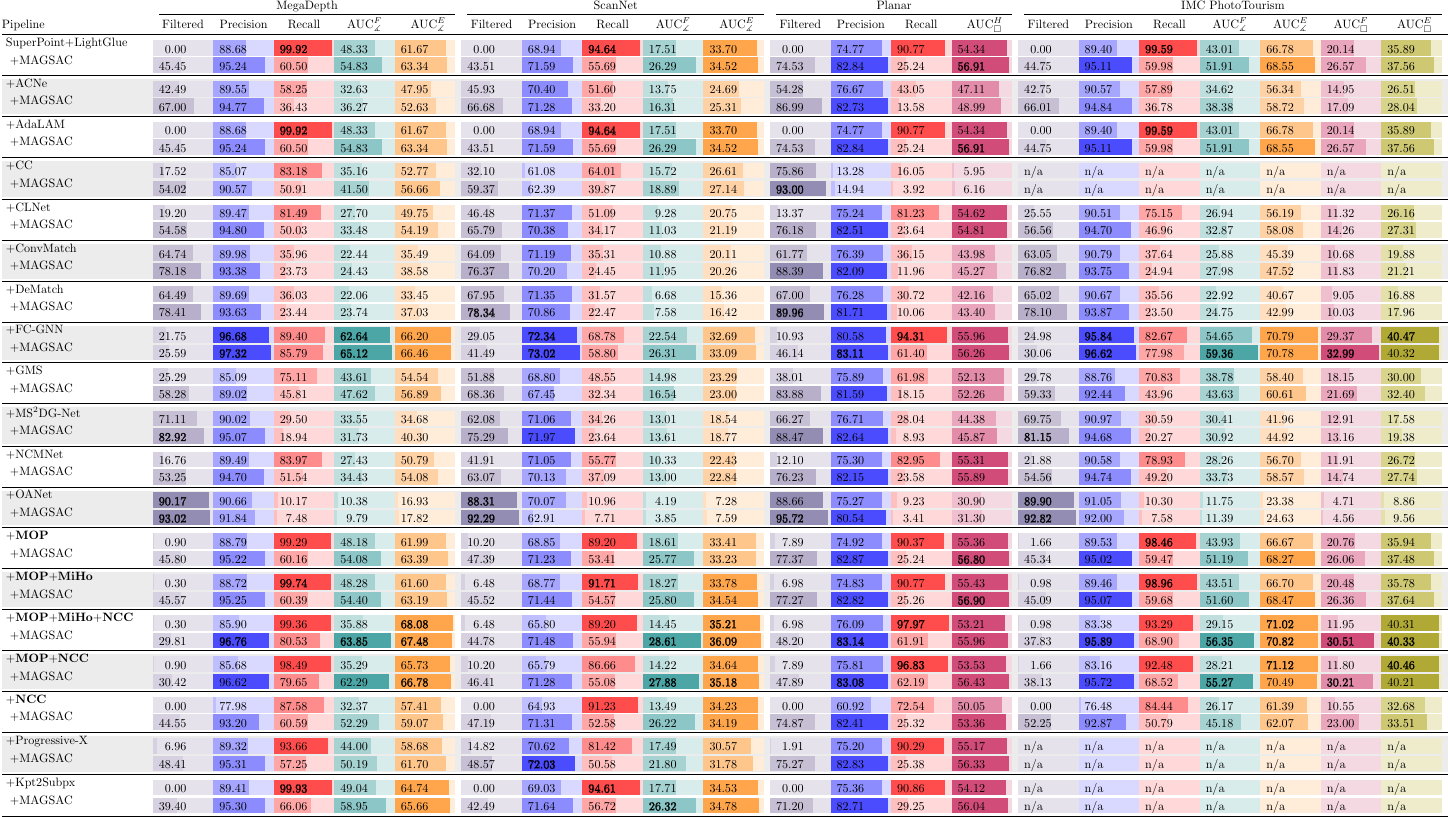}
\end{table}

\subsubsection{SuperPoint+LightGlue}\label{sp_text}
Table~\ref{superpoint_res} reports the results for SuperPoint. Overall, the relative rank among the datasets and pipelines for SuperPoint is comparable with Key.Net. This is not accidental since both Key.Net and SuperPoint are based on corner-like keypoints, so the same considerations and observations done for Key.Net hold for SuperPoint. Moreover, in terms of absolute score among all the evaluated sparse or semi-dense pipelines, SuperPoint with MOP+MiHo+NCC achieves on MegaDepth the best AUC$^E_\measuredangle$ value, or a value almost equal to the best absolute AUC$^E_\measuredangle$ provided by LoFTR for ScanNet. The improvement in adding NCC in terms of recall and precision with respect to Key.Net is inferior but still relevant, implying a more accurate corner relocalization of the base SuperPoint.

Note that for the planar case, MOP+MiHo+ NCC is slightly inferior with respect to the base pipeline, achieving the top score, and with respect to MOP+MiHo, for which the score is almost the same of the base pipeline. It is reasonable considering that the base SuperPoint training procedure involves the homographic adaptation which is actually achieved by performing image matching on planar scenes, so that the base pipeline training overfits on this specific task.

\begin{figure}
	\centering
	\subcaptionbox{\label{xa}}{\includegraphics[width=0.4\textwidth]{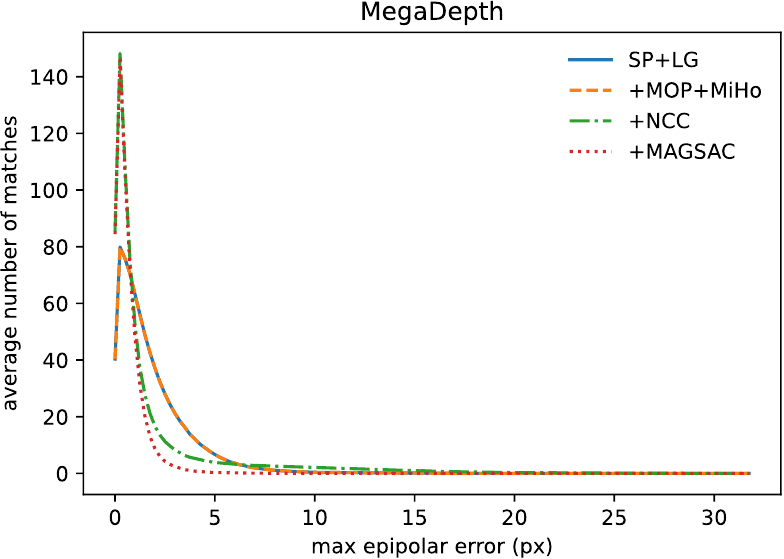}}
	\hspace{3em}
	\subcaptionbox{\label{xb}}{\includegraphics[width=0.4\textwidth]{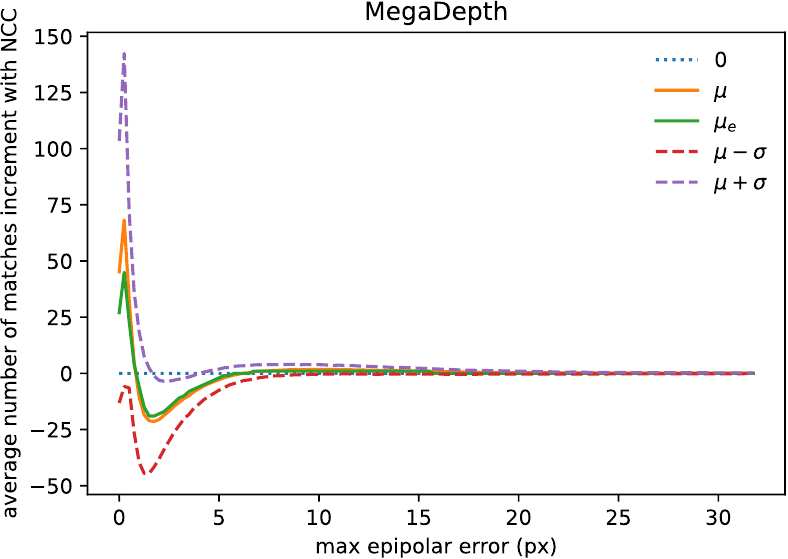}}
	\caption{(\subref{xa}) Average distribution of the maximum epipolar error of the SuperPoint pipeline after applying incrementally MOP+MiHo, NCC and MAGSAC on MegaDepth. (\subref{q2}) Average distribution variation after applying NCC. The mean, median and standard deviation are shown respectively as $\mu$, $\mu_e$ and $\sigma$. Notice the negative bump within 1 px and 5 px, with the increment of the left peak as well as of the right tail, which indicate that NCC refinement succeeds in most cases but sometimes increase the errors. Best viewed in color and zoomed in.\label{ncc_diff}}
\end{figure}

MOP-based filters without NCC are better in fundamental matrix estimation when no MAGSAC is employed, while the corresponding MOP-based filters with NCC work better when MAGSAC follows. Figure~\ref{xa} shows the distribution of the epipolar error of the matches as the next step of the pipeline is executed. NCC succeeds in refining most of the matches within a mid-range epipolar error, which for the specific configuration is about within 1 px and 5 px, as highlighted in the corresponding hollow in Fig~\ref{xb}. Nevertheless, NCC may fail in some cases as indicated by the small increase of the right tail of the plot, leading to higher errors for these few matches. Without MAGSAC the fundamental matrix estimation is not robust and hence a few strong outliers with high errors may affect the whole estimation. By adding MAGSAC these outliers are sweep out and the increased precision of the correctly refined matches improves the final result.

Finally, notice that results with Keypt2Subpx, which only refines and does not filter matches, are also included for MegaDepth and ScanNet with SuperPoint. While Keypt2-Subpx is able to improve the results of the base pipeline, its performances are inferior to those of MOP+MiHo+NCC and FC-GNN. This highlights that also RANSAC-based global filters like MAGSAC can benefits from a local filtering stage.

\begin{table}
	\caption{LoFTR pipeline evaluation results. All values are percentages, please refer to Sec.~\ref{loftr_text}. Best viewed in color and zoomed in.}\label{loftr_res}
	\includegraphics[width=1\textwidth]{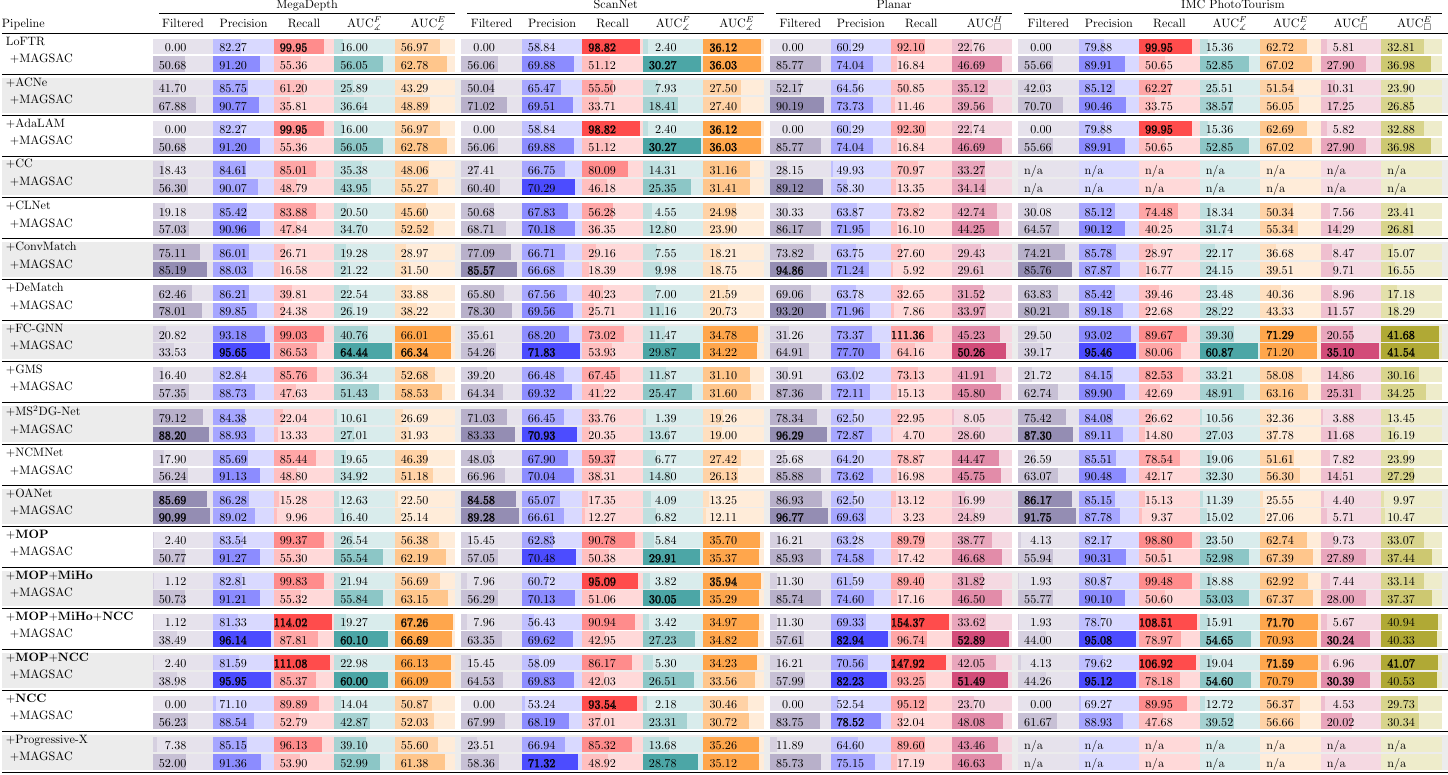}
\end{table}

\subsubsection{LoFTR}\label{loftr_text}
Table~\ref{loftr_res} reports the results for LoFTR. Unlike other architectures relying on SuperGlue, MAGSAC greatly improves the LoFTR base pipeline. FC-GNN and MOP+MiHo+NCC provide the best results with the exception of ScanNet. Moreover, NCC on this latter dataset degrades the accuracy noticeably. These results are reasonable considering that LoFTR is a semi-dense architecture specifically designed for ScanNet scenes, and there are many keypoints corresponding to flat regions not feasible for NCC template matching.

Excluding ScanNet, for the other datasets NCC greatly improves the AUCs, achieving the top scores, which are also the best as absolute values with respect to the sparse pipelines. This is highlighted by the high precision and the recall values exceeding those of the base LoFTR when using MOP-based filtering with NCC and without MAGSAC. As for previous base pipelines, FC-GNN is better than MOP-based filters with NCC in the case of pose estimation from the fundamental matrix. 

\begin{table}
	\caption{ALIKED+SuperGlue pipeline evaluation results. All values are percentages, please refer to Sec.~\ref{other_sparse}. Best viewed in color and zoomed in.}\label{aliked_res}
	\includegraphics[width=1\textwidth]{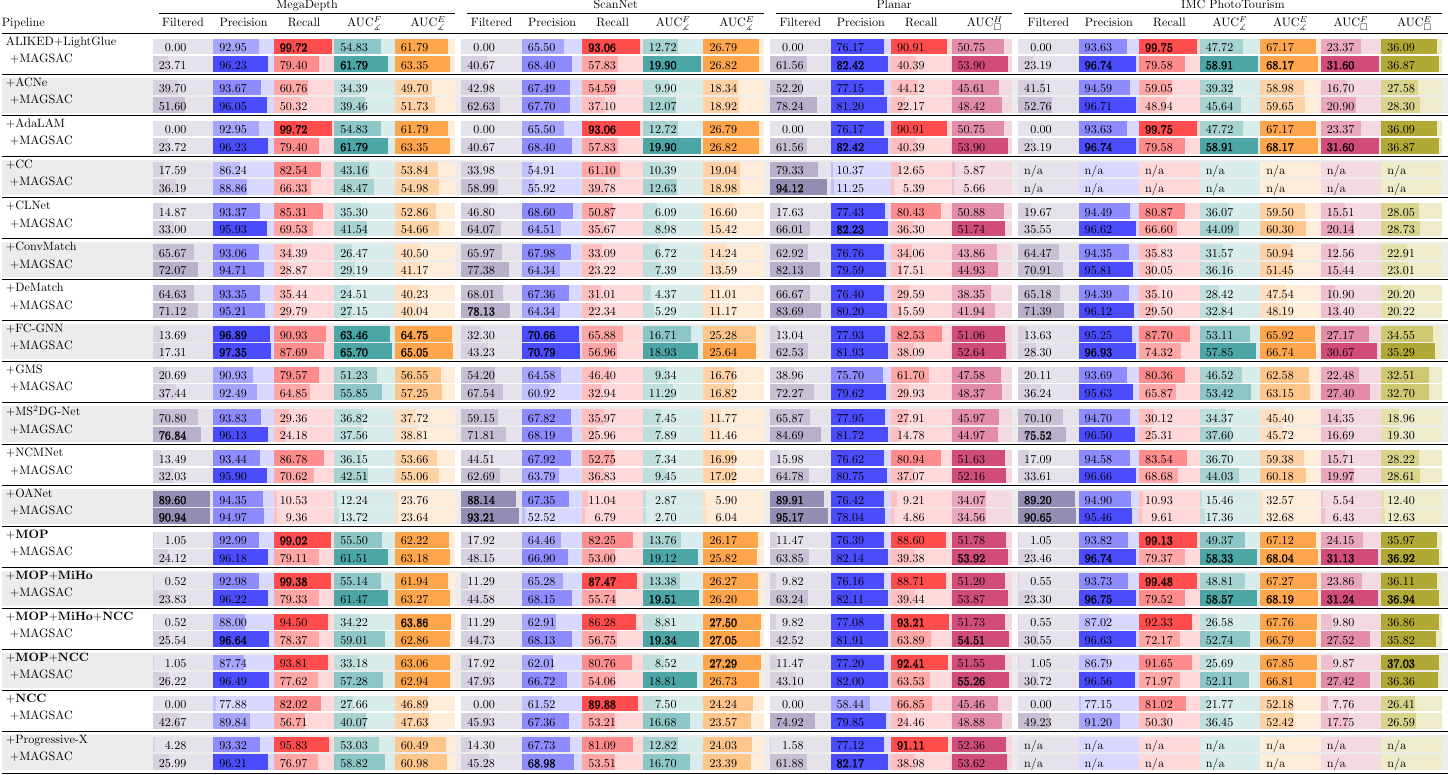}
\end{table}

\subsubsection{Other sparse pipelines}\label{other_sparse}
Table~\ref{aliked_res} reports the results for ALIKED, while those for DISK and DeDoDe v2 can be found in Appendix~\ref{other_sparse_appendix} since similar considerations hold. For ALIKED, the base pipeline is the best in almost all cases or very close to the best one, since by design ALIKED is trained to select only accurate matches and discard others. There are no match to be filtered and refined so that in the optimal case a filter or a refining module should not degrade the original score. Similar results on a different setups are reported by~\cite{keypt2subpx}. In this sense, only MOP-based filters, AdaLAM and FC-GNN are robust and stable. Notice that FC-GNN only surpasses the base ALIKED on MegaDepth where it was trained on, and NCC can slight degrade MOP-based filters since the ALIKED matches are already sub-pixel accurate. 

In the planar dataset, not involved in the ALIKED training, only MiHo-based filtering is better than the base pipeline while most of the other filters are slightly inferior to the base pipeline. The original ALIKED provides overall only robust and accurate base input matches differently from SuperPoint, that leads to better absolute AUC scores considering the base pipelines only. As a downside, this strict accuracy of the input limits the number of less probable correct matches which could be adjusted by a robust filtering approach, so that Superpoint matches filtered by MOP-based methods or FC-GNN provide overall remarkably better results.

Analogous observations hold for DISK, with the additional note that on IMC-PT the DISK base pipeline is much better than the corresponding base pipeline of SuperPoint, yet with FC-GNN or MOP-based filtering SuperPoint is able to almost fill the gap with DISK.

For DeDoDe v2 the results are quite unstable. Overall, this pipeline can only slight benefits by filtering and refinements, yet only using FC-GNN or MOP-based filters, but anyways the original or processed matches provide inferior results than those of SuperPoint.

\subsubsection{Recent dense pipelines}\label{dense}
Further analyses were done using the recent dense pipelines RoMa and MASt3R. While these pipelines provide SOTA accuracy with respect to the previous generation of the sparse approaches, these are currently less scalable and not suitable for general and common workflow tasks. The evaluations have been restricted to MegaDepth, ScanNet and the planar scenes, including MOP-based filtering, FC-GNN and Progressive-X. Detailed tables are reported in Appendix~\ref{dense_appendix}.

For RoMa, filtering does not alter the accuracy, while NCC refinement strongly decreases the performances, likewise ALIKED. RoMa accuracy is generally higher than that of sparse pipelines and the RoMa dense approach provides many accurate matches. Note that MOP-based filters in most cases remove less matches than others in non-planar scenes. As for ALIKED, the accuracy of matches is very high, but since RoMa is dense there would be many keypoints on flat areas. Both these factors negatively contribute to NCC suitability, leading to significatively degraded performances.

MASt3R performances are better than RoMa on ScanNet and close to those of SuperPoint with filtering for MegaDepth. In any case, MOP-based filtering without NCC does not offer relevant benefits on these datasets. For the planar dataset, MASt3R performs very well with MOP-based filtering only with NCC, with an accuracy level comparable with that of RoMa and SuperPoint. Nevertheless, on non-planar scenes NCC strongly reduces the accuracy while the precision, and especially the recall, are oddly boosted. As presented in the next section through visual inspection, the accuracy of MASt3R matches is generally low due to its regular grid sampling strategy, but these inaccurate matches are consistent with the scene epipolar geometry. The MASt3R approach, inherited from DUSt3R, obtains the 3D structure of the scene for each image, aligns these 3D structures and then computes the outputs matches, which will be consistent with the 3D alignment and so will be the derived epipolar geometry. Please refer to the next visual analysis.

\begin{figure}
	\centering
	\resizebox{!}{0.5\textwidth}{	
	\begin{tabular}{r@{\hskip 0.3em}l@{\hskip 0.3em}l@{\hskip 0.3em}c}
		\midrule
		\multirow{-3}{*}{\footnotesize SIFT+NNR} & \includegraphics[height=0.075\textwidth]{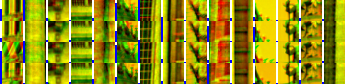} & \includegraphics[height=0.075\textwidth]{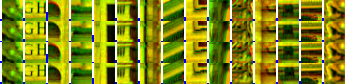} & \multirow{9}{*}{\includegraphics[height=0.09\textwidth, angle=270]{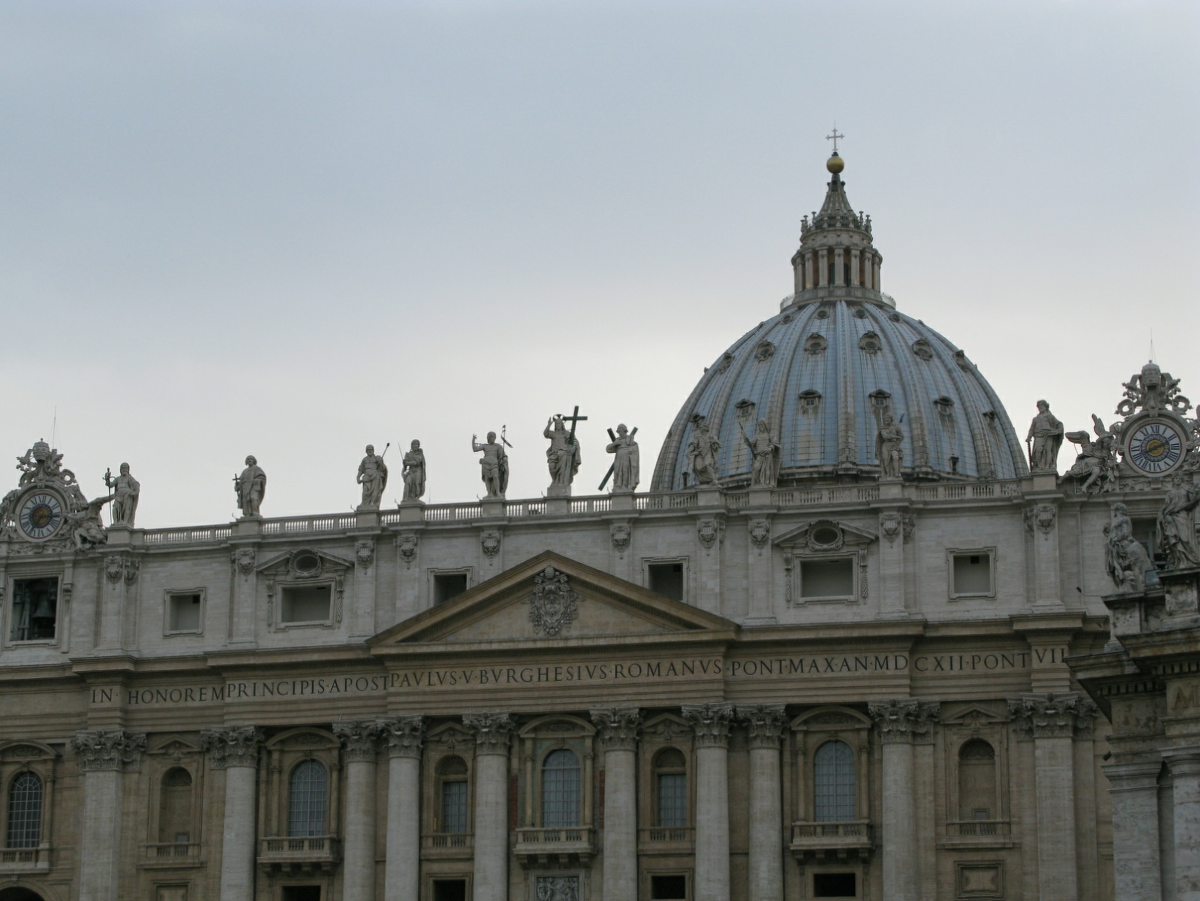}} \\ 		
		\multirow{-3}{*}{\footnotesize Key.Net+$\protect\substack{\text{AffNet}\\\text{HardNet}}$+NNR} &
		\includegraphics[height=0.075\textwidth]{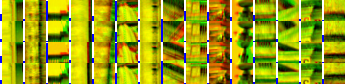} & \includegraphics[height=0.075\textwidth]{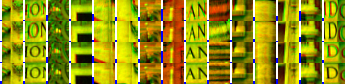} \\ 		
		\multirow{-3}{*}{\footnotesize SuperPoint+LightGlue} & \includegraphics[height=0.075\textwidth]{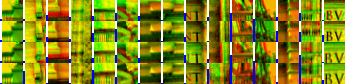} & \includegraphics[height=0.075\textwidth]{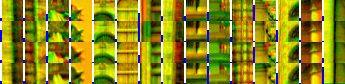} \\ 		
		\multirow{-3}{*}{\footnotesize ALIKED+LightGlue} & \includegraphics[height=0.075\textwidth]{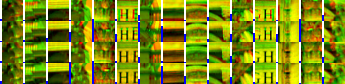} & \includegraphics[height=0.075\textwidth]{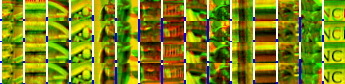} & \multirow{5.5}{*}{\includegraphics[height=0.09\textwidth, angle=270]{m281_im2.png}} \\ 		
		\multirow{-3}{*}{\footnotesize DISK+LightGlue} & \includegraphics[height=0.075\textwidth]{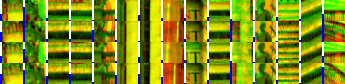} & \includegraphics[height=0.075\textwidth]{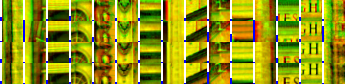} \\ 		
		\multirow{-3}{*}{\footnotesize LoFTR} & \includegraphics[height=0.075\textwidth]{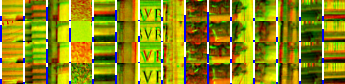} & \includegraphics[height=0.075\textwidth]{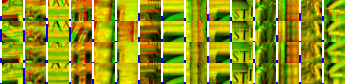} \\ 		
		\multirow{-3}{*}{\footnotesize DeDoDe v2} & \includegraphics[height=0.075\textwidth]{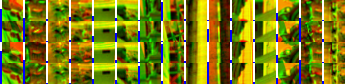} & \includegraphics[height=0.075\textwidth]{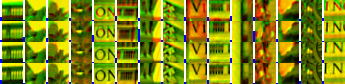} \\ 
		\midrule	
		\multirow{-3}{*}{\footnotesize SIFT+NNR} & \includegraphics[height=0.075\textwidth]{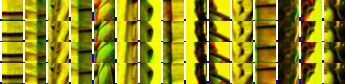} & \includegraphics[height=0.075\textwidth]{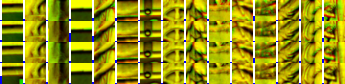} & \multirow{9}{*}{\includegraphics[height=0.09\textwidth, angle=270]{m1489_im1.png}} \\ 		
		\multirow{-3}{*}{\footnotesize Key.Net+$\protect\substack{\text{AffNet}\\\text{HardNet}}$+NNR} & \includegraphics[height=0.075\textwidth]{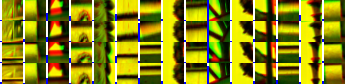} & \includegraphics[height=0.075\textwidth]{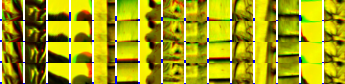} \\ 		
		\multirow{-3}{*}{\footnotesize SuperPoint+LightGlue} & \includegraphics[height=0.075\textwidth]{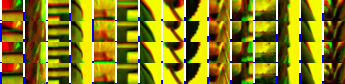} & \includegraphics[height=0.075\textwidth]{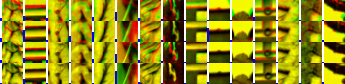} \\ 		
		\multirow{-3}{*}{\footnotesize ALIKED+LightGlue} & \includegraphics[height=0.075\textwidth]{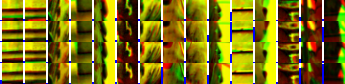} & \includegraphics[height=0.075\textwidth]{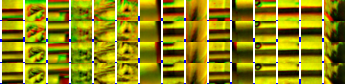} & \multirow{5.5}{*}{\includegraphics[height=0.09\textwidth, angle=270]{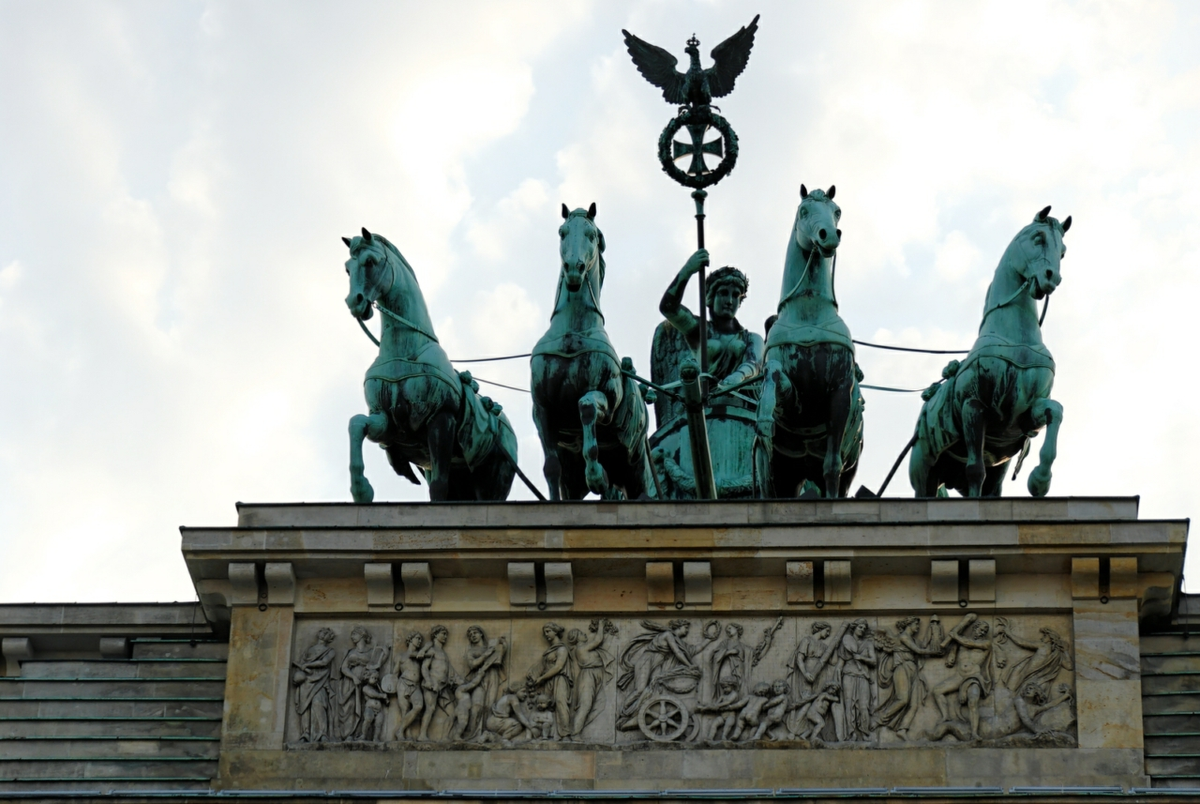}} \\ 		
		\multirow{-3}{*}{\footnotesize DISK+LightGlue} & \includegraphics[height=0.075\textwidth]{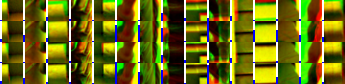} & \includegraphics[height=0.075\textwidth]{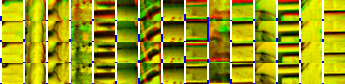} \\ 		
		\multirow{-3}{*}{\footnotesize LoFTR} & \includegraphics[height=0.075\textwidth]{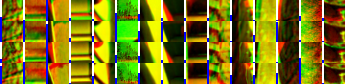} & \includegraphics[height=0.075\textwidth]{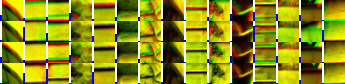} \\ 		
		\multirow{-3}{*}{\footnotesize DeDoDe v2} & \includegraphics[height=0.075\textwidth]{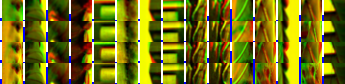} & \includegraphics[height=0.075\textwidth]{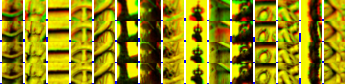} \\ 
		\midrule
	\end{tabular}
	}	
	\caption{Overlapped patches as red-green stereographs for 30 random matches for the MegaDepth image pairs shown in the rightmost column of the table. For each base pipeline, the original patches without NCC, with NCC, MOP+MiHo, and MOP+MiHo+NCC are shown from the top to the bottom rows, respectively. The left and right patch clusters correspond to the case where NCC correlation is higher without and with MOP+MiHo, respectively. The blue bar on the left of each patch indicates the maximum epipolar error with respect to the GT up to 10 px, i.e. the patch radius. Please refer to Sec.~\ref{patch_showcase_discussion}. Best viewed in color and zoomed in.}\label{patches}
\end{figure}
  
\subsubsection{Visual qualitative analysis}\label{patch_showcase_discussion}
Figure~\ref{patches} shows visual results of the patch alignment by NCC for two example image pairs, superimposed as red-green anaglyphs. Patches have been randomly chosen for visualization, and the full patch sets for each evaluated pipeline and dataset combination can be generated by the supporting tool included in the code. Inside each multirow in the figure, the top row shows the initial patch alignment specified by the homography pair $(\mathrm{H}^b_1,\mathrm{H}^b_2)$ defined in Sec.~\ref{ncc}, while the second row shows the results after applying NCC without MOP-based homography solutions. Likewise, the third row shows the initial alignment of the MOP+MiHo extended homography pair $(\mathrm{H}_{m_1},\mathrm{H}_{m_2})$ and the bottom row the final results with NCC. In the left clusters $(\mathrm{H}^b_1,\mathrm{H}^b_2)$, with or without NCC, has the highest correlation value. The blue bar on the left of each patch superimposition indicates the corresponding epipolar error of the patch center with respect to the pseudo GT. The reader is invited to zoom in and inspect the above figure, which provides examples of the possible configurations and alignment failure cases.

For true planar regions, such as the patches including the St. Peter's Basilica epigraph, it is evident that MOP+MiHo extended patches are better than the base ones in terms of alignment and distortions errors with respect to the original images. Nevertheless, MOP patch alignment can go wrong in some circumstances as shown in the leftmost column of the top multirow. This happens for instance when the homography assignment described in Sec.~\ref{assign_homo} picks a homography with a small reprojection error but also a poor wrong support set. Furthermore, NCC similarity maximization is not always the best choice as highlighted by the patches containing the inscription in the left clusters, where the registration is visibly better with MOP although base patches achieve the highest NCC correlation value.

In the case of patches representing almost flat regions or straight contour edges according to the Harris corner definition~(\cite{harris}), NCC can be misled and can degrade the alignment increasing the epipolar error, as in the rightmost patch overlay in the third multirow. Notice also that although visually improving the overall patch alignment, NCC can increase the epipolar error of the effective keypoint center, as for the `BV' inscription in the third multirow. Anyways, when NCC keypoint accuracy decreases assuming that the pseudo-GT sub-pixel estimation can be always trusted, the epipolar error increment is generally limited as reported in the figure and by the results of Tables~\ref{sift_res}-\ref{aliked_res}.

\begin{figure}
	\centering
	\begin{tabular}{c@{\hskip 0.6em}c@{\hskip 0.6em}c@{\hskip 0.6em}c@{\hskip 0.6em}c}
		& & \resizebox{!}{0.54em}{+MOP+MiHo+NCC\hphantom{g}} & \resizebox{!}{0.54em}{+FC-GNN\hphantom{j}} \\
		\midrule
		& \includegraphics[width=0.27\textwidth]{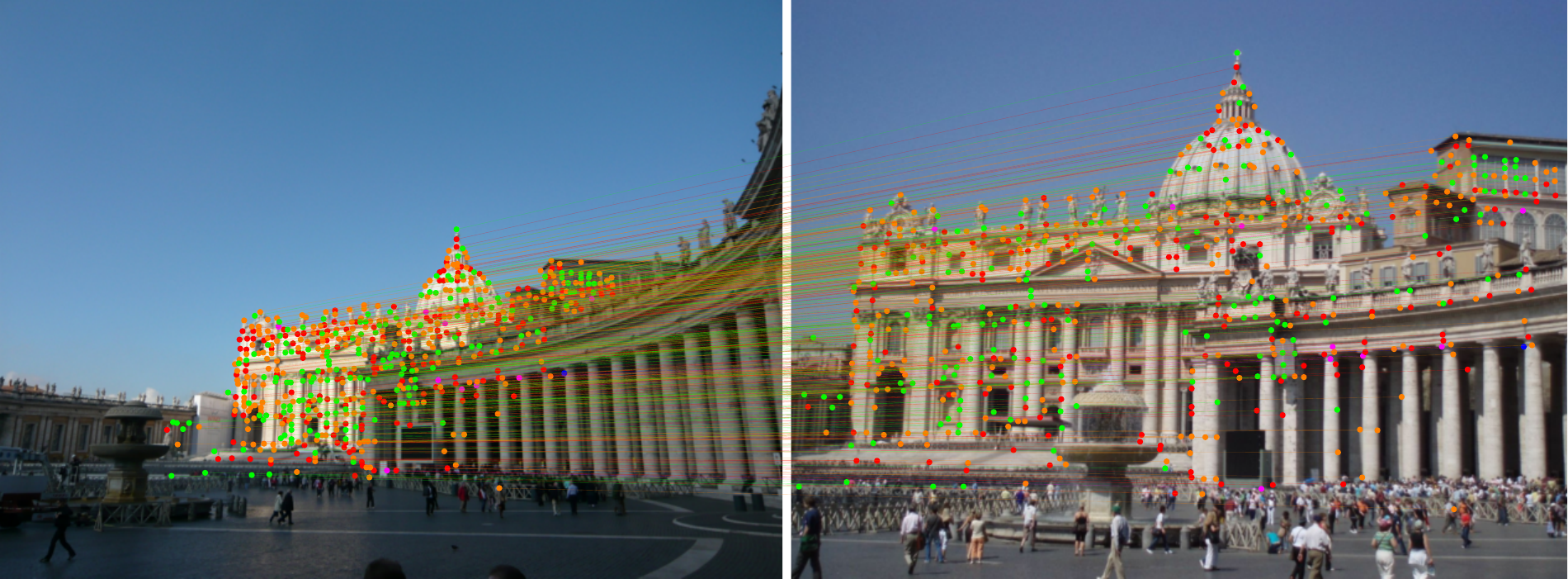} & \includegraphics[width=0.27\textwidth]{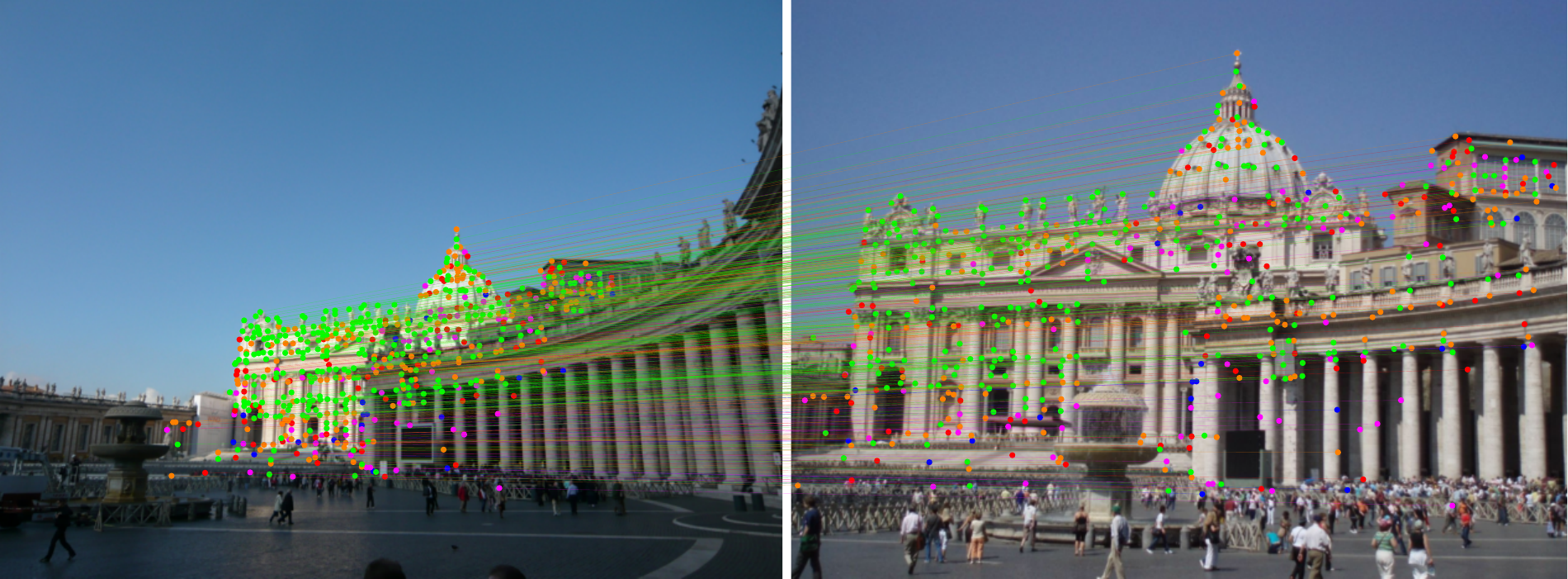} & \includegraphics[width=0.27\textwidth]{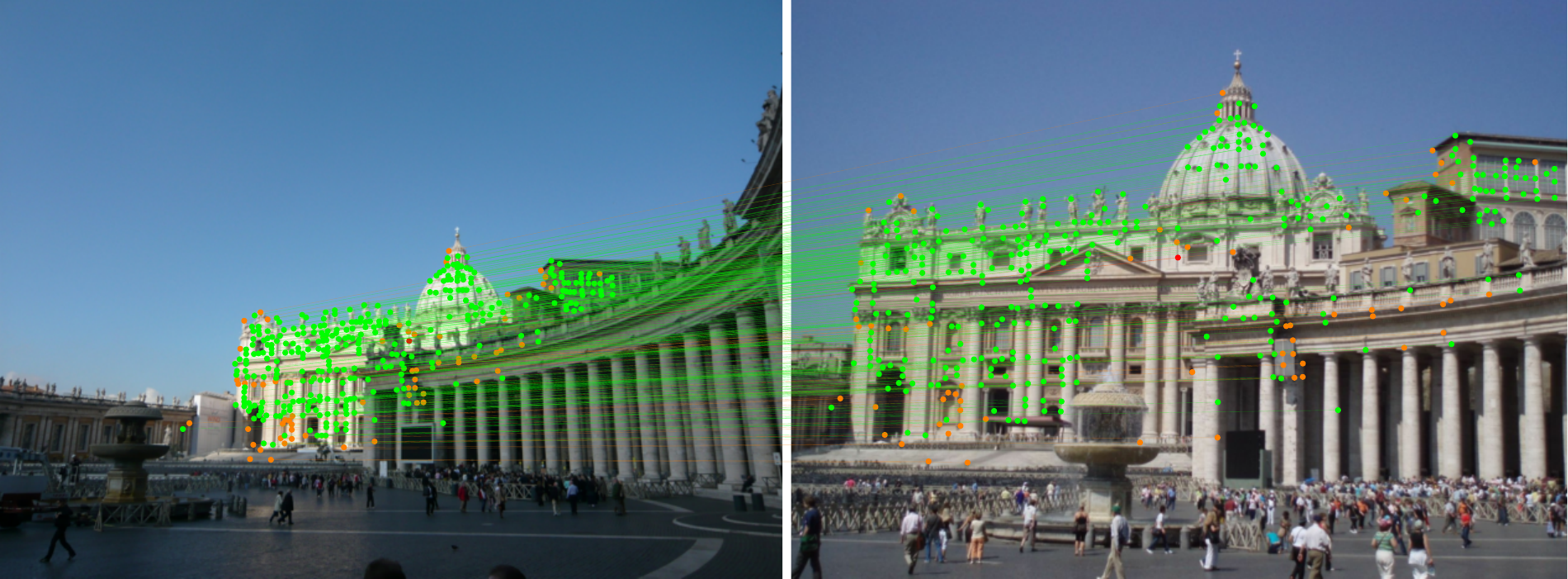} \\ 
		\multirow{-8}{1em}{\rotatebox{90}{\resizebox{!}{0.57em}{SuperPoint+LightGlue}}} & \includegraphics[width=0.27\textwidth]{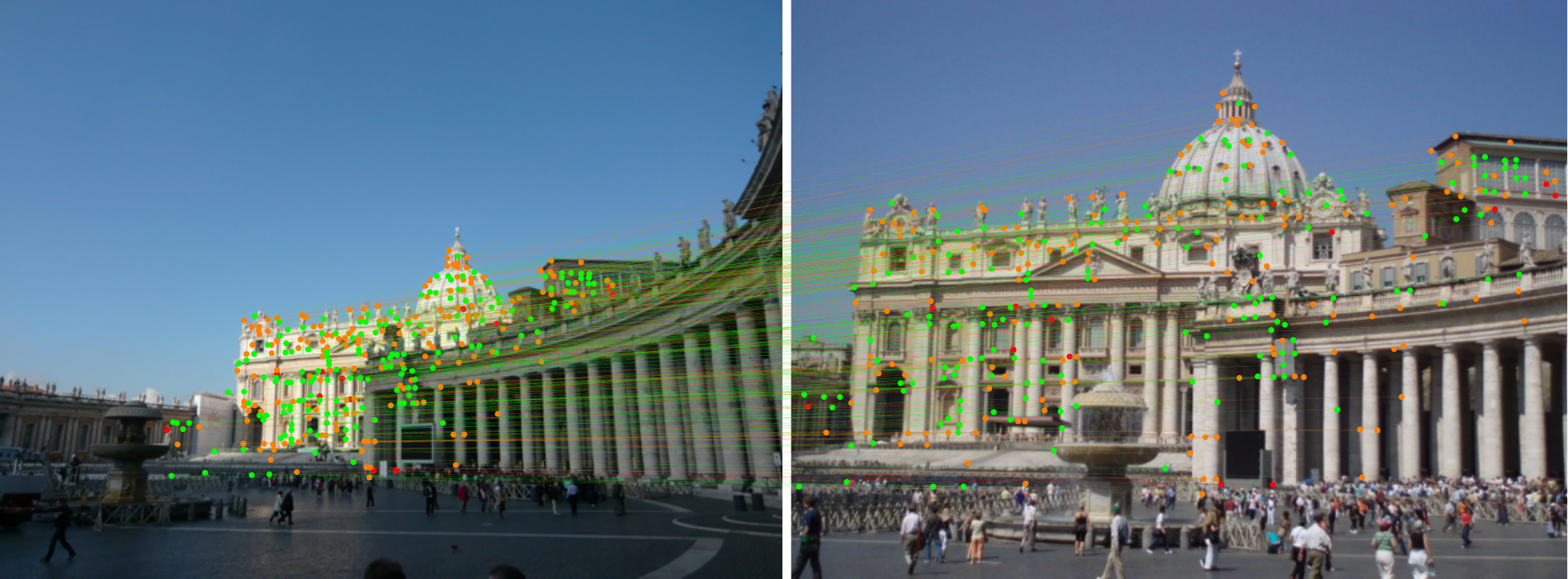} & \includegraphics[width=0.27\textwidth]{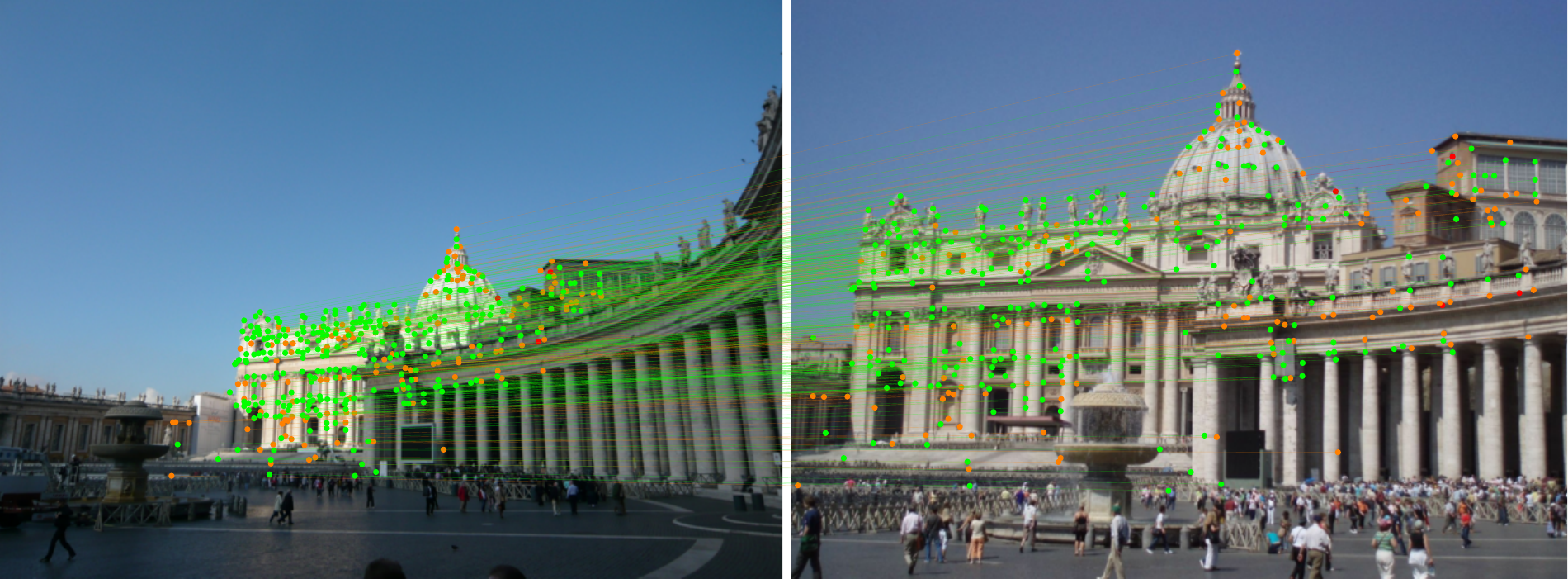} & \includegraphics[width=0.27\textwidth]{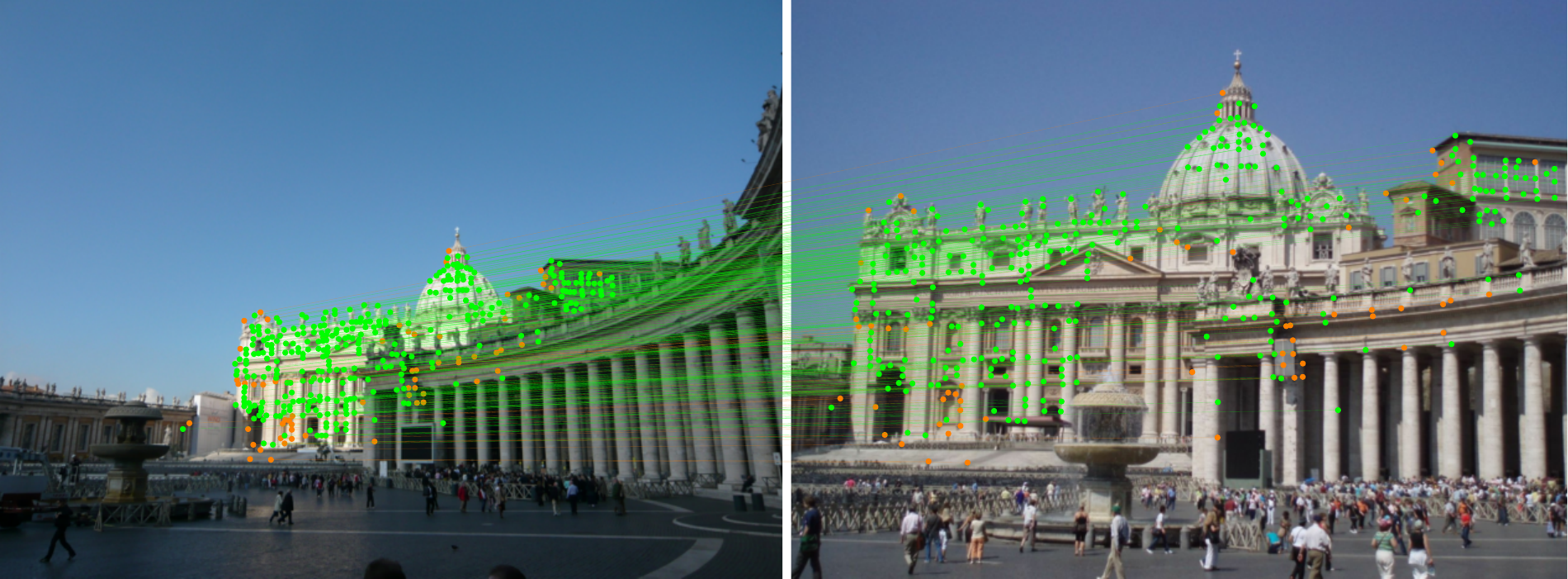} & \multirow{-5.5}{1em}{\rotatebox{-90}{\resizebox{!}{0.57em}{\hspace{0.5em}+MAGSAC}}} \\ 				
		\midrule
		& \includegraphics[width=0.27\textwidth]{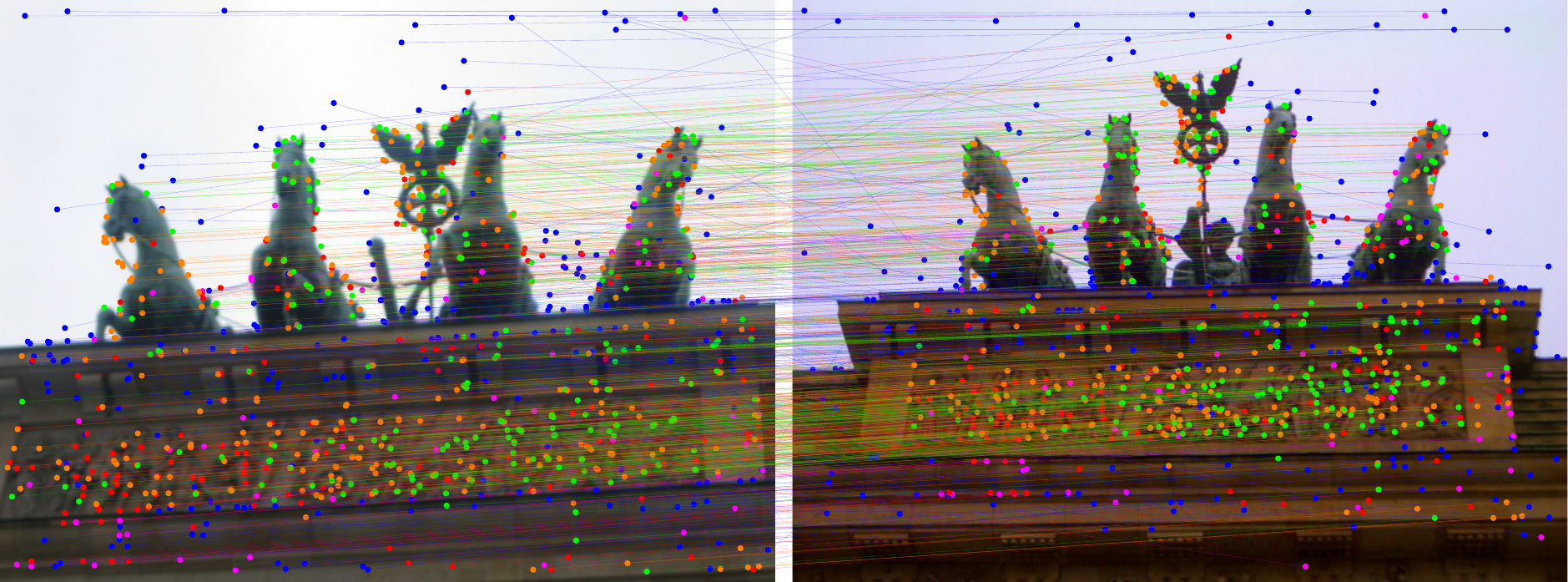} & \includegraphics[width=0.27\textwidth]{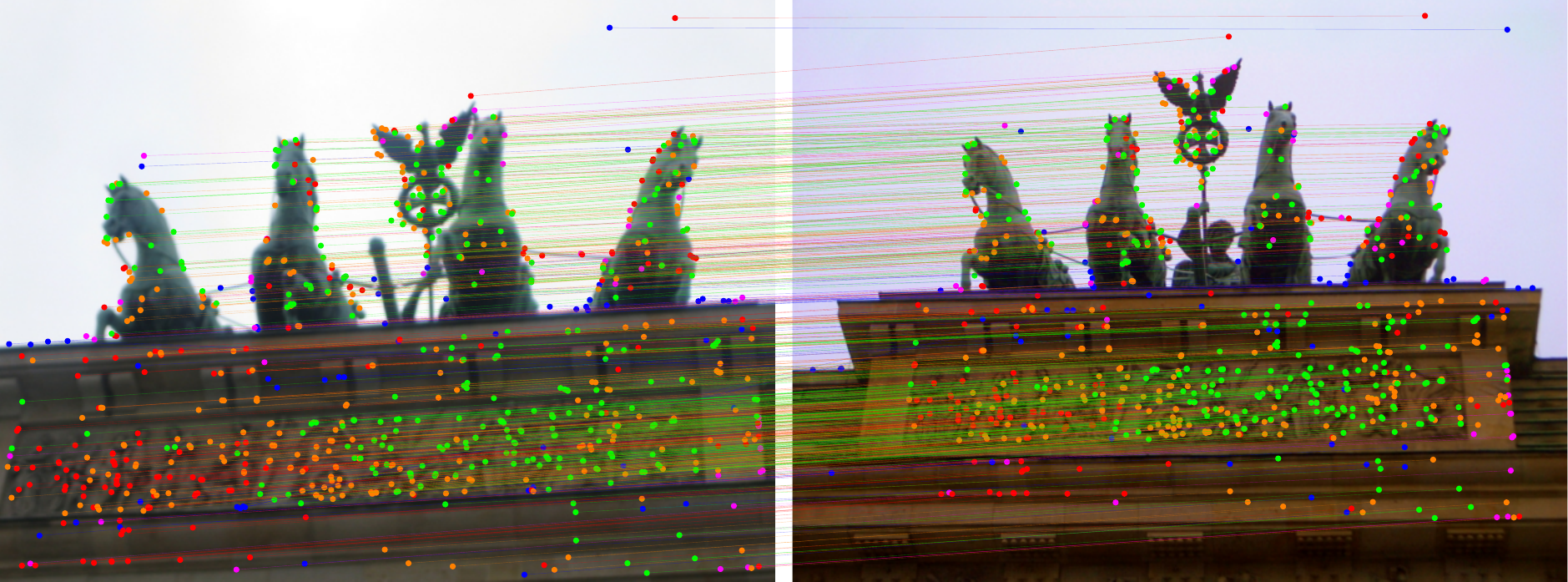} & \includegraphics[width=0.27\textwidth]{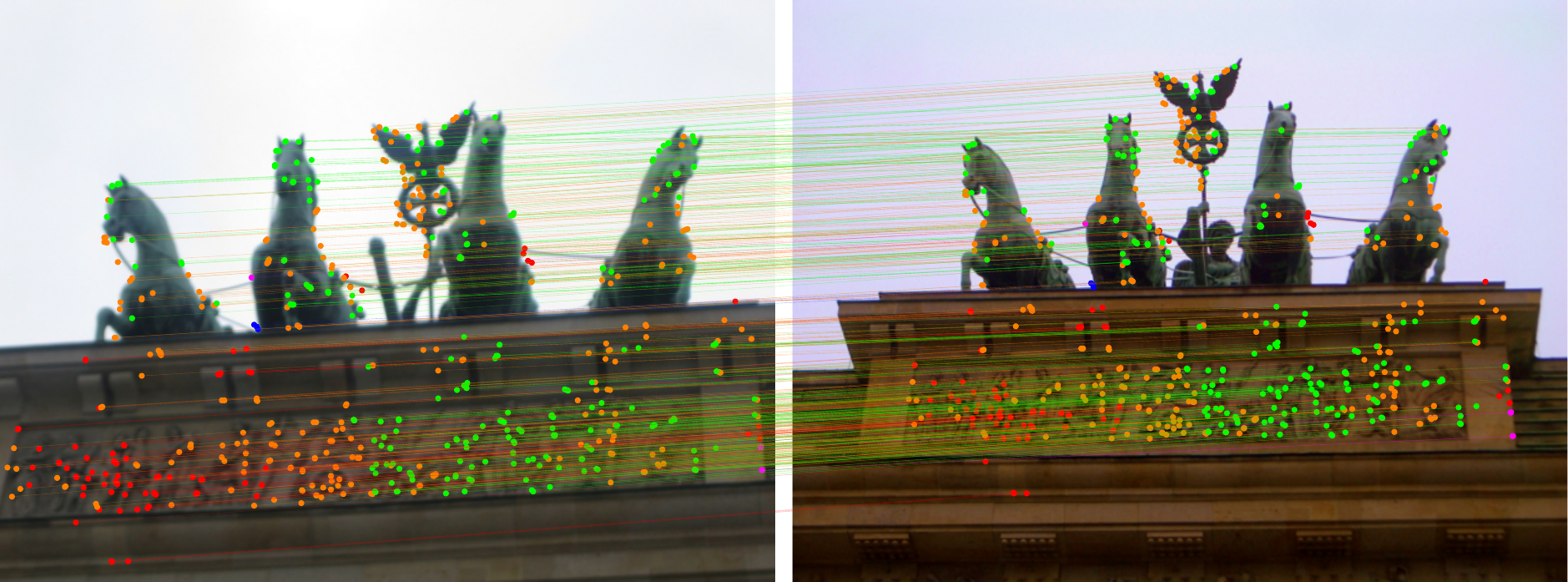} \\ 		
		\multirow{-8}{1em}{\rotatebox{90}{\resizebox{!}{0.63em}{Key.Net+$\substack{\text{AffNet}\\\text{HardNet}}$+NNR}}} & \includegraphics[width=0.27\textwidth]{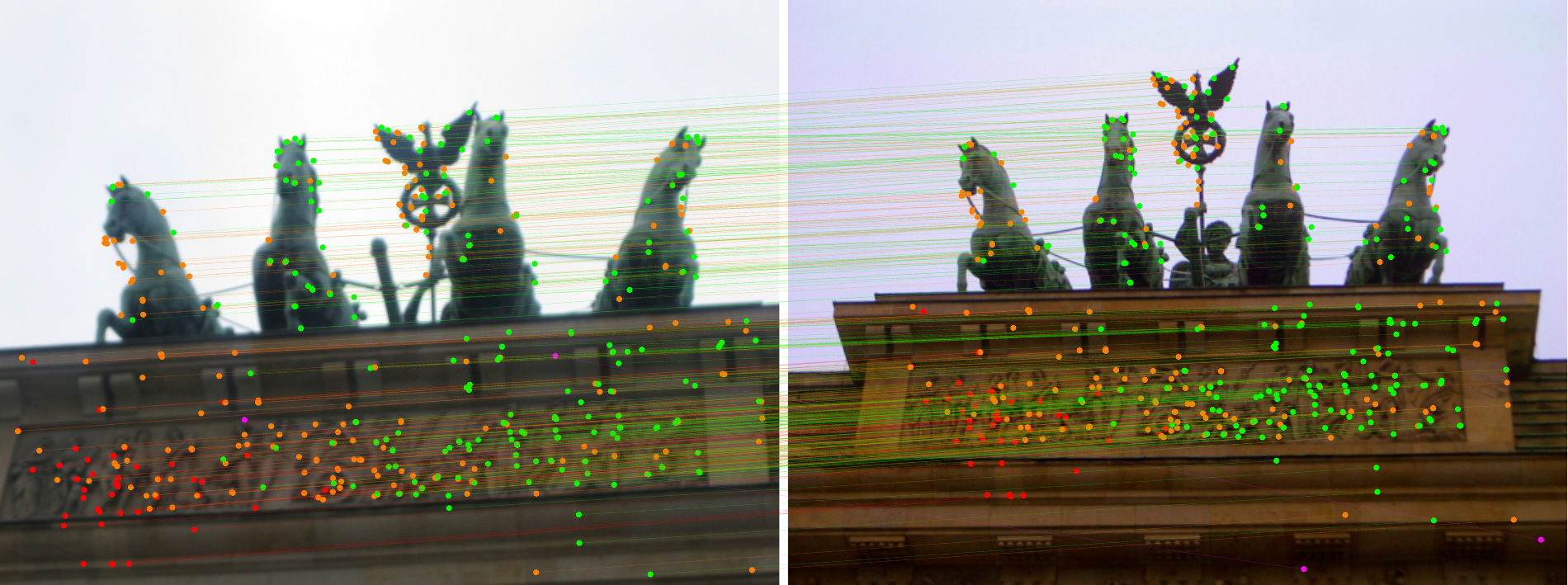} & \includegraphics[width=0.27\textwidth]{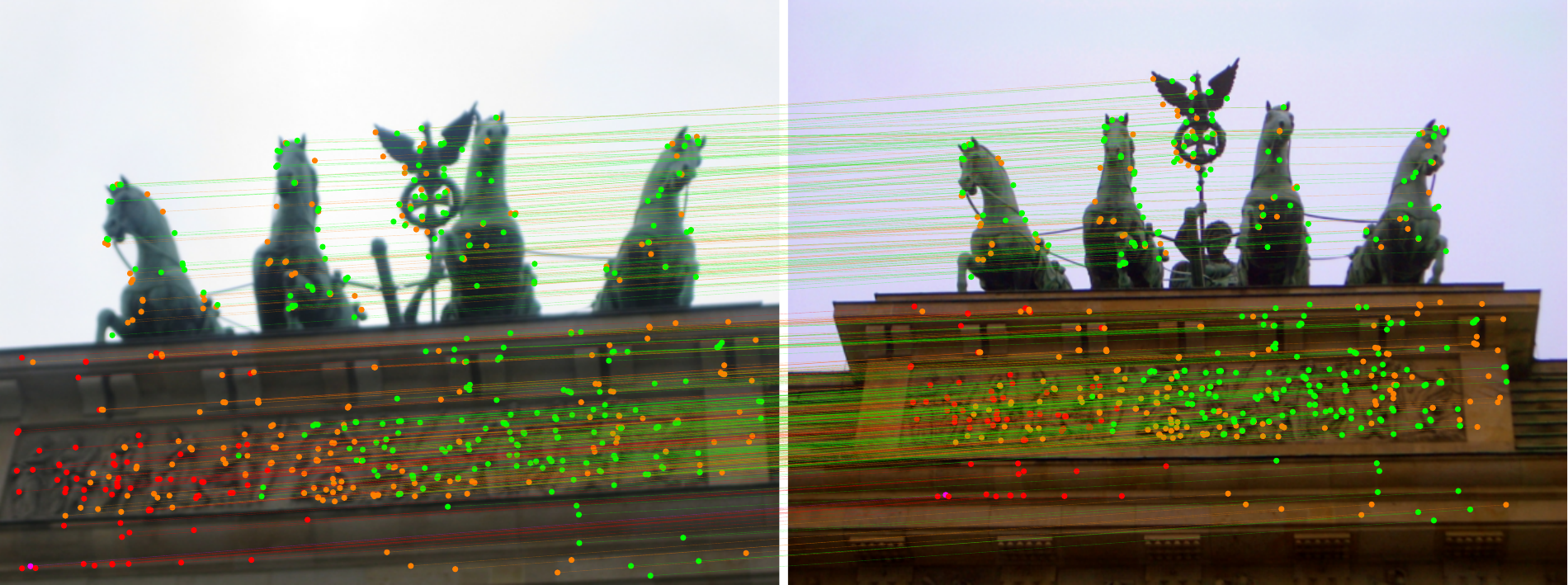} & \includegraphics[width=0.27\textwidth]{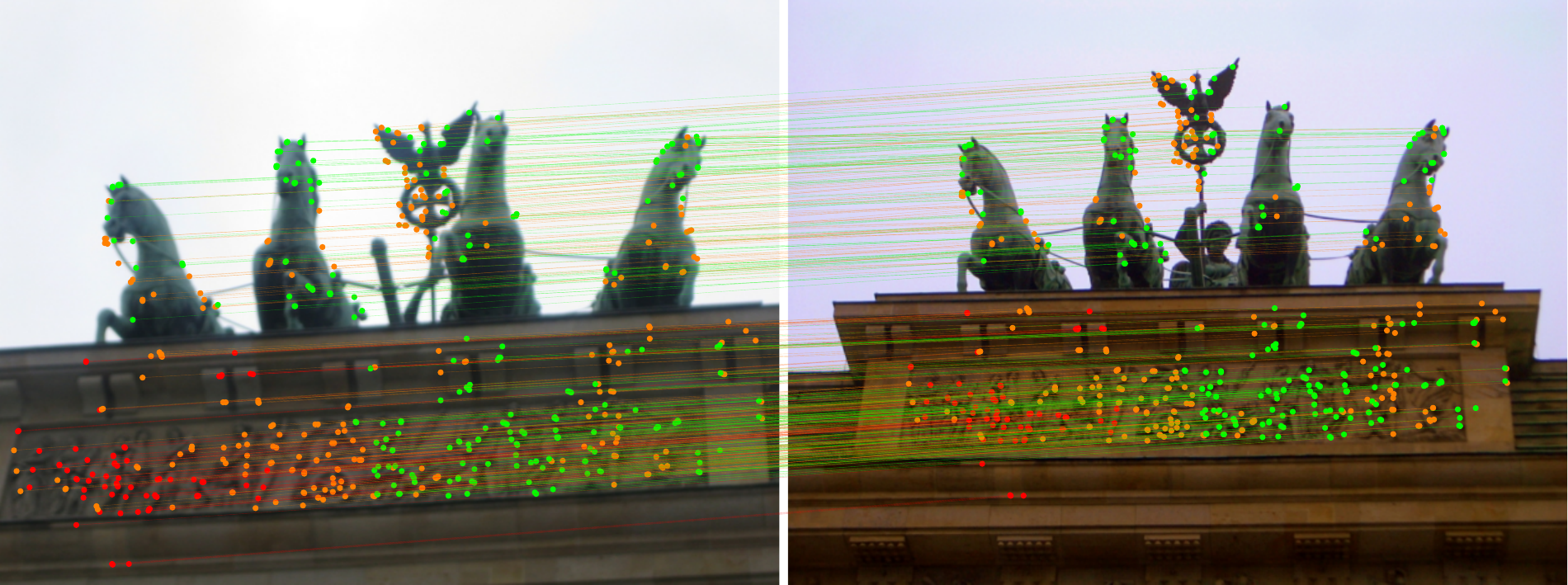} & \multirow{-5.5}{1em}{\rotatebox{-90}{\resizebox{!}{0.57em}{\hspace{0.5em}+MAGSAC}}} \\ 		
		\midrule
		& \includegraphics[width=0.27\textwidth]{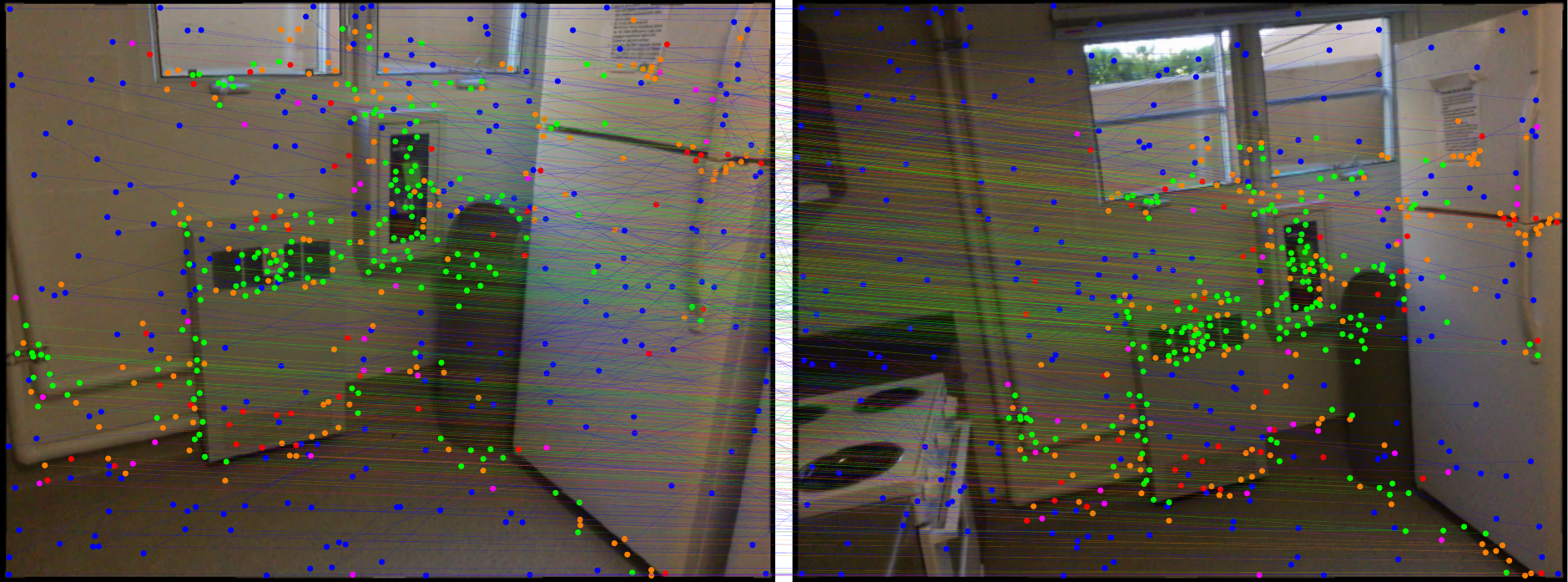} & \includegraphics[width=0.27\textwidth]{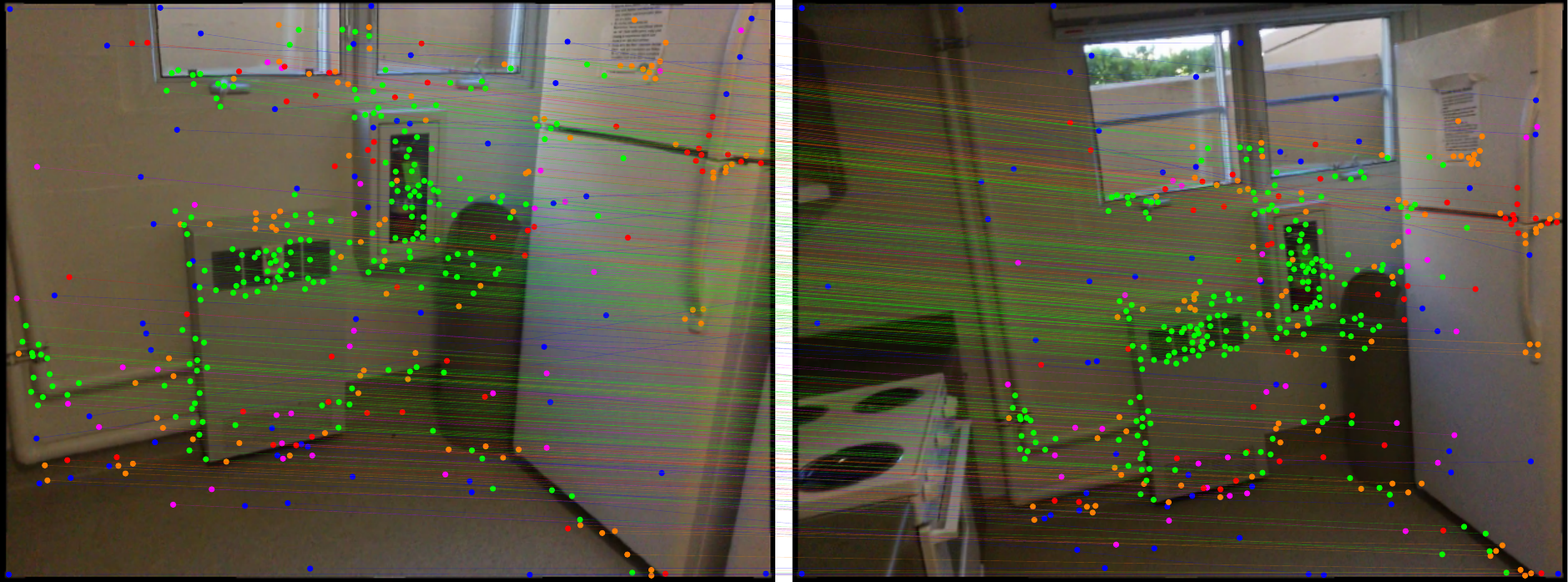} & \includegraphics[width=0.27\textwidth]{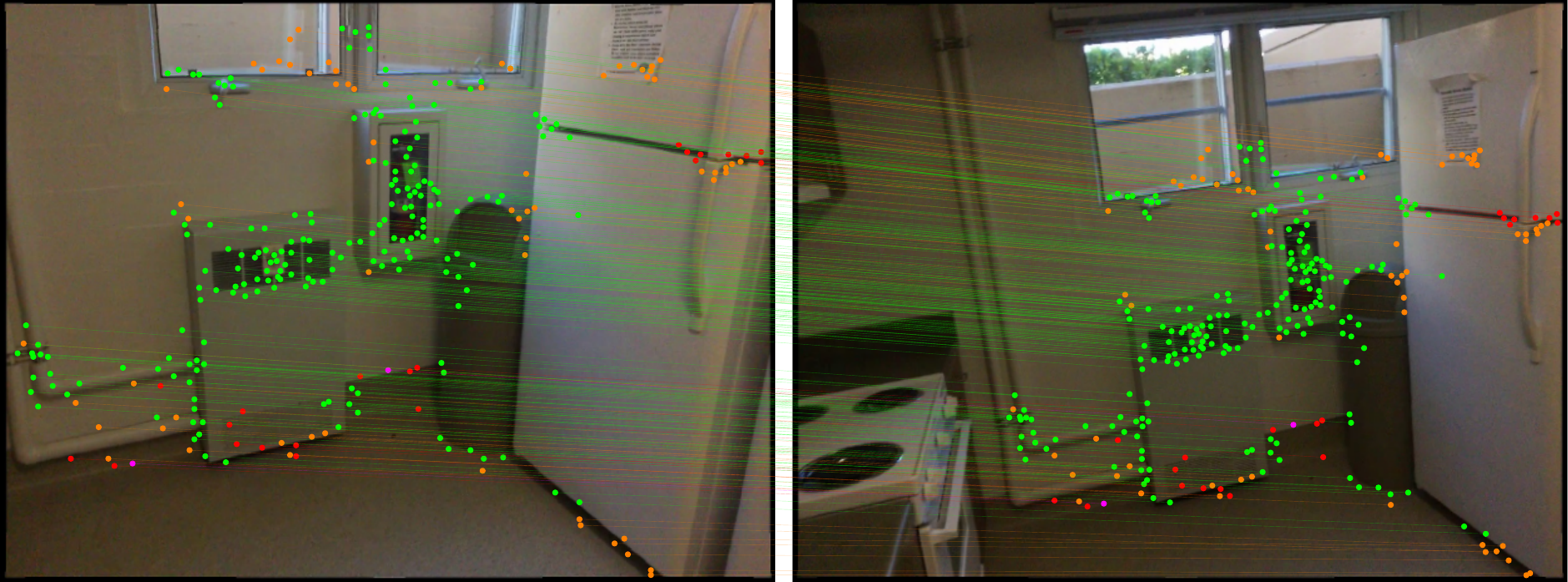} \\ 		
		\multirow{-8}{1em}{\rotatebox{90}{\resizebox{!}{0.57em}{SIFT+NNR\hphantom{g}}}} & \includegraphics[width=0.27\textwidth]{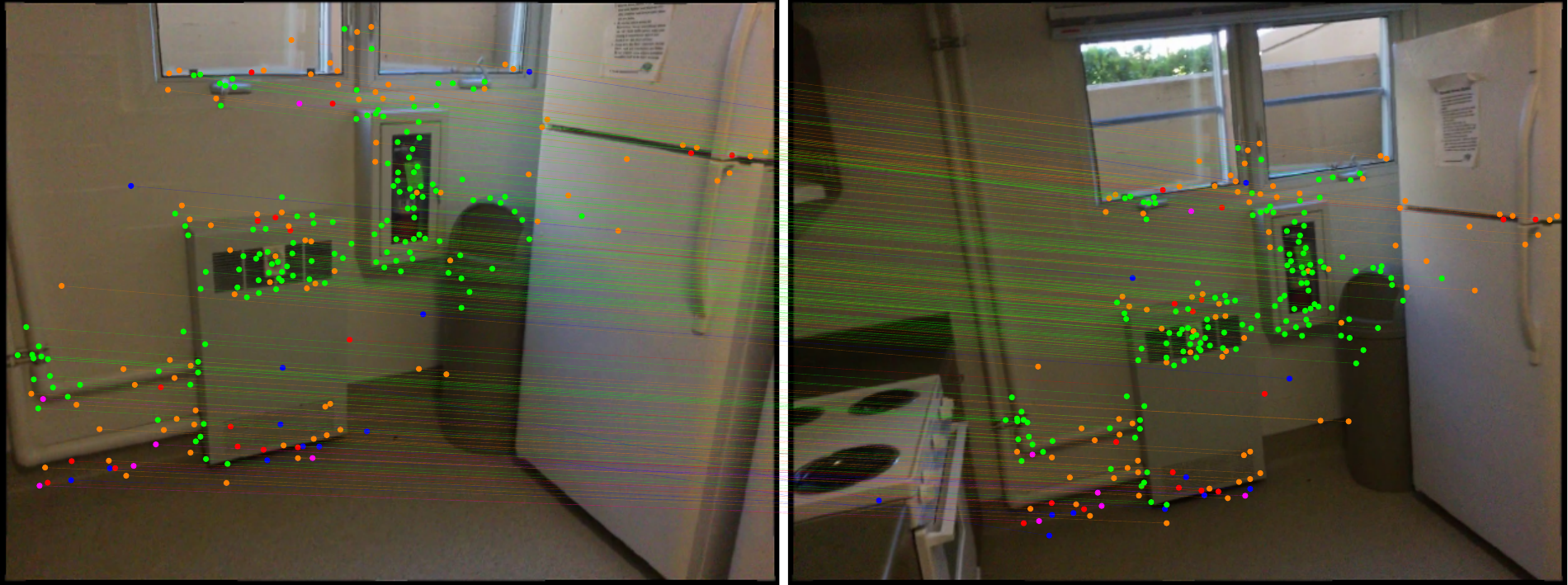} & \includegraphics[width=0.27\textwidth]{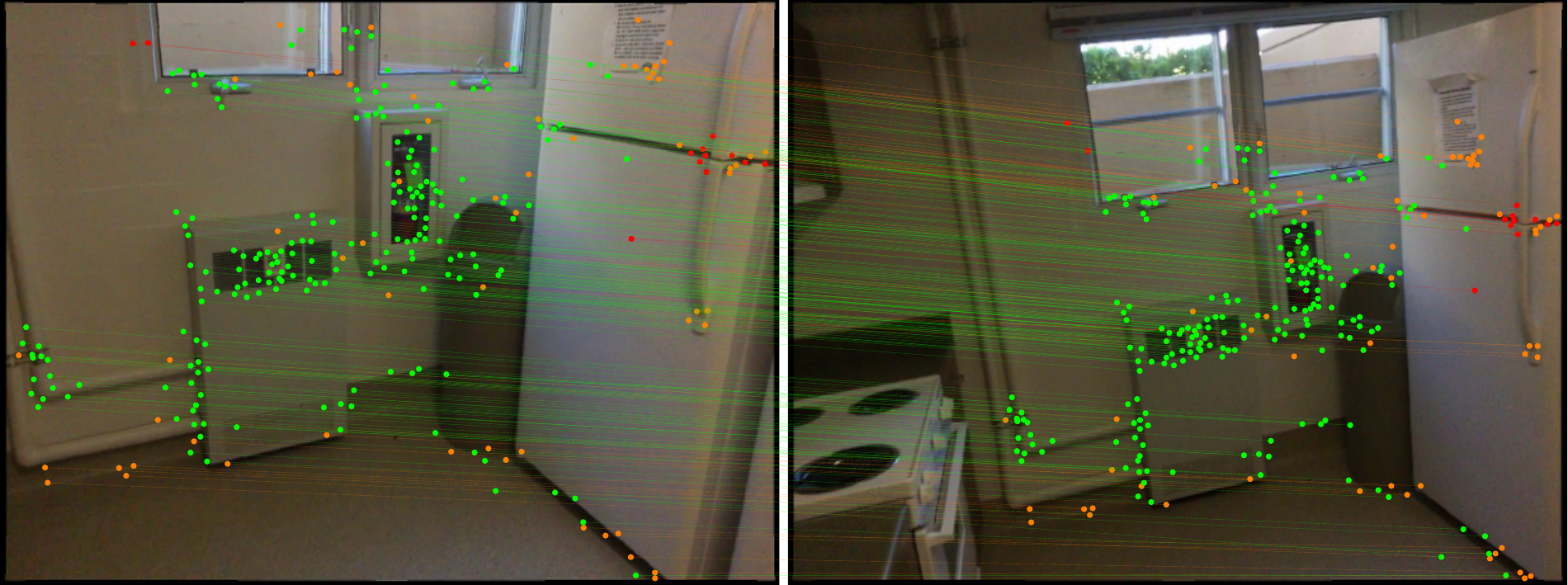} & \includegraphics[width=0.27\textwidth]{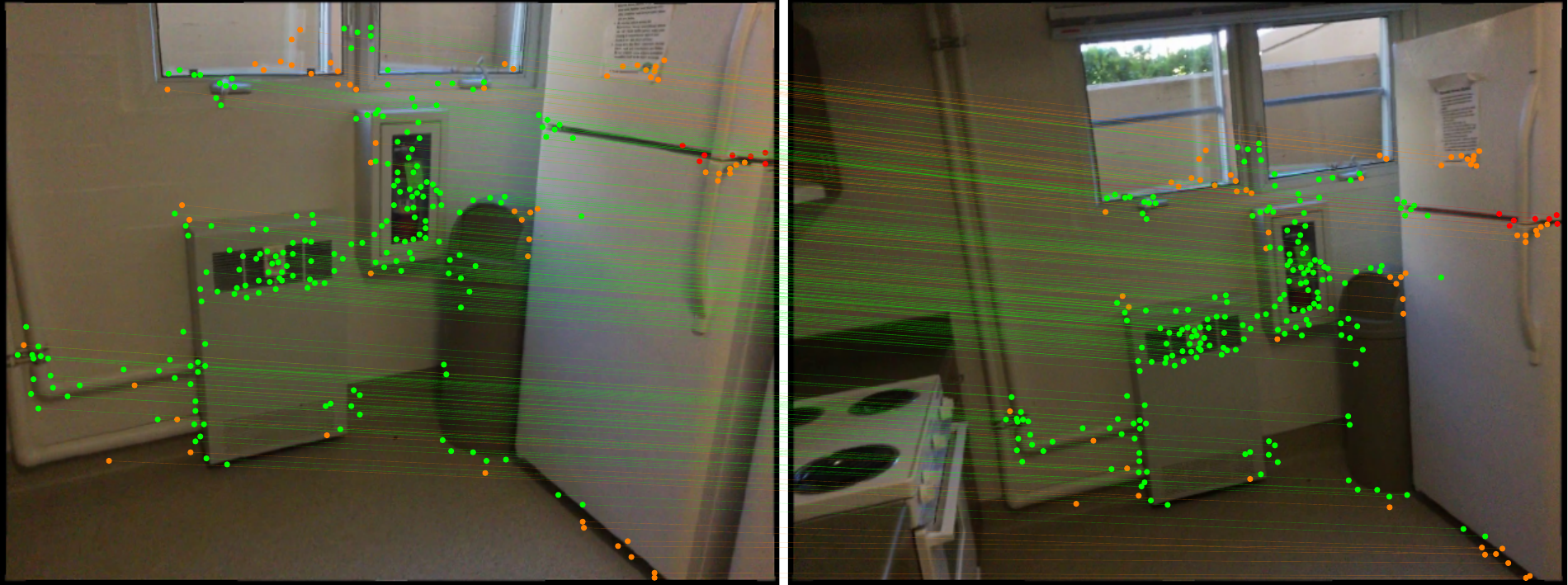} & \multirow{-5.5}{1em}{\rotatebox{-90}{\resizebox{!}{0.57em}{\hspace{0.5em}+MAGSAC}}} \\ 
		\midrule
		& \includegraphics[width=0.27\textwidth]{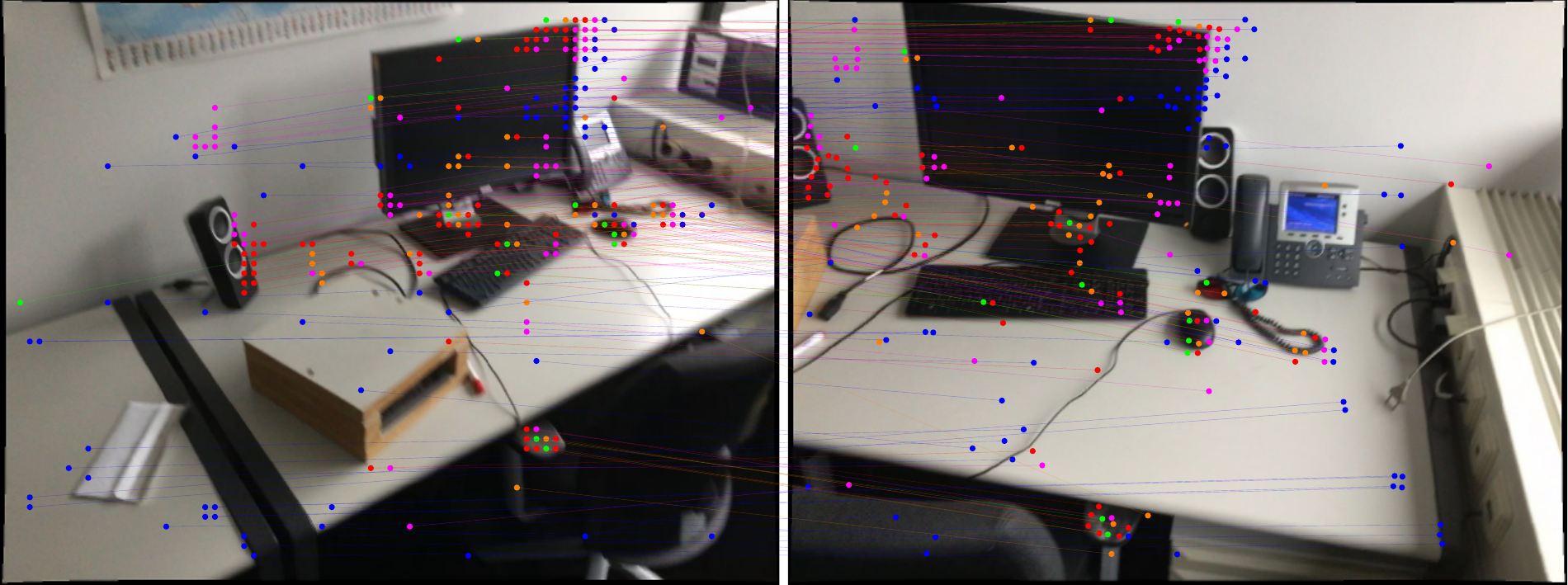} & \includegraphics[width=0.27\textwidth]{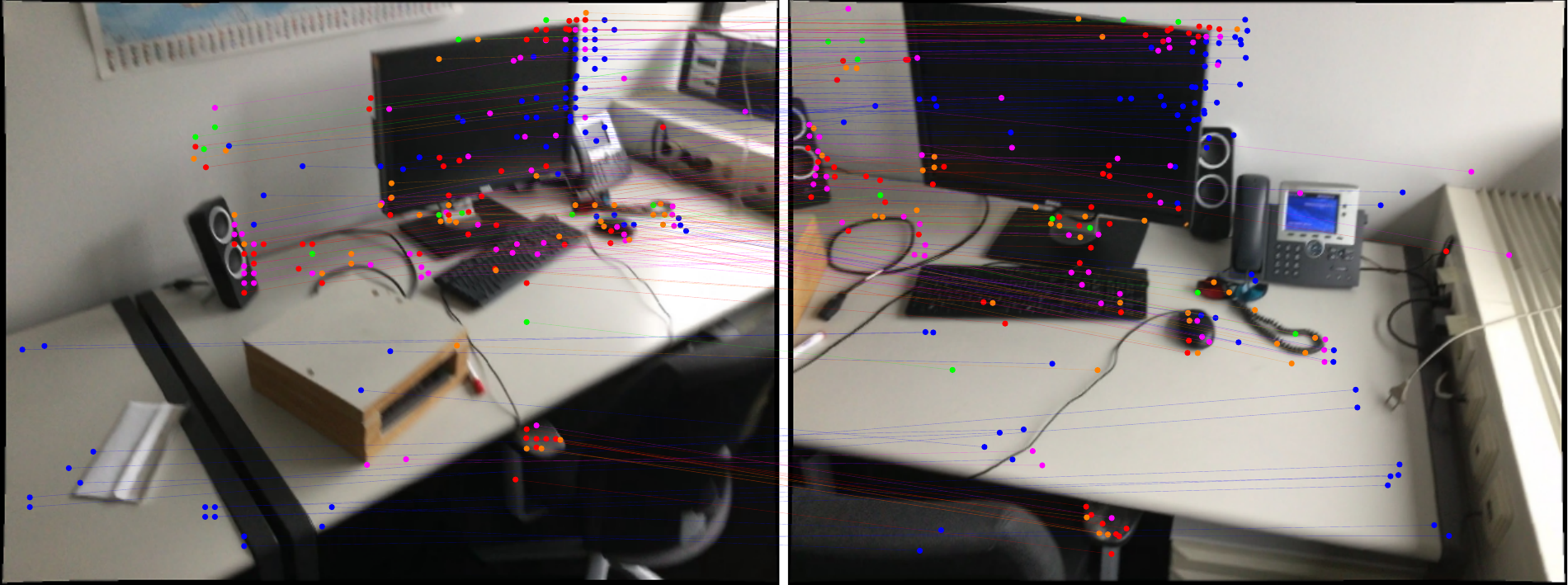} & \includegraphics[width=0.27\textwidth]{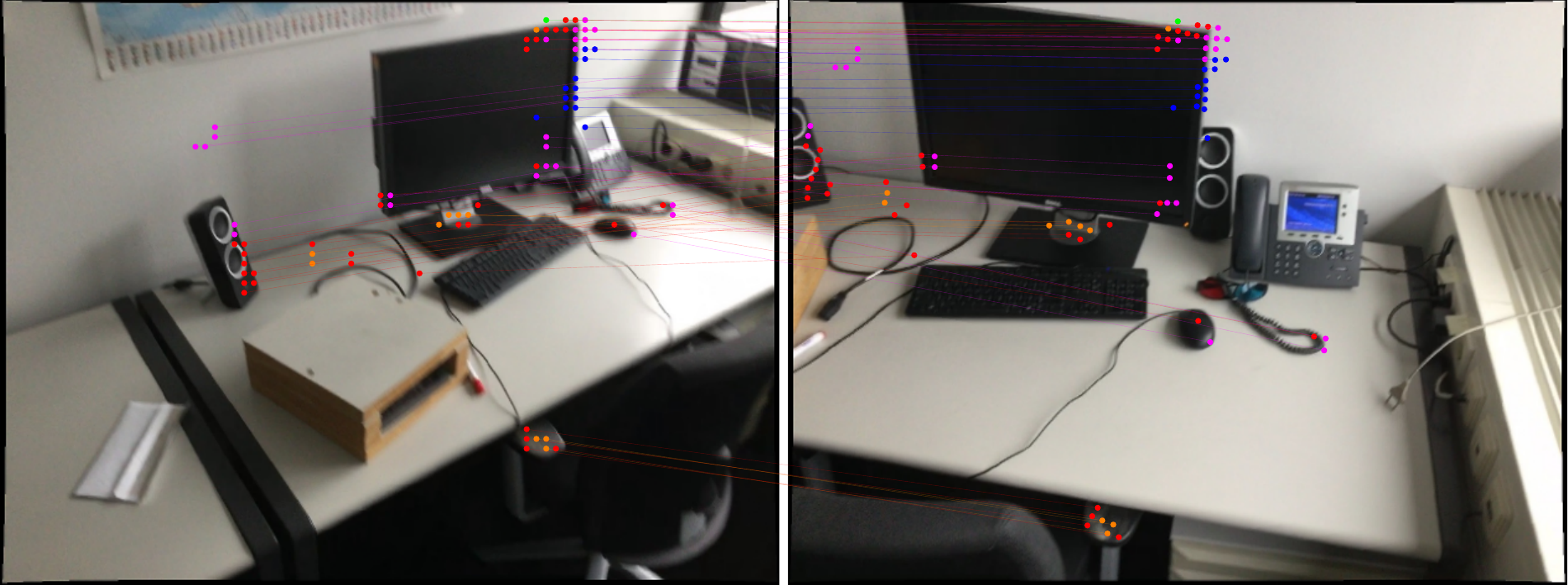} \\ 
		\multirow{-8}{1em}{\rotatebox{90}{\resizebox{!}{0.57em}{LoFTR}}} & \includegraphics[width=0.27\textwidth]{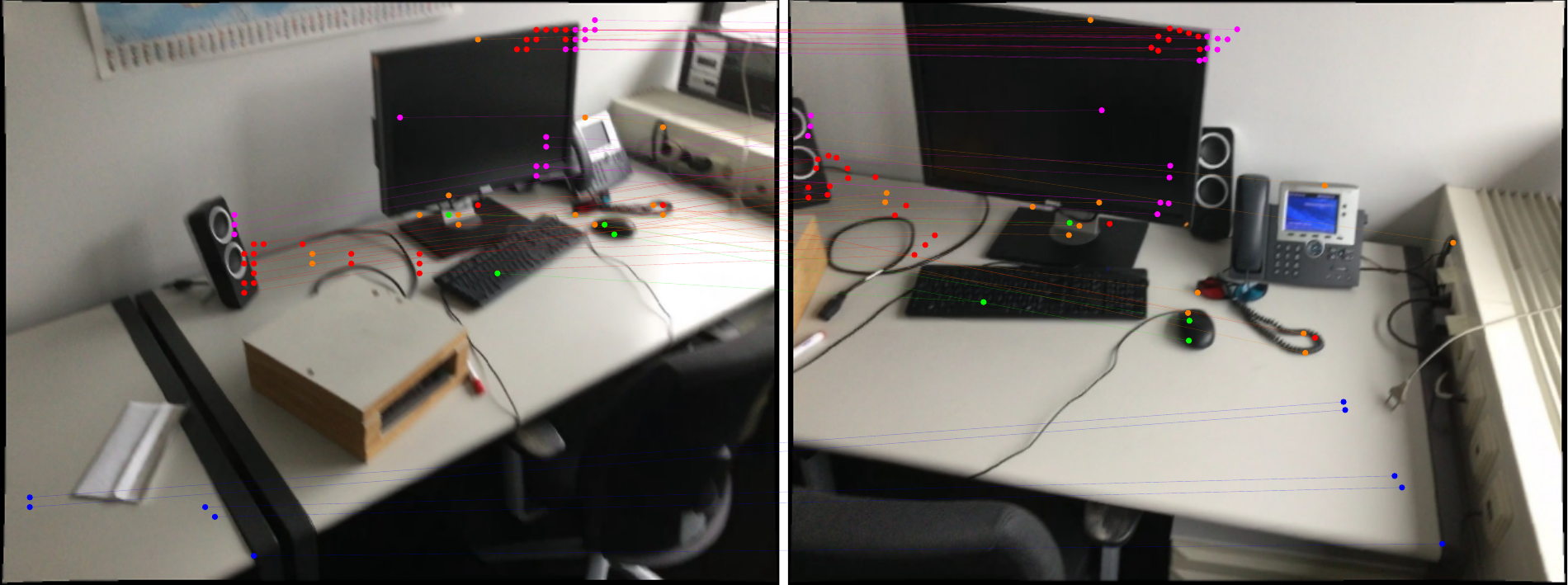} & \includegraphics[width=0.27\textwidth]{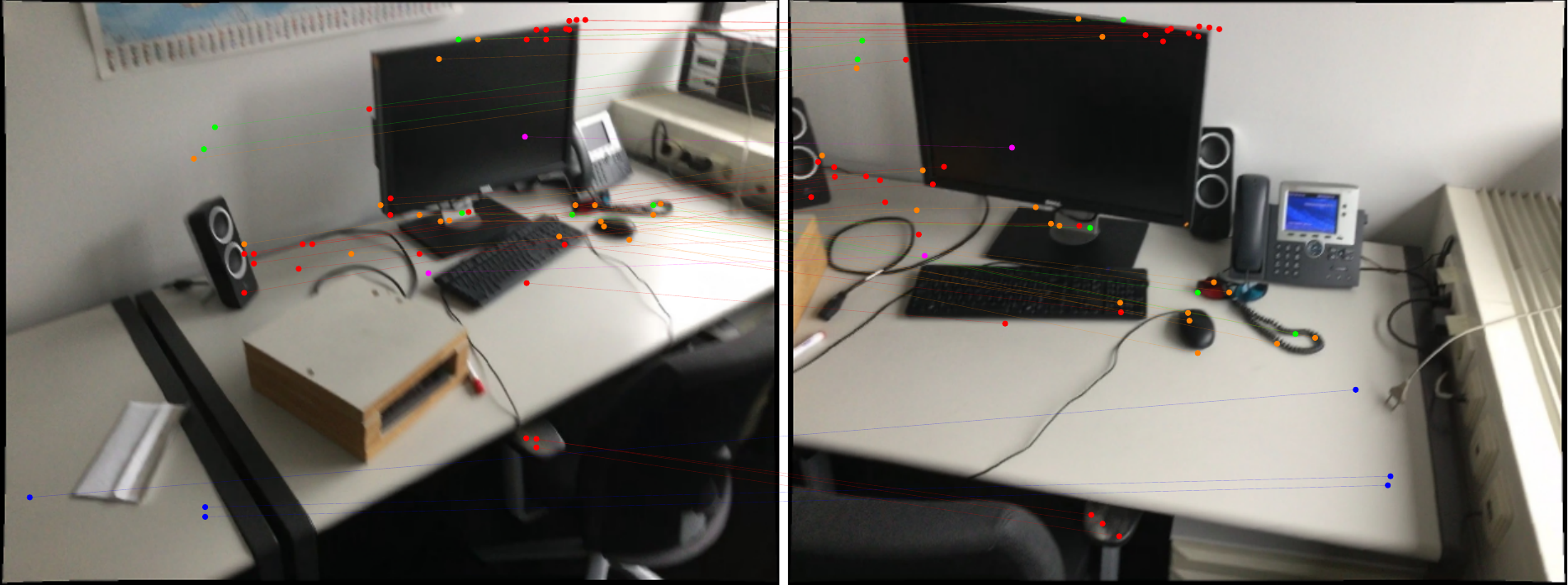} & \includegraphics[width=0.27\textwidth]{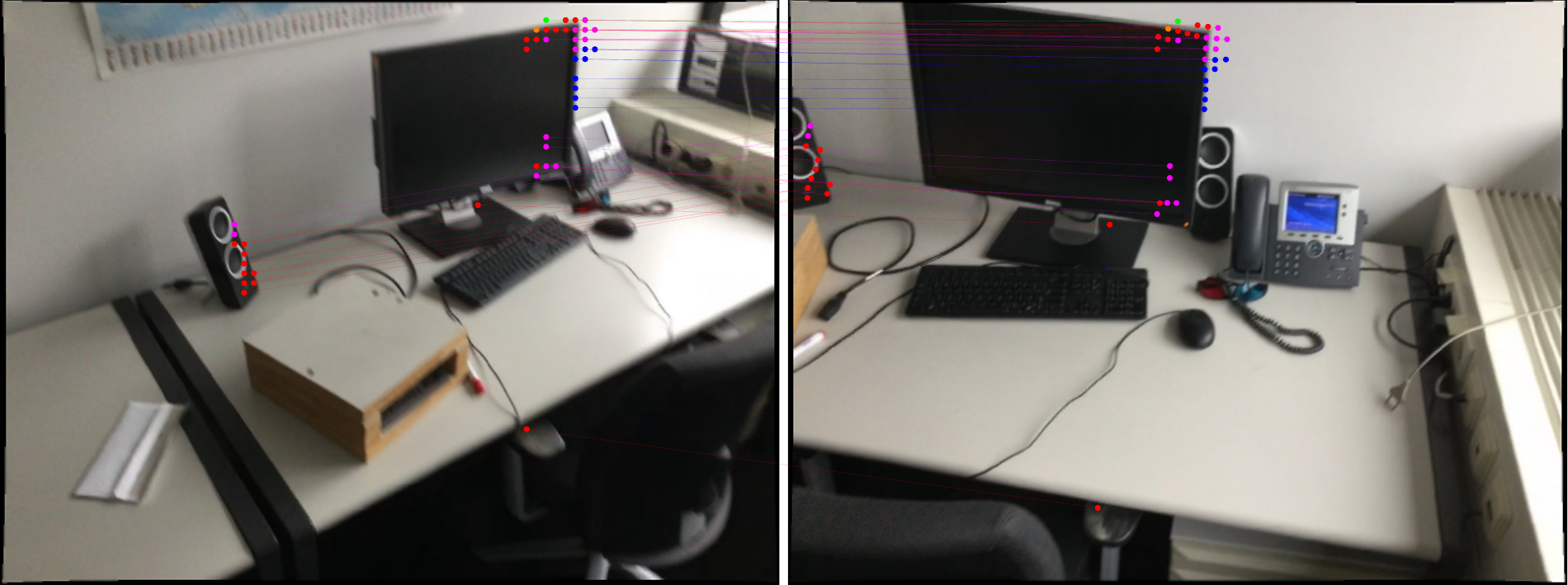} & \multirow{-5.5}{1em}{\rotatebox{-90}{\resizebox{!}{0.57em}{\hspace{0.5em}+MAGSAC}}} \\ 				
		\midrule		
	\end{tabular}	
	\caption{Visual qualitative comparison for non-planar scenes on example image pairs of MegaDepth (top row sections) and ScanNet (bottom row section). The base pipeline specified on the left is optionally combined with MOP+MiHo+NCC or FC-GGN, and then MAGSAC, as indicated by the top and right labels, respectively. Matches are colored according to the maximum epipolar error with respect to the GT such that the five consecutive intervals in $[0,1,3,7,15,\infty]$ px are colored as \textcolor{green}{\rule{1em}{1em}}\textcolor{orange}{\rule{1em}{1em}}\textcolor{red}{\rule{1em}{1em}}\textcolor{magenta}{\rule{1em}{1em}}\textcolor{blue}{\rule{1em}{1em}}. Please refer to Sec.~\ref{patch_showcase_discussion}. Best viewed in color and zoomed in.}\label{showcase1}
\end{figure}

\begin{figure}
	\centering
	\begin{tabular}{c@{\hskip 0.6em}c@{\hskip 0.6em}c@{\hskip 0.6em}c@{\hskip 0.6em}c}
		& & \resizebox{!}{0.54em}{+MOP+MiHo+NCC\hphantom{g}} & \resizebox{!}{0.54em}{+FC-GNN\hphantom{j}} \\
		\midrule
		& \includegraphics[width=0.27\textwidth]{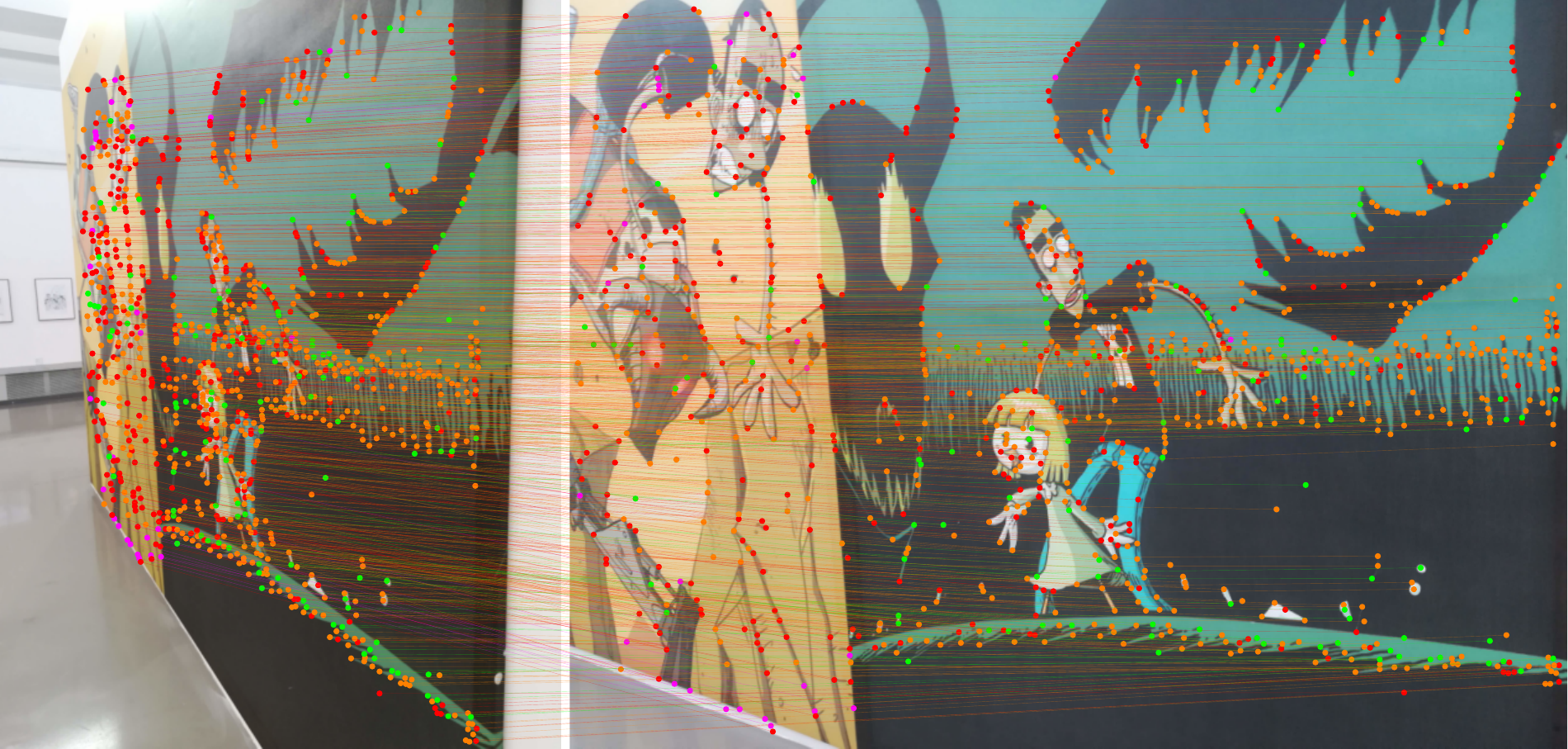} & \includegraphics[width=0.27\textwidth]{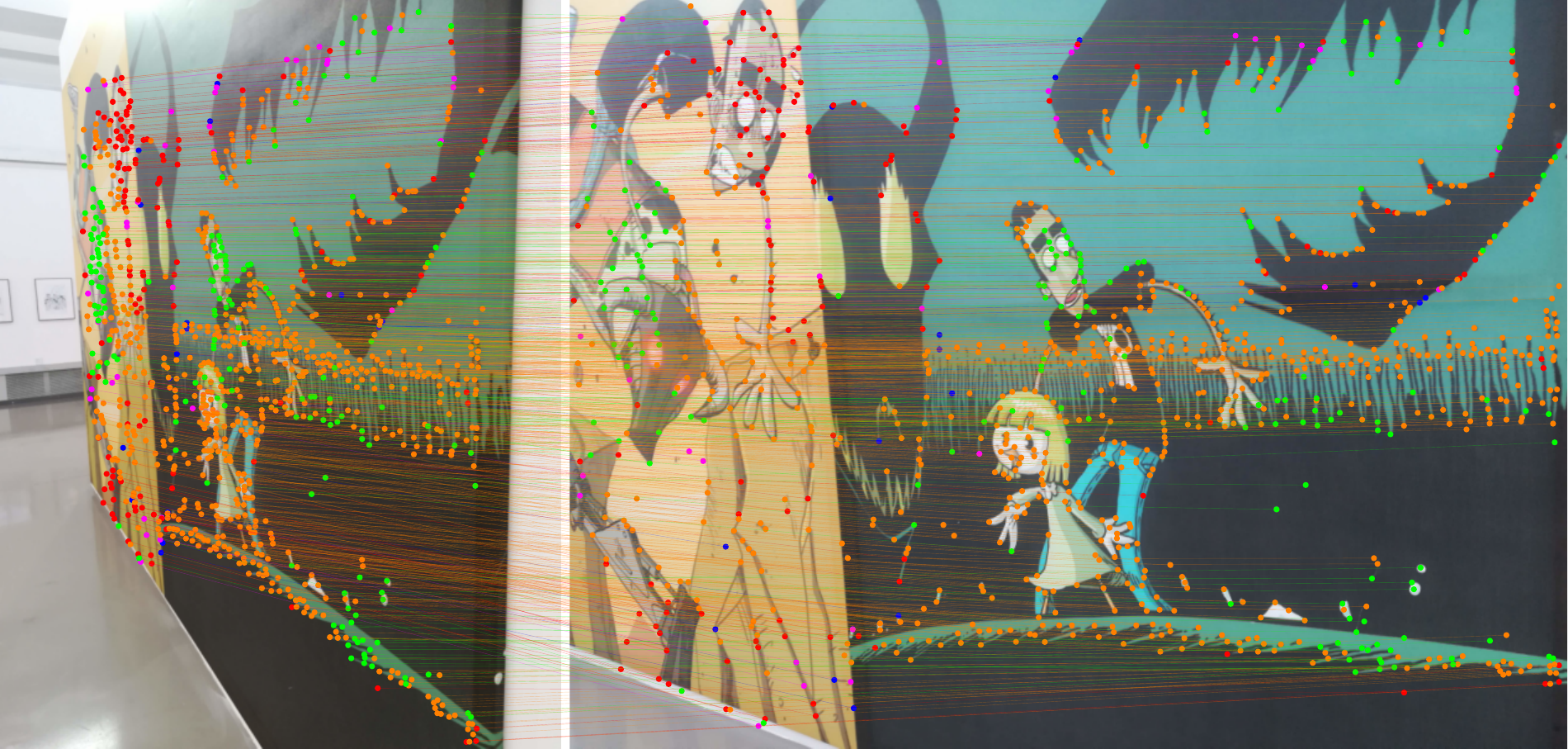} & \includegraphics[width=0.27\textwidth]{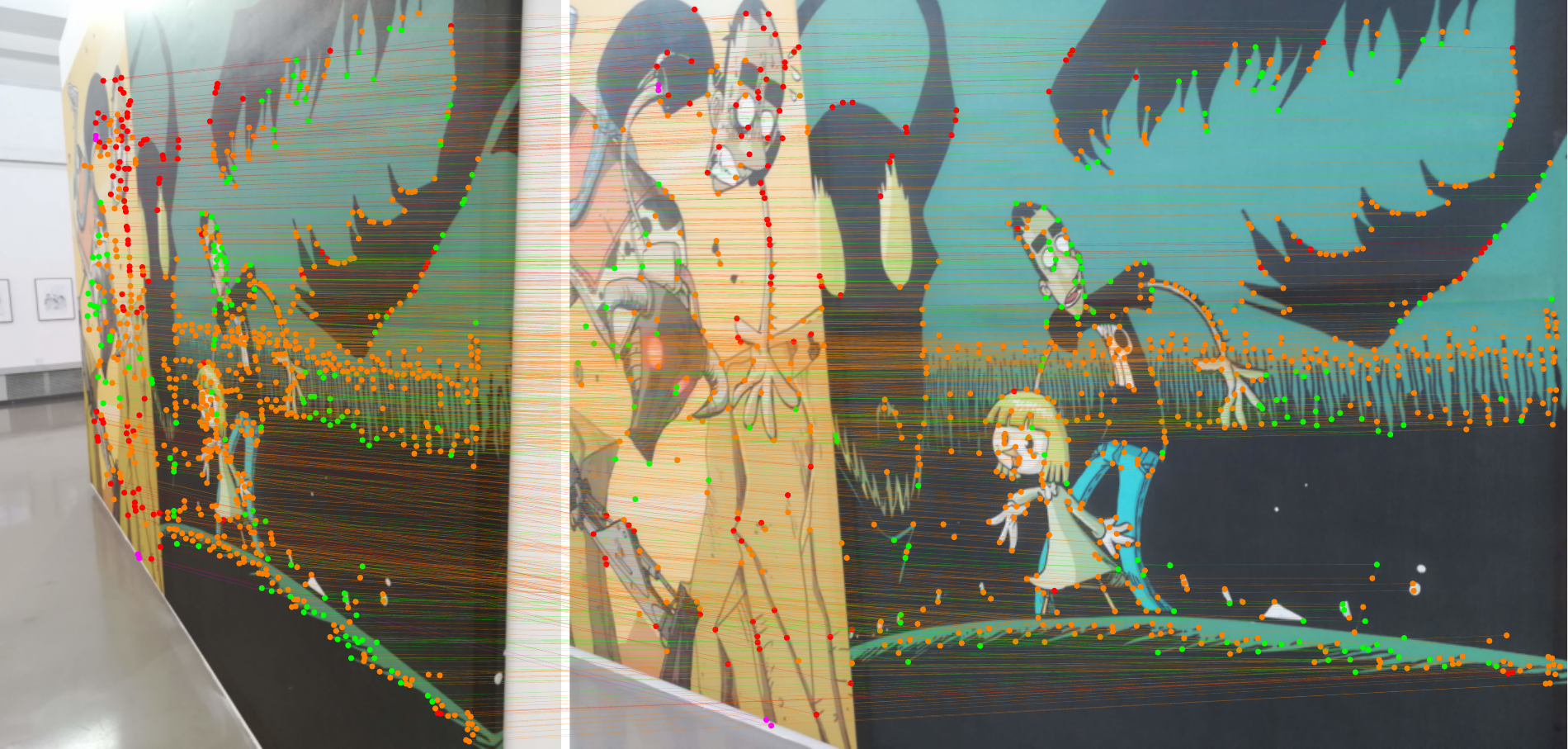} \\ 		
		\multirow{-10}{1em}{\rotatebox{90}{\resizebox{!}{0.57em}{SuperPoint+LightGlue}}} & \includegraphics[width=0.27\textwidth]{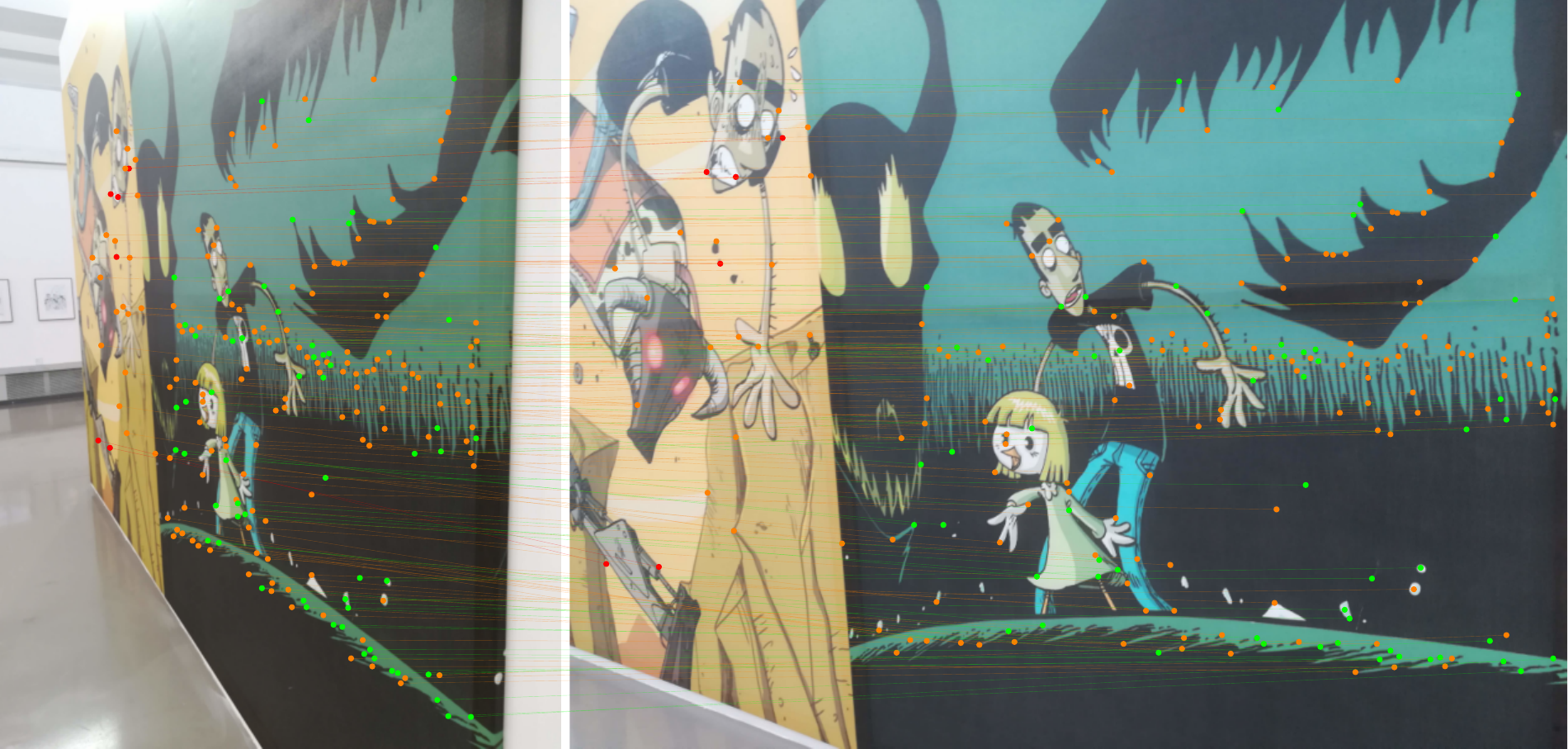} & \includegraphics[width=0.27\textwidth]{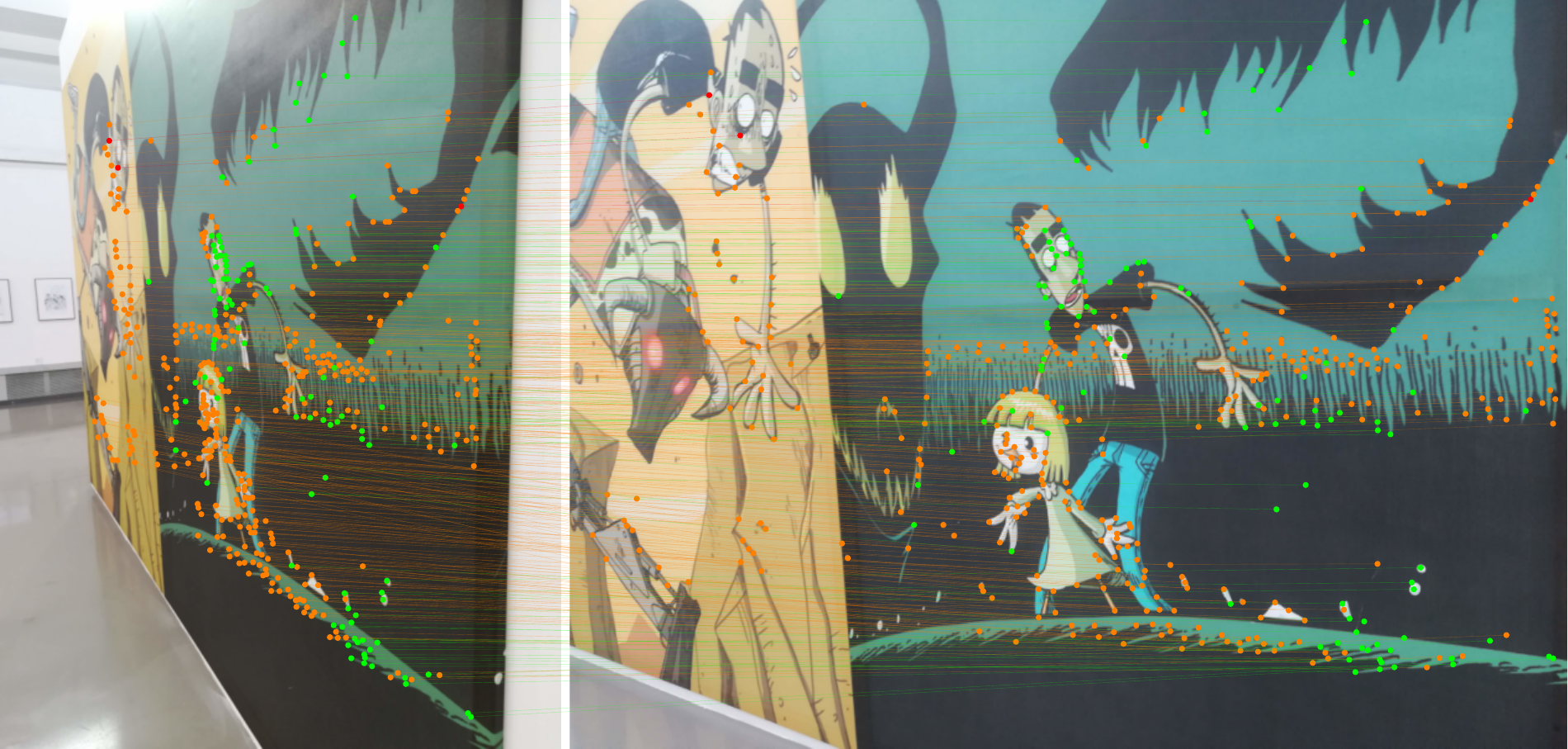} & \includegraphics[width=0.27\textwidth]{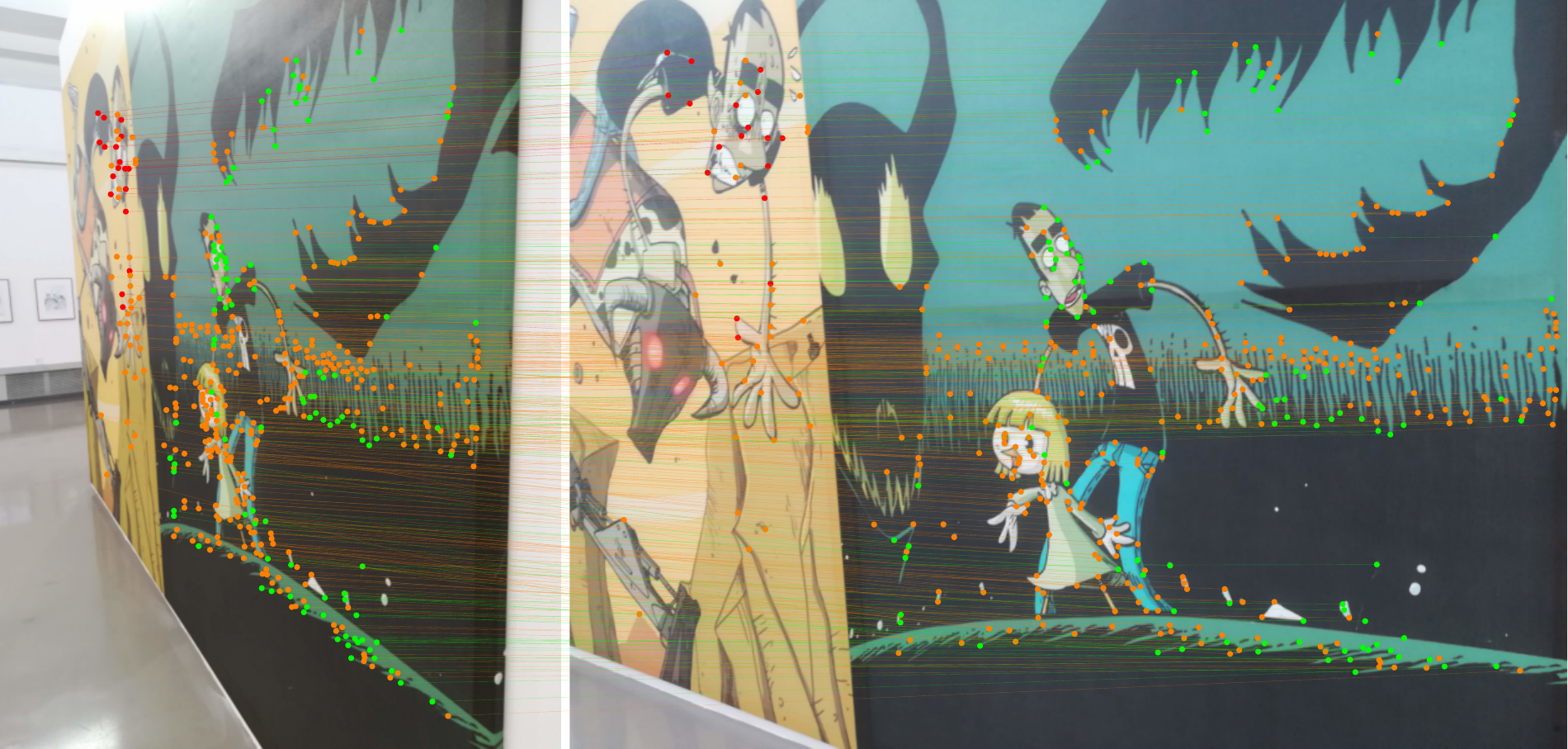} & \multirow{-6.5}{1em}{\rotatebox{-90}{\resizebox{!}{0.6em}{\hspace{0.57em}+MAGSAC}}} \\ 		
		\midrule
		& \includegraphics[width=0.27\textwidth]{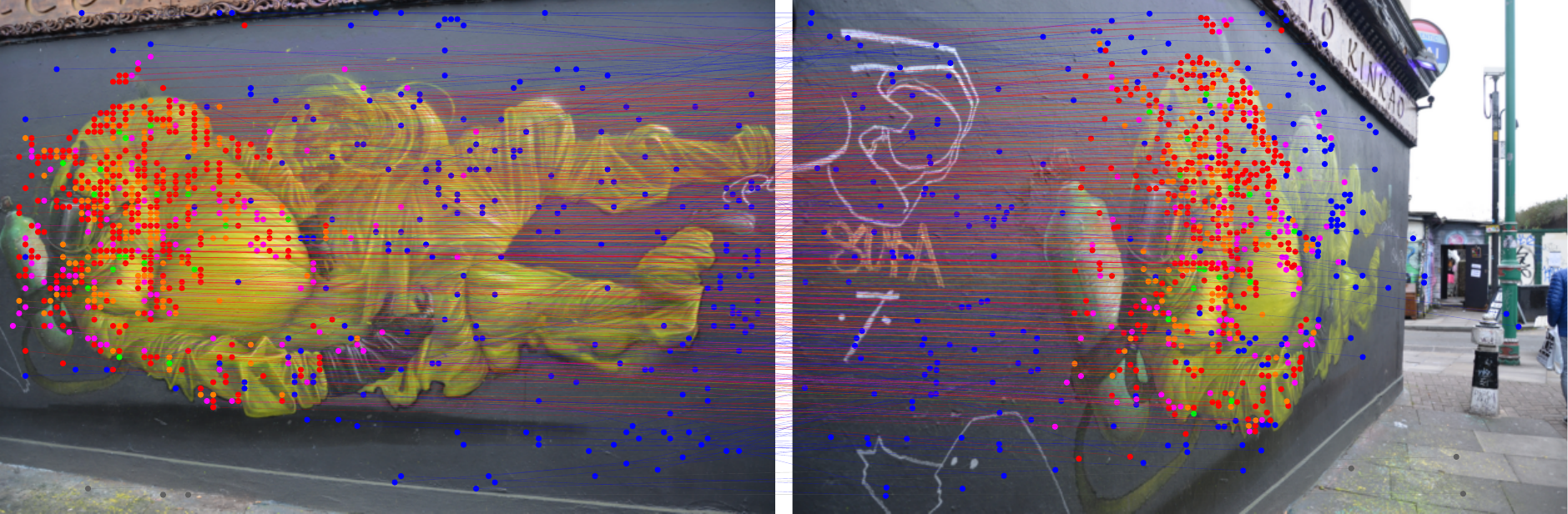} & \includegraphics[width=0.27\textwidth]{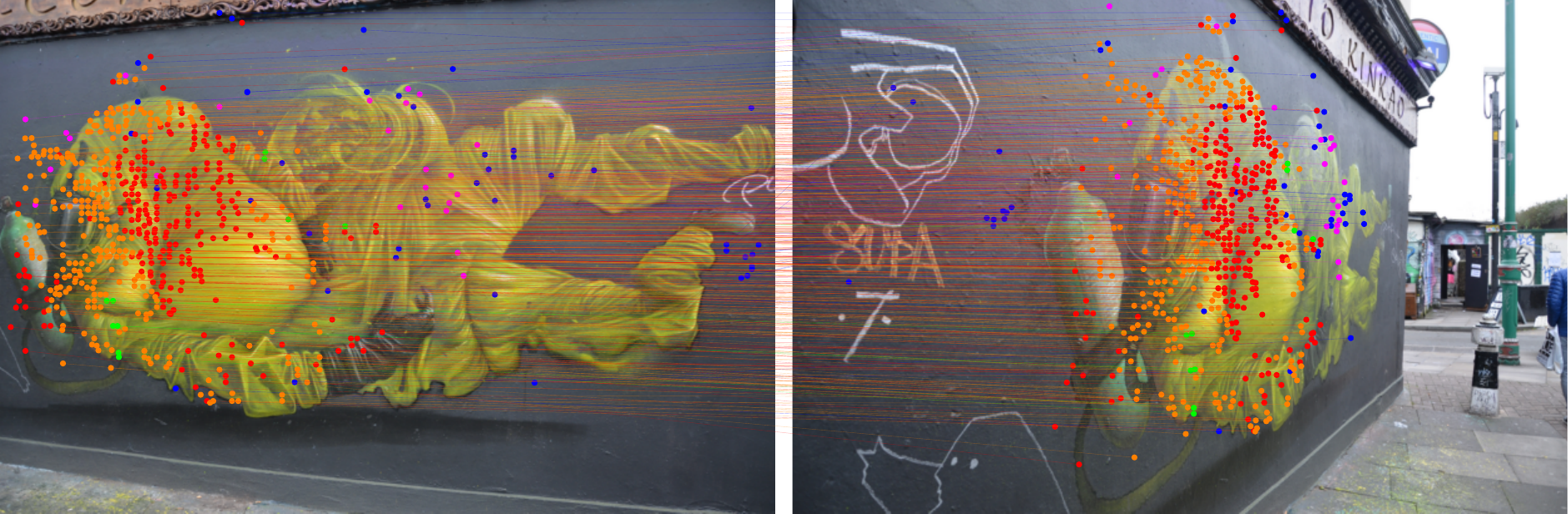} & \includegraphics[width=0.27\textwidth]{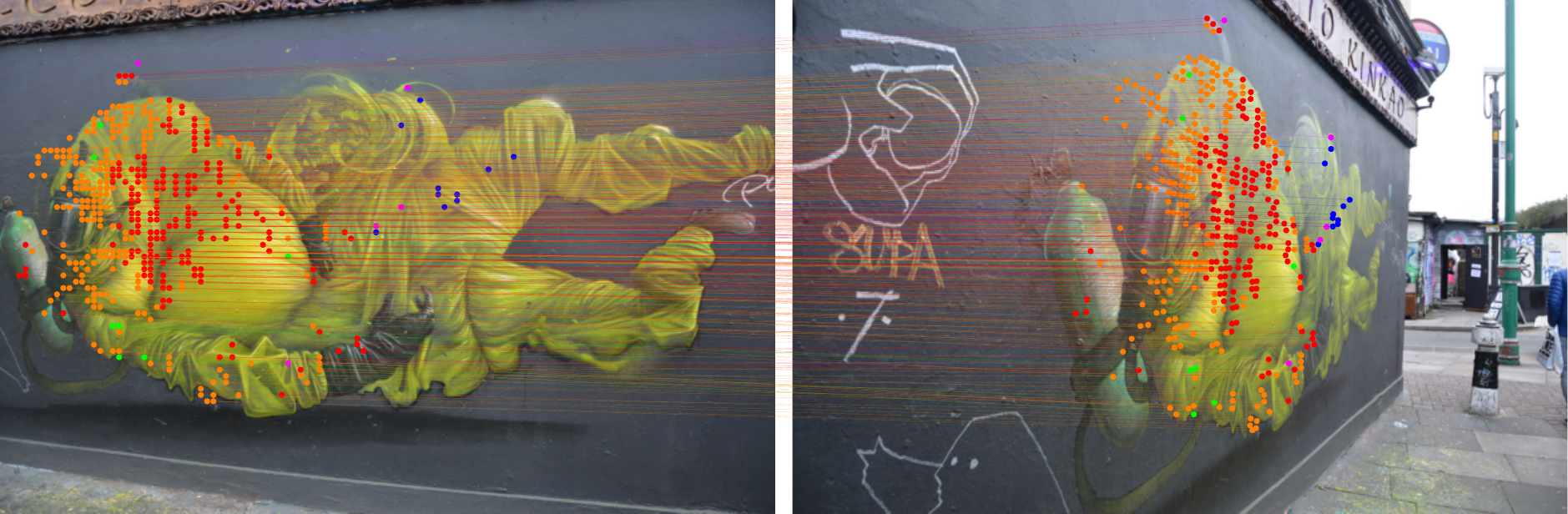} \\ 		
		\multirow{-8}{1em}{\rotatebox{90}{\resizebox{!}{0.57em}{LoFTR\hphantom{g}}}} & \includegraphics[width=0.27\textwidth]{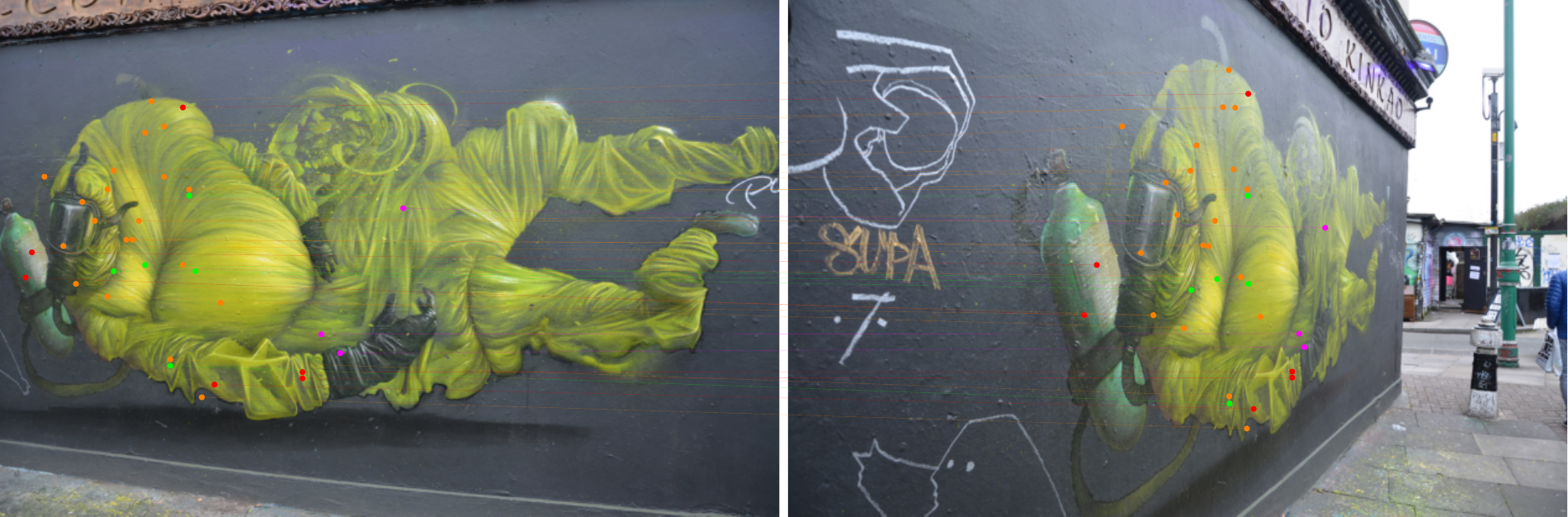} & \includegraphics[width=0.27\textwidth]{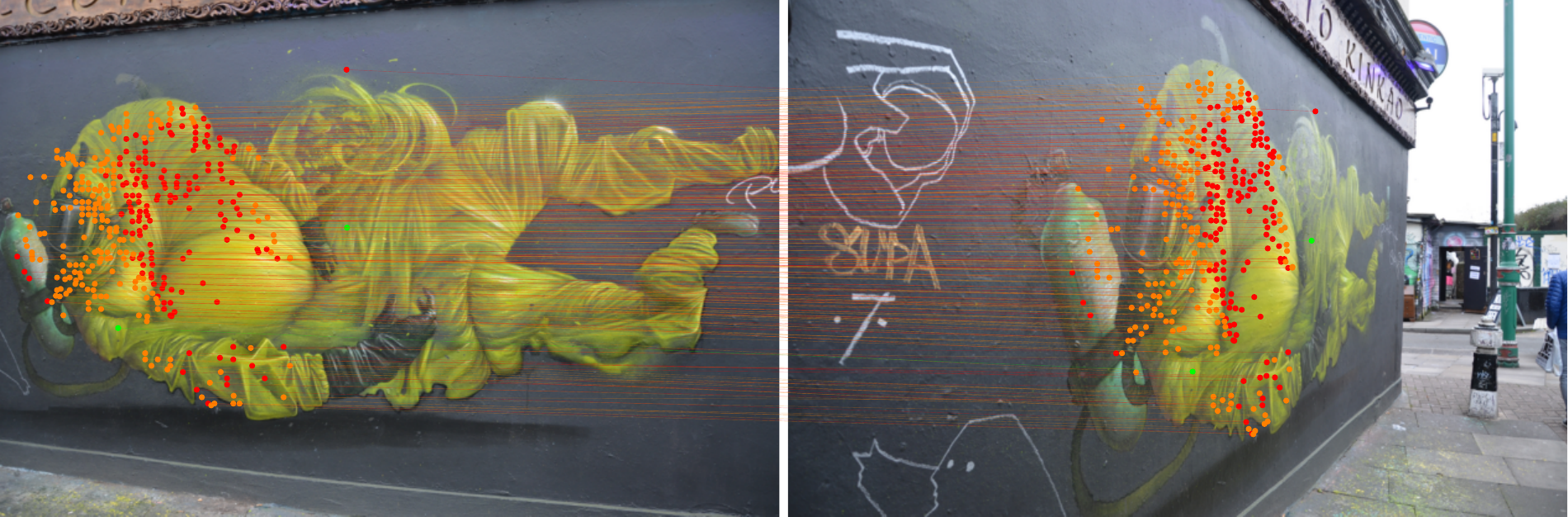} & \includegraphics[width=0.27\textwidth]{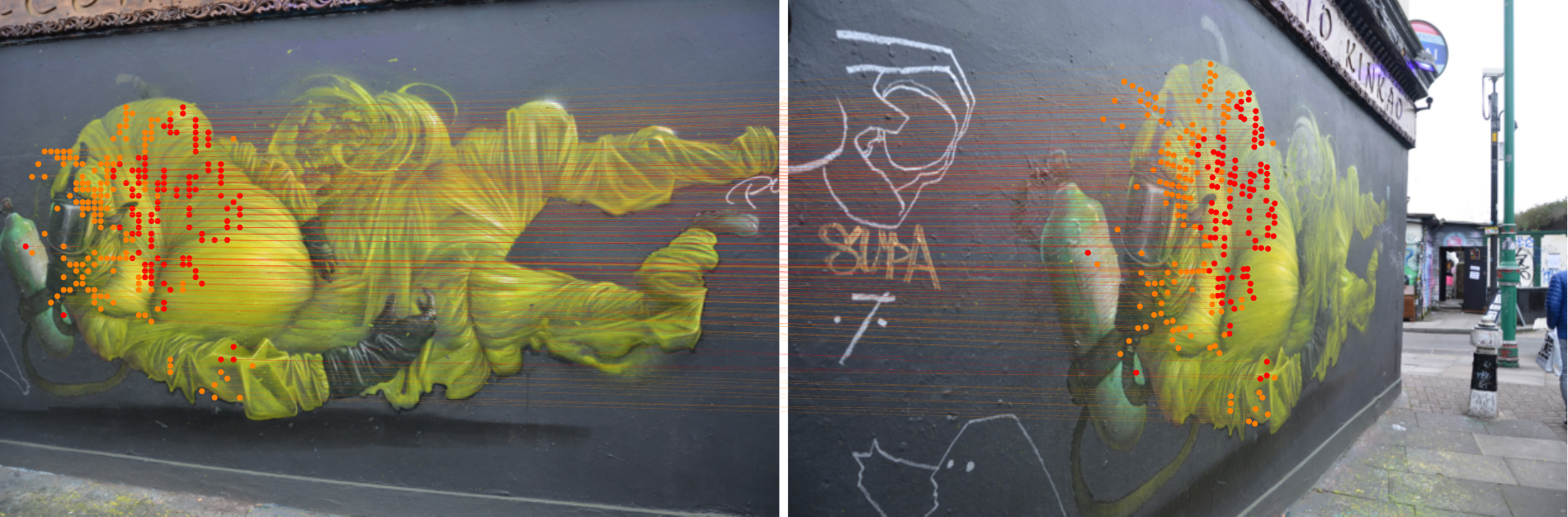} & \multirow{-5.3}{1em}{\rotatebox{-90}{\resizebox{!}{0.6em}{\hspace{0.57em}+MAGSAC}}} \\  				
		\midrule		
	\end{tabular}	
	\caption{Visual qualitative comparison for planar scenes in the Planar dataset. The same conventions of Fig.~\ref{showcase1} are employed. Matches are colored according to the maximum epipolar error with respect to the GT such that the five consecutive intervals in $[0,1,3,7,15,\infty]$ px are colored as \textcolor{green}{\rule{1em}{1em}}\textcolor{orange}{\rule{1em}{1em}}\textcolor{red}{\rule{1em}{1em}}\textcolor{magenta}{\rule{1em}{1em}}\textcolor{blue}{\rule{1em}{1em}}. Matches where both keypoints are outside the considered plane so that the GT is not valid are marked in gray. Please refer to Sec.~\ref{patch_showcase_discussion}. Best viewed in color and zoomed in.}\label{showcase2}
\end{figure}

Figures~\ref{showcase1}-\ref{showcase2} present some example image pairs, respectively in the case of non-planar and planar scenes, for base matching pipelines where NCC refinement is effective according to the quantitative evaluation. The base matches after being filtered by MOP+MiHo+NCC and FC-GNN, with or without RANSAC, are reported, where the match color indicates the corresponding maximum epipolar error within the pair images.

Comparing the first row for each scene, i.e. without RANSAC-based filtering, both MOP+MiHo and FC-GNN remove outliers, highlighted in blue, but MOP+MiHo is more relaxed in the task, acting as a sort of coarse Local Optimized RANSAC (LO-RANSAC,~\cite{lo-ransac}) step. As reported in Tables~\ref{sift_res}-\ref{aliked_res}, this is beneficial for the effective RANSAC especially when the outlier contamination of the input matches is high, such as for SIFT and Key.Net, and does not negatively affect the final results in the other cases. Concerning the refinement, both NCC and FC-GNN are able to improve the keypoint accuracy, as indicated by the color shift of the matches, but the epipolar error variation of their refined matches differs. For NCC the same observations and considerations discussed for Figure~\ref{patches} hold. 

\begin{figure}
	\centering
	\begin{tabular}{c@{\hskip 0.6em}c@{\hskip 0.6em}c@{\hskip 0.6em}c@{\hskip 0.6em}c}
		& & \resizebox{!}{0.54em}{+MOP+MiHo+NCC\hphantom{g}} & \resizebox{!}{0.54em}{+FC-GNN\hphantom{j}} \\
		\midrule
		& \includegraphics[width=0.27\textwidth]{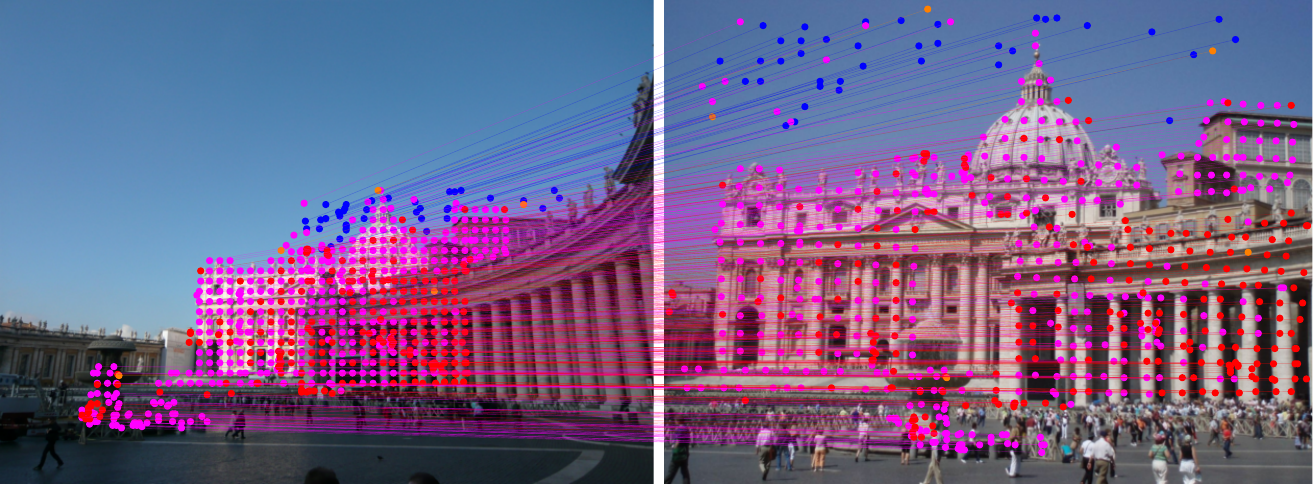} & \includegraphics[width=0.27\textwidth]{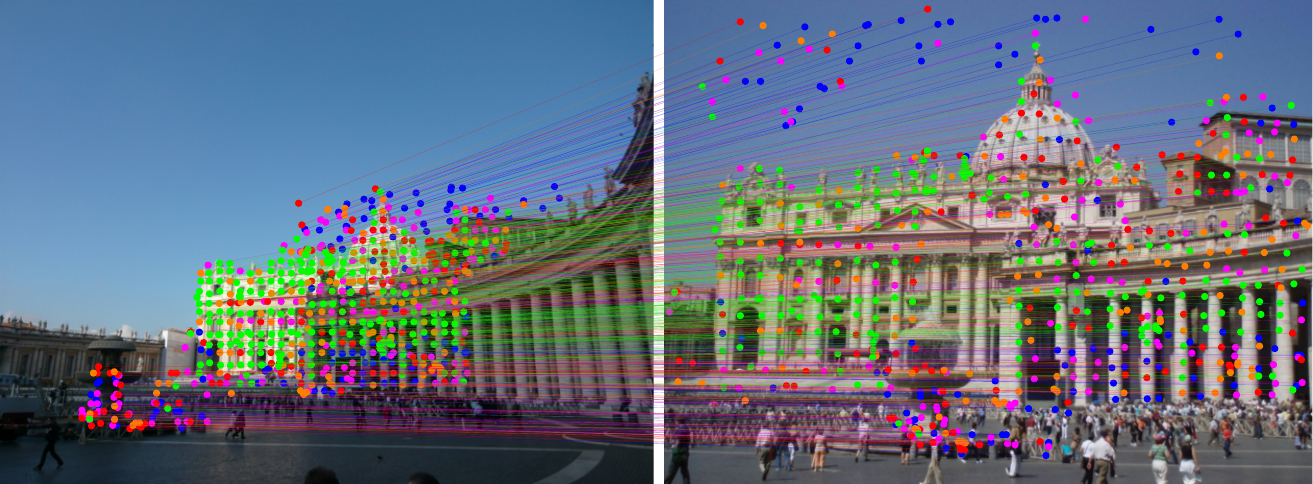} & \includegraphics[width=0.27\textwidth]{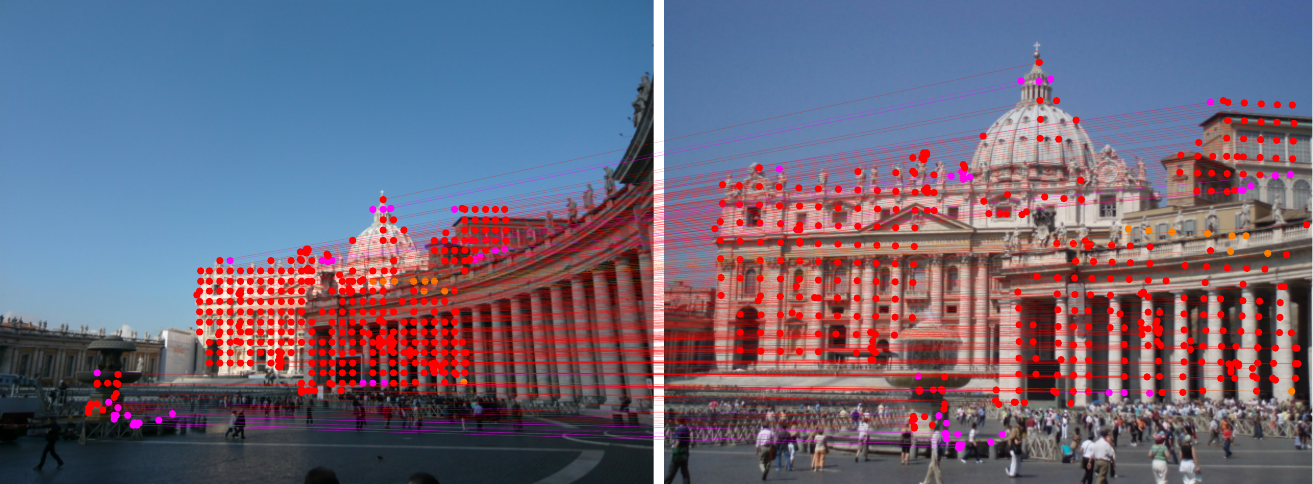} \\ 		
		\multirow{-8}{1em}{\rotatebox{90}{\resizebox{!}{0.57em}{MASt3R}}} & \includegraphics[width=0.27\textwidth]{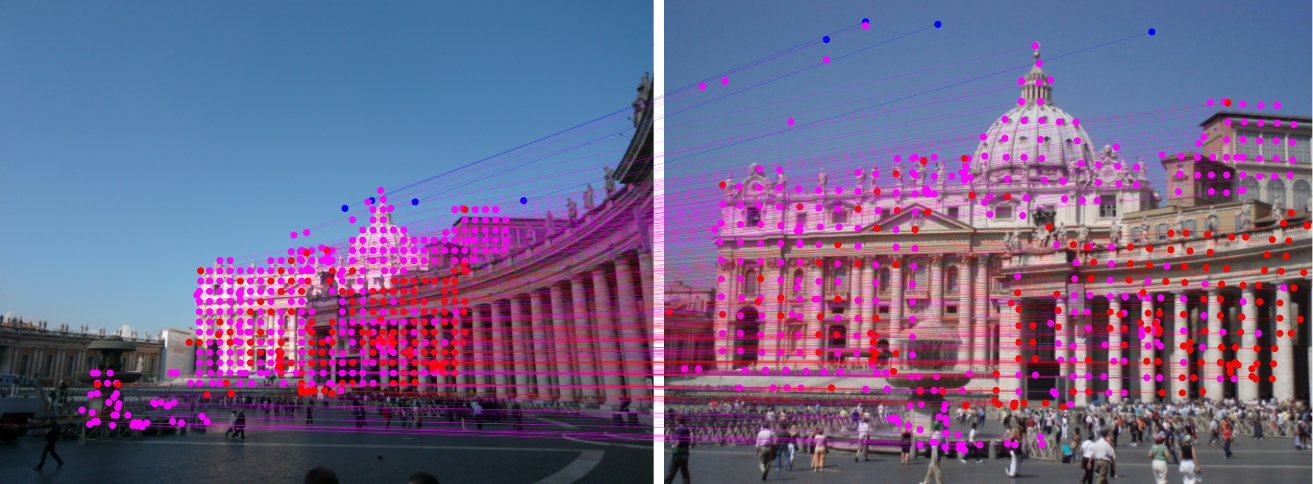} & \includegraphics[width=0.27\textwidth]{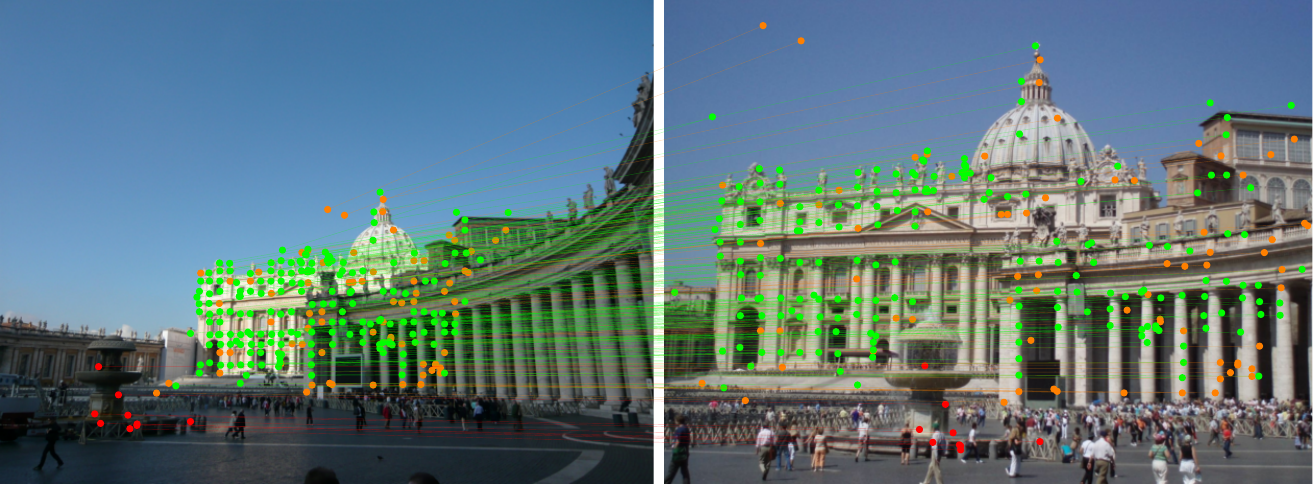} & \includegraphics[width=0.27\textwidth]{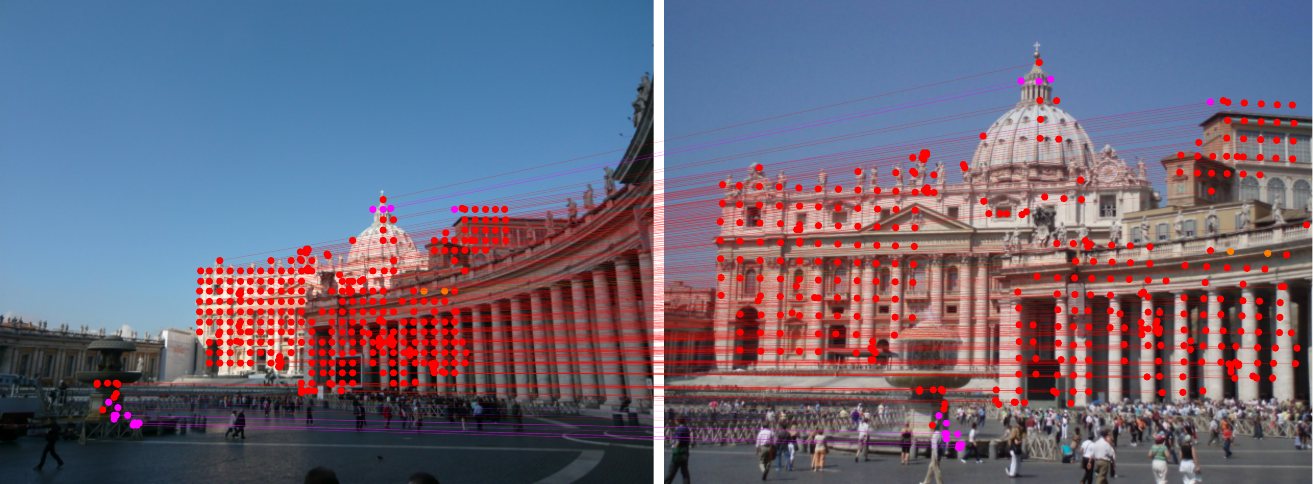} & \multirow{-5.5}{1em}{\rotatebox{-90}{\resizebox{!}{0.6em}{\hspace{0.57em}+MAGSAC}}} \\ 		
		\midrule
		& \includegraphics[width=0.27\textwidth]{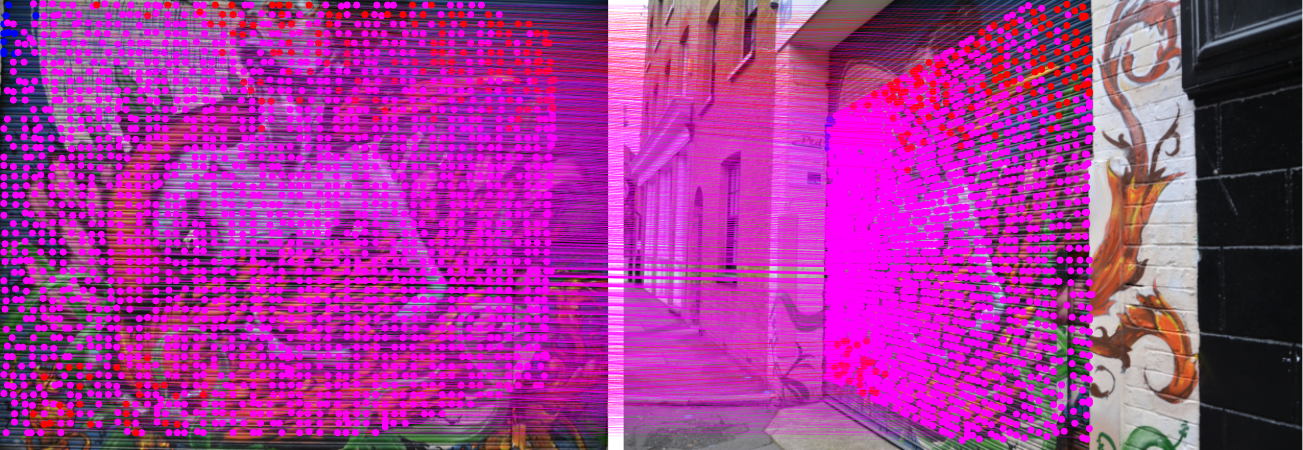} & \includegraphics[width=0.27\textwidth]{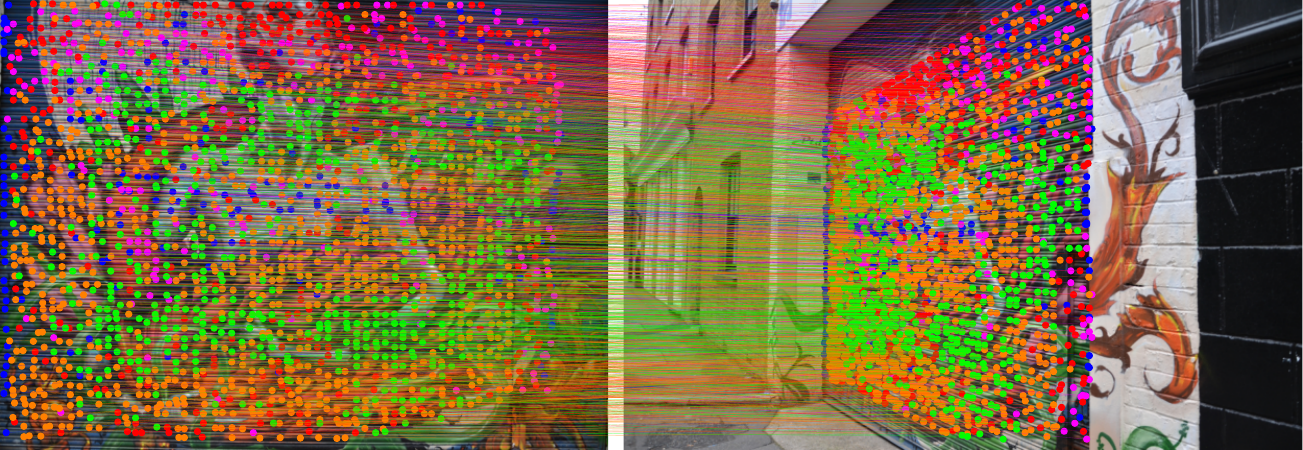} & \includegraphics[width=0.27\textwidth]{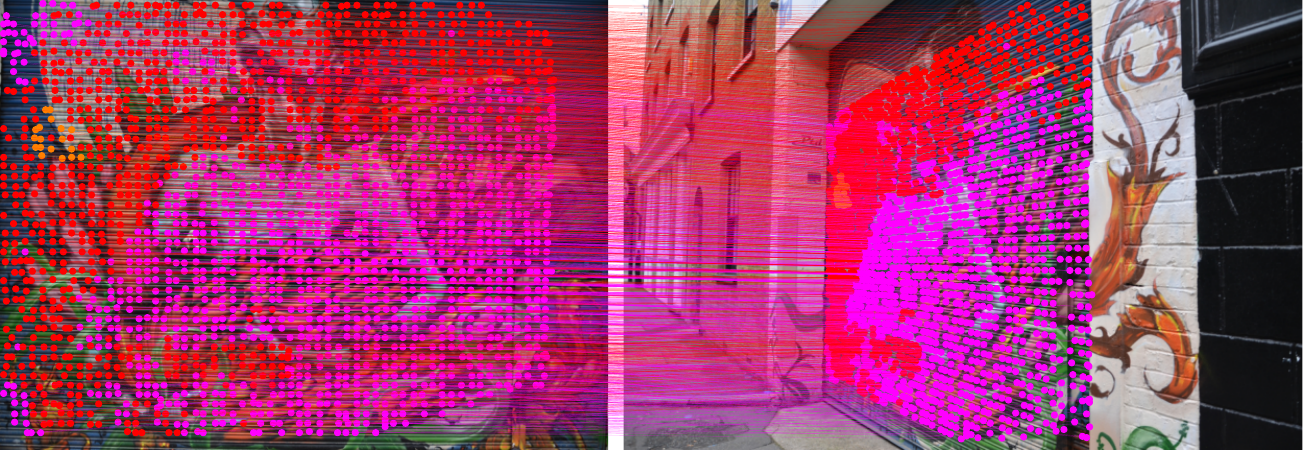} \\ 		
		\multirow{-8}{1em}{\rotatebox{90}{\resizebox{!}{0.57em}{MASt3R\hphantom{g}}}} & \includegraphics[width=0.27\textwidth]{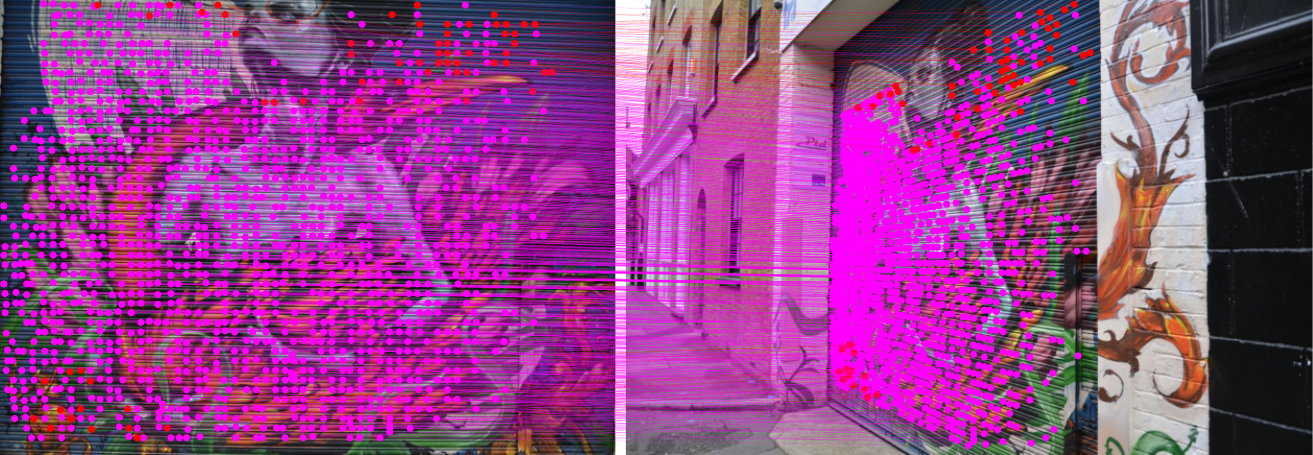} & \includegraphics[width=0.27\textwidth]{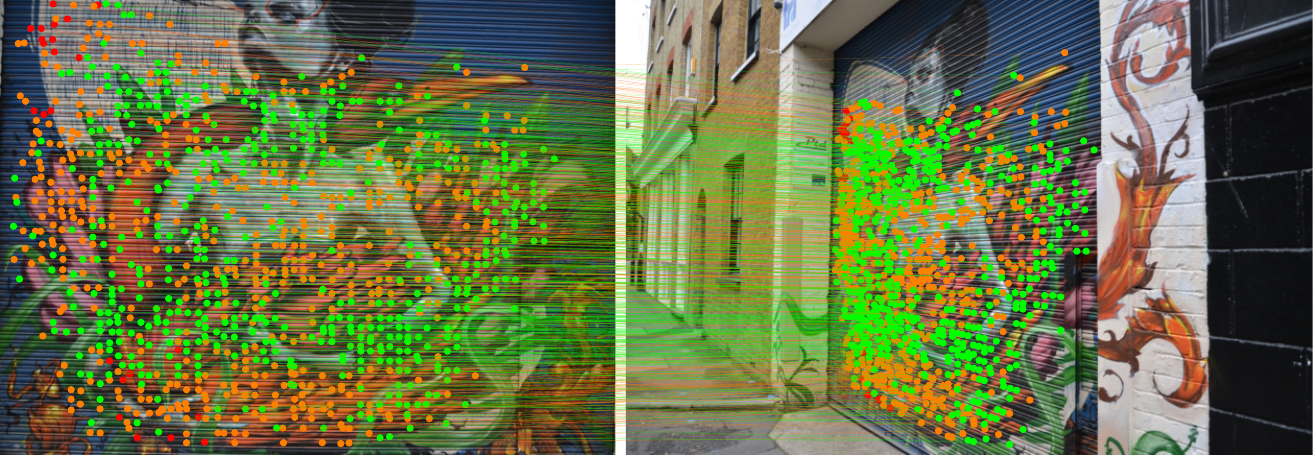} & \includegraphics[width=0.27\textwidth]{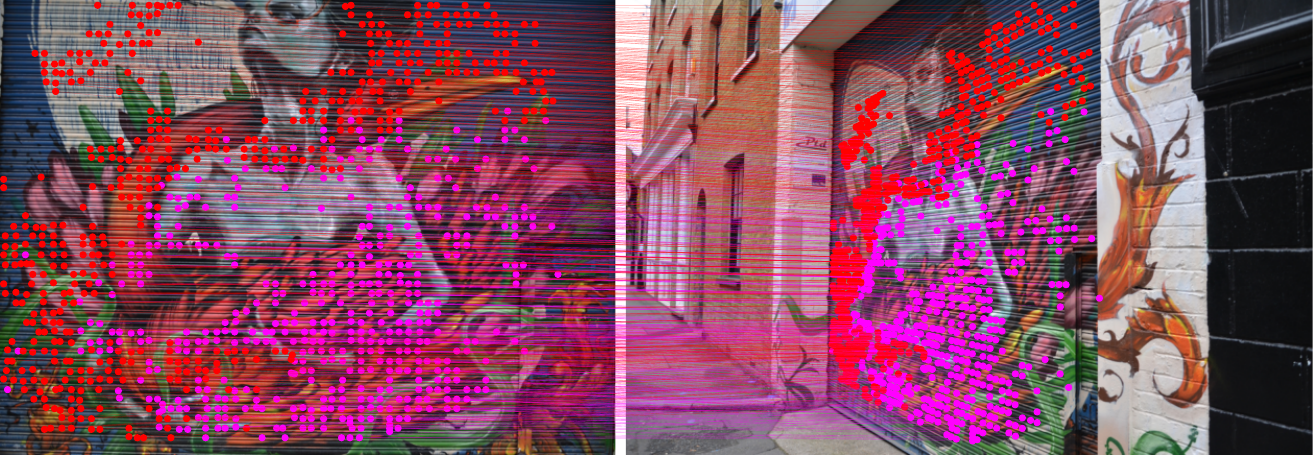} & \multirow{-5.3}{1em}{\rotatebox{-90}{\resizebox{!}{0.6em}{\hspace{0.57em}+MAGSAC}}} \\  				
		\midrule		
	\end{tabular}	
	\caption{Visual qualitative comparison for MASt3R. The same conventions of Figs.~\ref{showcase1}-\ref{showcase2} are employed. Matches are colored according to the maximum epipolar error with respect to the GT such that the five consecutive intervals in $[0,1,3,7,15,\infty]$ px are colored as \textcolor{green}{\rule{1em}{1em}}\textcolor{orange}{\rule{1em}{1em}}\textcolor{red}{\rule{1em}{1em}}\textcolor{magenta}{\rule{1em}{1em}}\textcolor{blue}{\rule{1em}{1em}}. Please refer to Sec.~\ref{patch_showcase_discussion}. Best viewed in color and zoomed in.}\label{showcase3}
\end{figure}

Finally, Fig.~\ref{showcase3} presents some example image pairs in the case of non-planar and planar scenes when MASt3R is the base pipeline, with the aim to complete the observations presented in Sec.~\ref{dense}. The MASt3R base matches are not very accurate in terms of maximum epipolar error as indicated by the red and purple colored matches in the left column, but these are quite consistent with the true epipolar geometry so these are almost all preserved by MAGSAC. In the center column, MOP does not remove ambiguous matches in the sky which are globally consistent with a virtual plane, while NCC effectively increases the accuracy for most of the matches. Nevertheless, in the case where the are no real matches such as for the fountain and crowd in the bottom part of the non-planar scene image, NCC provides non-matches with higher epipolar error which are no more globally consistent with the real true scene pose. Moreover, NCC fails in the  bottom right part of the arcade due to edge-like patches not feasible for template matching. Thus the pose estimation becomes less accurate. This does not happen in the planar scenes due to the different nature of the problem and of the constraints so that NCC provides a remarkable improvement. In the last column, one can see that FC-GNN in also able to improve the precision of the matches but it obtains more yet less accurate matches than MOP+MiHo+NCC. Differently from the proposed strategy, both MASt3R and FC-GNN are trained deep models relying on epipolar pose priors.

\begin{table}[h]
	\caption{Ablation study on MOP-based filtering and refinement. All values are percentages, please refer to Sec.~\ref{ablation}. Best viewed in color and zoomed in.}\label{ablation_res}
	\centering
	\includegraphics[width=0.7\textwidth]{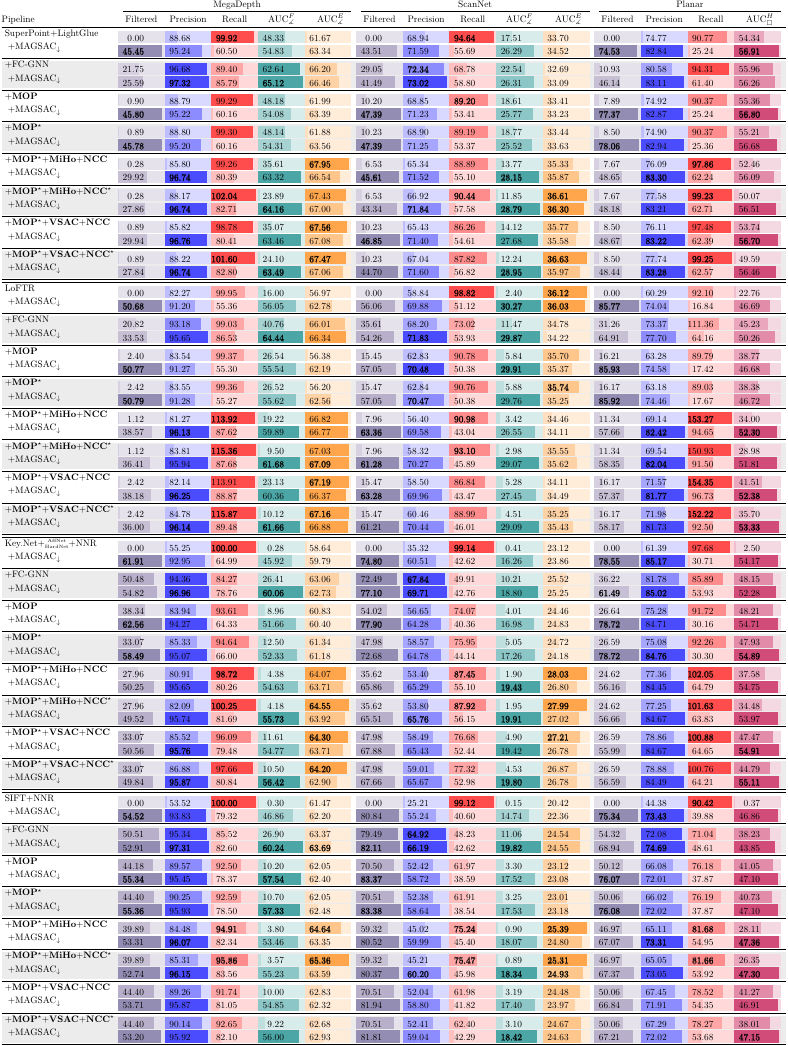}
\end{table}

\subsubsection{Ablation study}\label{ablation}
Table~\ref{ablation_res} reports the results obtained using several variations of the MOP-based approaches, including the reflection check through MOP$^\star$ (see Sec.~\ref{miho}), the replacement of MiHo with VSAC (see Sec.~\ref{miho}), and taking into account in template matching the patch gradient orientation through NCC$^\star$ (see Sec.~\ref{ncc}). The results are reported for the base pipelines where MOP-based approaches are effective and, besides the base pipeline, include FC-GNN which is the best overall deep alternative. Extended results are available as additional material\footref{github_results}.

Concerning MOP and MOP$^\star$ there are basically no differences. As there are no real planar reflections in the datasets, this confirms the robustness of the process, and the small fluctuations are likely due to the randomness of the MOP inner RANSAC process.

NCC$^\star$ usually provides an improvement upon NCC in terms of accuracy around 0.5-1\%, especially for the fundamental matrix estimation. NCC$^\star$ can be slightly worse in some cases, but even in the most noticeable cases, i.e. the essential matrix estimation for MegaDepth and the planar homography case for both SuperPoint and Key.Net, the degradation is quite negligible.

VSAC on average is slightly worse than MiHo, the differences between the two approaches are minimal. Only in the planar homography estimation VSAC is slightly better than MiHo, probably due to the fact that VSAC tends to adhere to the strict planar homography model more than MiHo.

Overall, MOP$^\star$+MiHo+NCC$^\star$ seem to provide the best combination, even if compared with MOP+MiHo+NCC these are not impressive. With respect to the base pipeline the MOP-based approaches provide evident improvements for the non-planar datasets, while on the planar dataset the differences are minimal with the exception of LoFTR which benefits of the filtering and refinements of the matches. 

Compared against the deep FC-GNN, MOP-based combinations with NCC offer a valid alternative, being the results comparable or better. Notice that FC-GNN is effectively preferable only for the fundamental estimation on MegaDepth were it was trained on, and in the planar dataset is noticeably worse than the base pipeline. 

Note also that even for the base pipelines presented in Sec.~\ref{other_sparse} and Sec.~\ref{dense}, the degradation of MOP-based filtering is almost null compared with other filtering approaches. The only exception is for the non-planar scenes with MASt3R when NCC is added to MOP-base filters for the reasons explained in Sec.~\ref{dense}. This implies that the approach presented is robust and stable.

\section{Conclusions and future work}\label{conclusion}
This paper presented a thorough analysis of filtering and refinement strategies for image matching. The reported comparative evaluation highlighted several critical aspects including the dependency of the filtering methods on the specific dataset and the base pipeline, and the pose estimation performance drop observed when camera intrinsic are unknown, quite common in real scenarios. The error metric is critical in the benchmark design in order to highlight the effective difference between the approaches. 

The idea explored in MOP of filtering image matches relying on the principle that the scene flow can be approximated as virtual local and overlapping planar homographies is effective and robust. Moreover, the patch alignment addressed by MiHo or VSAC to refine keypoints through NCC has demonstrated to be valid.  

MOP+MiHo+NCC and its variants are able to effectively discard outliers and to improve the match localization as well as the pose accuracy with SuperPoint, LoFTR, Key.Net and SIFT. When the base pipeline only contains accurate matches such as for ALIKED, DISK, and DeDoDe v2 and RoMa, the proposed pipeline does not alter the final output being stable and robust. In terms of absolute accuracy, LoFTR and SuperPoint reached overall best results among all the evaluated sparse pipelines when boosted by the proposed approach for MegaDepth and ScanNet. For the remaining non-planar IMC-PT dataset, the best is reached by DISK with FC-GNN, which is the current deep competitor and alternative to MOP+MiHo+NCC.

Overall, the above behavior and the gap reduction between handcrafted and hybrid pipelines with respect to deep SOTA networks achieved using the proposed ideas suggest that current deep architectures are consistent and implicitly compatible with the proposed strategies of more traditional computer vision. As indirect consequences, the proposed approach can be used upon SIFT or SuperPoint to build better baselines for future development and evaluation of matching approaches.  

MOP has proved to be effective in finding spatial planar clusters. As future work these clusters could be organized and exploited more in-depth to further filter matches as well as to expand them within the framework described by~\cite{slime}, also taking into account the epipolar geometry of the scene. Moreover, planar clusters could be used also to characterize the scene structure at higher levels in terms of object segmentation and the relative positions and occlusions of the found elements.

MiHo is able to improve the alignment in almost-planar motion, yet it lacks a rigorous mathematical formulation which would help to better understand its properties and limitations. Future development in this direction will be explored, as well as its possible applications to current image stitching solutions.

NCC inherits the strengths and weaknesses of the base template matching approach it is based on. NCC is effective on corner-like patches but on strong edge-like structures or flat homogeneous regions, more typical of blobs, NCC can mislead the alignment refinement. Nevertheless, NCC$^\star$ shows that base template matching can be improved exploiting further options. More elaborated and robust approaches based on the gradient orientation histogram correlation or variational methods will be taken in consideration for future developments. Match densification through local patch homographies will be investigated too.

MOP+MiHo+NCC shifts the same initial keypoint belonging to an image in different directions when this image is paired with other images of the same scene. This implies that the proposed method cannot be embedded directly within current SfM pipelines since it would prevent the generation of keypoint tracks. Solutions for handling multiple keypoint shifts in SfM, for instance by constraint keypoint refinement to only one image by a greedy strategy as more images are added to the model, will be evaluated in future work. 

MOP+MiHo+NCC offers the advantage of handcrafted design with respect to deep architectures, including explainability, direct design tuning, and adaptability for specific tasks and no training data dependence. Nevertheless, MOP+MiHo+NCC sequential nature inherited by RANSAC makes it less efficient in terms of computation in nowadays parallel hardware and programming infrastructures exploited straightly by deep networks. Future works will investigate design approaches and optimization strategies to speed up the computational times.  


\section*{Appendix}
\appendix

\section{MOP}\label{mop_appendix}
\subsection{Preliminaries}
MOP assumes that motion flows within the images can be piecewise approximated by local planar homography.

Representing with an abuse of notation the reprojection of a point $\mathbf{x}\in\mathbb{R}^2$ through a planar homography $\mathrm{H}\in\mathbb{R}^{3\times3}$ as $\mathrm{H}\mathbf{x}$, where $\mathrm{H}$ is non-singular, a match $m=(\mathbf{x}_1,\mathbf{x}_2)$ within two keypoints $\mathbf{x}_1\in I_1$ and $\mathbf{x}_2\in I_2$ is considered an inlier based on the maximum reprojection error
\begin{equation}\label{reproj_error}
	\varepsilon_\mathrm{H}(m)=\max\left(\left\|\mathbf{x}_2-\mathrm{H}\mathbf{x}_1\right\|,\left\|\mathbf{x}_1-\mathrm{H}^{-1}\mathbf{x}_2\right\|\right)
\end{equation}
For a set of matches $\mathcal{M}=\{m\}$, the inlier subset is provided according to a threshold $t$ as
\begin{equation}\label{inliers}
	\mathcal{J}^\mathcal{M}_{\mathrm{H},t}=\left\{m\in\mathcal{M}: \varepsilon_\mathrm{H}(m)\leq t\right\}
\end{equation}
A proper scene planar transformation must be quasi-affine to preserve the convex hull~(\cite{multiview}) with respect to the minimum sampled model generating $\mathrm{H}$ in RANSAC. The the inlier set effectively employed is
\begin{equation}\label{full_inliers}
	\mathcal{I}^\mathcal{M}_{\mathrm{H},t}=\mathcal{J}^\mathcal{M}_{\mathrm{H},t}\cap\mathcal{Q}^\mathcal{M}_\mathrm{H}
\end{equation}
where $\mathcal{Q}^\mathcal{M}_\mathrm{H}$ is the set of matches in $\mathcal{M}$ satisfying the quasi-affinity property with respect to the RANSAC minimum model of $\mathrm{H}$, described later in Eq.~\ref{quasi_affine}.

\subsection{Main loop}\label{main_loop}
MOP iterates through a loop where a new homography is discovered and the match set is reduced for the next iteration. Iterations halt when the counter $c_f$ of the sequential failures at the end of a cycle reaches the maximum allowable value $c_f^\star=3$. If not incremented at the end of the cycle, $c_f$ is reset to 0. As output, MOP returns the set $\mathcal{H}^\star$ of found homographies $\mathrm{H}$ which covers the image visual flow. 

Starting with $c_f=0$, $\mathcal{H}^0=\emptyset$ and the initial set of input matches $\mathcal{M}^0$ provided by the image matching pipeline, at each iteration $k$ MOP looks for the best planar homography $\mathrm{H}^k$ compatible with the current match set $\mathcal{M}^k$ according to a relaxed inlier threshold $t_l=15$ px. When the number of inliers provided by $t_l$ is less than the minimum required amount, set to $n=12$, i.e.
\begin{equation}\label{h_bad}
	\left|\mathcal{I}^{\mathcal{M}^k}_{\mathrm{H}^k,t_l}\right| < n
\end{equation}
the current $\mathrm{H}^k$ homography is discarded and the failure counter $c_f$ is incremented without updating the iteration counter $k$, the next match set $\mathcal{M}^{k+1}$ and the output list $\mathcal{H}^{k+1}$. Otherwise, MOP removes the inlier matches of $\mathrm{H}^k$ for a given threshold $t^k$ in the next iteration $k+1$
\begin{equation}
	\mathcal{M}^{k+1}=\mathcal{M}^{k}\setminus\mathcal{I}^k
\end{equation}
where $\mathcal{I}^k=\mathcal{I}^{\mathcal{M}^k}_{\mathrm{H}^k,t^k}$ and updates the output list
\begin{equation}
	\mathcal{H}^{k+1}=\mathcal{H}^k\cup\{\mathrm{H}^k\}    
\end{equation}
The threshold $t^k$ adapts to the evolution of the homography search and is equal to a strict threshold $t_h=t_l/2$ in case the cardinality of the inlier set using $t_h$ is greater than half of $n$, i.e. in the worst case no more than half of the added matches should belong to previous overlapping planes. If this condition does not hold $t^k$ turns into the relaxed threshold $t_l$ and the failure counter $c_f$ is incremented
\begin{equation}
	t^k =
	\begin{cases}
		t_h & \text{if}\quad \left|\mathcal{I}^{\mathcal{M}^k}_{\mathrm{H}^k,t_h}\right| > \frac{n}{2} \\
		t_l & \text{otherwise}
	\end{cases}	
\end{equation}
Using the relaxed threshold $t_l$ to select the best homography $\mathrm{H}^k$ is required since the real motion flow is only locally approximated by planes, and removing only strong inliers by $t_h$ for the next iteration $k+1$ guarantees smooth changes within overlapping planes and more robustness to noise. Nevertheless, in case of slow convergence or when the homography search gets stuck around a wide overlapping or noisy planar configuration, limiting the search to matches excluded by previous planes can provide a way out. The final set of homography returned is
\begin{equation} 
	\mathcal{H}^\star=\mathcal{H}^{k^\star}
\end{equation}
found by the last iteration $k^\star$ when the failure counter reaches $c_f^\star$.

\subsection{Inside RANSAC}\label{mop_ransac}
Within each MOP iteration $k$, a homography $\mathrm{H}^k$ is extracted by RANSAC on the match set $\mathcal{M}^k$. The vanilla RANSAC is modified according to the following heuristics to improve robustness and avoid degenerate cases.

A minimum number $c_{\min}=50$ of RANSAC loop iterations is required besides the max number of iterations $c_{\max}=2000$. In addition, a minimum hypothesis sampled model
\begin{equation}\label{sample}
	\mathcal{S}=\left\{\left(\mathbf{s}_{1i},\mathbf{s}_{2i}\right)\right\}\subseteq\mathcal{M}^k,\quad 1\leq i\leq 4    
\end{equation}
is discarded if the distance within the keypoints in one of the images is less than the relaxed threshold $t_l$, i.e. it must hold
\begin{equation}
	\min_{\substack{1\leq i,j\leq 4 \\ i\neq j \\ w=1,2}}\left\|\mathbf{s}_{wi}-\mathbf{s}_{wj}\right\|\geq t_l
\end{equation}
Furthermore, being $\mathrm{H}$ the corresponding homography derived by the normalized Direct Linear Transform (DLT) algorithm~\cite{multiview}, the smaller singular value found by solving through SVD the $8\times9$ homogeneous system
must be relatively greater than zero for a stable solution not close to degeneracy. According to these observations, the minimum singular value is constrained to be greater than 0.05. Note that an absolute threshold is feasible since data are normalized before SVD.

Next, matches in $\mathcal{S}$ must satisfy quasi-affinity with respect to the derived homography $\mathrm{H}$. This can be verified by checking that the sign of the last coordinate of the reprojection $\mathrm{H}\mathbf{s}_{1i}$ in non-normalized homogenous coordinates is the same for $1\leq i\leq 4$, i.e. for all the four keypoint in the first image. This constraint can be expressed as
\begin{equation}
	[\mathrm{H}\mathbf{s}_{1i}]_3 \cdot [\mathrm{H}\mathbf{s}_{1j}]_3 = 1,\quad\forall 1\leq i,j\leq 4
\end{equation}
where $[\mathbf{x}]_3$ denotes the third and last vector element of the point $\mathbf{x}$ in non-normalized homogeneous coordinates. Following the above discussion, the quasi-affine set of matches $\mathcal{Q}^\mathcal{M}_\mathrm{H}$ in Eq.~\ref{full_inliers} is formally defined as
\begin{equation}\label{quasi_affine}
	\mathcal{Q}^\mathcal{M}_\mathrm{H}=\left\{m\in\mathcal{M}: [\mathrm{H}\mathbf{x}_1]_3\cdot[\mathrm{H}\mathbf{s}_{11}]_3=1\right\}
\end{equation}
For better numerical stability, the analogous checks are executed also in the reverse direction through $\mathrm{H}^{-1}_\mathcal{S}$.

Lastly, to speed up the RANSAC search, a buffer $\mathcal{B}=\{\tilde{\mathrm{H}}_i\}$ with $1\leq i\leq z$ is globally maintained in order to contain the top $z=5$ discarded sub-optimal homographies encountered within a RANSAC run. For the current match set $\mathcal{M}^k$, if $\tilde{\mathrm{H}}_0$ is the current best RANSAC model maximizing the number of inliers, $\tilde{\mathrm{H}}_i\in\mathcal{B}$ are ordered by the number $v_i$ of inliers excluding those compatible with the previous homographies in the buffer, i.e. it must holds that
\begin{equation}
	v_{i}\geq v_{i+1}
\end{equation}
where
\begin{equation}
	v_i=\left|\mathcal{I}^{\mathcal{M}^k}_{\tilde{\mathrm{H}}_{i}}\setminus\left(\substack{\bigcup\\ {0\leq j<i}}\;\mathcal{I}^{\mathcal{M}^k}_{\tilde{\mathrm{H}}_j}\right)\right|   
\end{equation}\vspace{1em}

\noindent The buffer $\mathcal{B}$ can be updated when the number of inliers of the homography associated with the current sampled hypothesis is greater than the minimum $v_i$, by inserting in the case the new homography, removing the low-scoring one and resorting to the buffer. Moreover, at the beginning of a RANSAC run inside a MOP iteration, the minimum hypothesis model sampling sets $\mathcal{S}$ corresponding to homographies in $\mathcal{B}$ are evaluated before proceeding with true random samples to provide a bootstrap for the solution search. This also guarantees a global sequential inspection of the overall solution search space within MOP.   

\subsection{Homography assignment}\label{assign_homo}
After the final set of homographies $\mathcal{H}^\star$ is computed, the filtered match set $\mathcal{M}^\star$ is obtained by removing all matches $m\in\mathcal{M}$ for which there is no homography $\mathrm{H}$ in $\mathcal{H}^\star$ having them as inlier, i.e.
\begin{equation}
	\mathcal{M}^\star=\left\{m\in\mathcal{M}:\exists \mathrm{H}\in\mathcal{H}^\star\;m\in\mathcal{I}^\mathcal{M}_{\mathrm{H},t_l}\right\}
\end{equation}
The final set of pruned matches $\mathcal{M}^\star$ can be directly employed in the next step of the pipeline when no keypoint refinement is required, accomplishing the MOP filtering task. 

Otherwise, a homography must be associated with each survived match $m\in\mathcal{M}^\star$, which will be used for patch normalization in order to proceeds with NCC template matching. A possible choice is to assign the homography $\mathrm{H}_m^\varepsilon$ which gives the minimum reprojection error, i.e.
\begin{equation}
	\mathrm{H}_m^\varepsilon=\argmin_{\mathrm{H}\in\mathcal{H}^\star}\varepsilon_\mathrm{H}(m)
\end{equation}
However, this choice is likely to select homographies corresponding to narrow planes in the image with small consensus sets, which leads to a low reprojection error but also to a more unstable and prone to noise assignment.

Another choice is to assign to the match $m$ the homography $\mathrm{H}_m^\mathcal{I}$ compatible with $m$ with the wider consensus set, i.e.
\begin{equation}
	\mathrm{H}_m^\mathcal{I}=\argmax_{\mathrm{H}\in\mathcal{H}^\star,\,m\in\mathcal{I}^{\mathcal{M}^\star}_{\mathrm{H},t_l}}\left|\mathcal{I}^{\mathcal{M}^\star}_{\mathrm{H},t_l}\right|
\end{equation}
which points to more stable and robust homographies, but can lead to incorrect patches when the corresponding reprojection error is not among the lowest ones.

The homography $\mathrm{H}_m$ actually chosen for the match association provides an intermediate solution between the previous ones. Specifically, for a match $m$, defining $q_m$ as the median of the number of inliers of the top 5 compatible homographies according to the number of inliers for $t_l$, $\mathrm{H}_m$ is defined as the compatible homography with the minimum reprojection error among those with an inlier number at least equal to $q_m$
\begin{equation}\label{estimated_hom}
	\mathrm{H}_m=\argmin_{\mathrm{H}\in\mathcal{H}^\star,\,\left|\mathcal{I}^{\mathcal{M}^\star}_{\mathrm{H},t_l}\right|\geq q_m}\varepsilon_H(m)    
\end{equation}

Figure~\ref{miho_img} shows an example of the achieved solution directly in combination with MiHo homography estimation described in Sec.~\ref{miho}. MOP or MOP+MiHo cluster representation does not present visually appreciable differences. Keypoint belonging to discarded matches are marked with black diamonds, while the clusters highlighted by other combinations of markers and colors indicate the resulting filtered matches $m\in\mathcal{M}^\star$ with the selected planar homography $\mathrm{H}_m$. Notice that clusters are not designed to highlight scene planes but points that move according to the same homography transformation within the image pair. 

\section{MiHo implementation}\label{miho_appendix}
While a strict analytical formulation of the MiHo problem which should lead to a practical satisfying solution is not easy to derive, the heuristic found in~\cite{average_homography} can be exploited to modify the base RANSAC employed by MOP described in Appendix~\ref{mop_ransac} to account for the required constraints. Specifically, each match $m$ is replaced by two corresponding matches $m_1=(\mathbf{x}_1,\mathbf{m})$ and $m_2=(\mathbf{m},\mathbf{x}_2)$ where $\mathbf{m}$ is the midpoint within the two keypoints
\begin{equation}
	\mathbf{m}=\frac{\mathbf{x}_1+\mathbf{x}_2}{2}    
\end{equation}
Hence the RANSAC input match set $\mathcal{M}^k=\{m\}$ for the MOP $k$-th iteration is split in the two match sets $\mathcal{M}^k_1=\{m_1\}$ and $\mathcal{M}^k_2=\{m_2\}$. Within a RANSAC iteration, a sample $\mathcal{S}$ defined by Eq.~\ref{sample} is likewise split into
\begin{align}\notag
	&\mathcal{S}_1=\left\{\left(\mathbf{s}_{1i},\tfrac{\mathbf{s}_{1i}+\mathbf{s}_{2i}}{2}\right)\right\}\subseteq\mathcal{M}^k_1 \\    
	&\mathcal{S}_2=\left\{\right(\tfrac{\mathbf{s}_{1i}+\mathbf{s}_{2i}}{2},\mathbf{s}_{2i}\left)\right\}\subseteq\mathcal{M}^k_2, 1\leq i\leq 4    
\end{align}
leading to two concurrent homographies $\mathrm{H}_1$ and $\mathrm{H}_2$ to be verified simultaneously with the inlier set
in analogous way to Eq.~\ref{full_inliers}, i.e.
\begin{equation}
	\mathcal{I}^{\mathcal{M}^k}_{(\mathrm{H}_1,\mathrm{H}_2),t}=\left(\mathcal{J}^{\mathcal{M}^k_1}_{\mathrm{H}_1,t}\upharpoonleft\downharpoonright\mathcal{J}^{\mathcal{M}^k_2}_{\mathrm{H}_2,t}\right)\cap\left(\mathcal{Q}^{\mathcal{M}^k_1}_{\mathrm{H}_1}\upharpoonleft\downharpoonright\mathcal{Q}^{\mathcal{M}^k_2}_{\mathrm{H}_2}\right)
\end{equation}
where for two generic match set $\mathcal{M}_1$, $\mathcal{M}_2$, the $\upharpoonleft\downharpoonright$ operator rejoints splitted matches according to
\begin{align}\notag
	\mathcal{M}_1\upharpoonleft\downharpoonright\mathcal{M}_2=\{(\mathbf{x}_1,\mathbf{x}_2):&(\mathbf{x}_1,\mathbf{m})\in\mathcal{M}_1\\ \wedge&(\mathbf{x}_2,\mathbf{m})\in\mathcal{M}_2\}
\end{align}
All the other MOP steps detailed in Appendix~\ref{mop_ransac}, including RANSAC degeneracy checks, threshold adaptation, and final homography assignment, follow straightly in an analogous way.

Overall, the principal difference to the base MOP is that two concurrent RANSAC steps are executed, corresponding to the replacement of the  actual match set $\mathcal{M}$ by the two linked match sets $\mathcal{M}_1$ and $\mathcal{M}_2$. Analogously, the sampling minimal set $\mathcal{S}$ is replaced by $\mathcal{S}_1$ and $\mathcal{S}_2$. This leads to obtain pairs of homography $(\mathrm{H}_{m_1},\mathrm{H}_{m_2})$ instead of the single $\mathrm{H}_m$ homography inside the RANSAC loop, so that a match of $\mathcal{M}$ is to be considered inlier if the corresponding pair of matches are respectively inliers for $\mathcal{M}_1$ and $\mathcal{M}_2$.
	
\section{MiHo rotation handling}\label{miho_rot_fix}
MiHo is invariant to translations as long as the algorithm used to extract the homography provides the same invariance. This can be easily verified since when translation vectors $\mathbf{t}_a$, $\mathbf{t}_b$ are respectively added to all keypoints $\mathbf{x}_1\in I_1$ and $\mathbf{x}_2\in I_2$ to get $\mathbf{x}'_1=\mathbf{x}_1+\mathbf{t}_a$ and $\mathbf{x}'_2=\mathbf{x}_2+\mathbf{t}_b$ the corresponding midpoints are of the form
\begin{equation}
	\mathbf{m}'=\frac{\mathbf{x}'_1+\mathbf{x}'_2}{2}=\frac{\mathbf{x}_1+\mathbf{x}_2}{2}+\frac{\mathbf{t}_a+\mathbf{t}_b}{2}=\mathbf{m}+\mathbf{t}
\end{equation}
where $\mathbf{t}=\frac{\mathbf{t}_a+\mathbf{t}_b}{2}$ is a fixed translation vector added to all the corresponding original midpoints $\mathbf{m}$. Overall, translating keypoints in the respective images have the effect of adding a fixed translation to the corresponding midpoint. Hence MiHo is invariant to translations since the normalization employed by the normalized 8-point algorithm is invariant to similarity transformations~(\cite{multiview}), including translations.

MiHo is not rotation invariant. Figure~\ref{miho_rot_fig} illustrates how midpoints change with rotations. In the ideal MiHo configuration, where the images are upright, the area of the image formed in the midpoint plane is within the areas of the original ones, making a single cone when considering the images as stacked. The area of the midpoint plane is lower than the minimum area between the two images when there is a rotation about $180^\circ$, and could degenerate to the apex of the corresponding double cone. In analogous way, reflections in the worst case degenerate into segments.

Notice that a similar behavior can be observed in VSAC half homography. More in detail, the matrix entries of the square root decomposition $\mathrm{H}_v$ of the original homography $\mathrm{H}_m$ must not be complex. This implies that $\mathrm{H}_m$ must be in case multiplied by -1 if its determinant is negative since otherwise $\det(\mathrm{H}_v)=\det(\mathrm{H}_m)^{\frac{1}{2}}=\sqrt{-1}$. After this adjustment, all eigenvalues must be positive, so that reflections or rotation by $180^\circ$ cannot be handled, as well as any transformation which can be represented by a triangular matrix with one or two negative diagonal elements, including hence the related affine transformations.

A simple strategy to get an almost-optimal configuration with MiHo was experimentally verified to work well in practice under the observation that images acquired by cameras tend to have relative rotations which are multiples of $90^\circ$. The main idea is to choose the global rotation $\alpha$ which maximizes the configuration where midpoint inter-distances are between corresponding keypoint inter-distances and to adjust the images accordingly. This can be computed efficiently and can be extended to take into account also reflections. An analogous approach can be applied to VSAC. 

\begin{figure}
	\centering
	\hfill
	\includegraphics[width=0.47\textwidth]{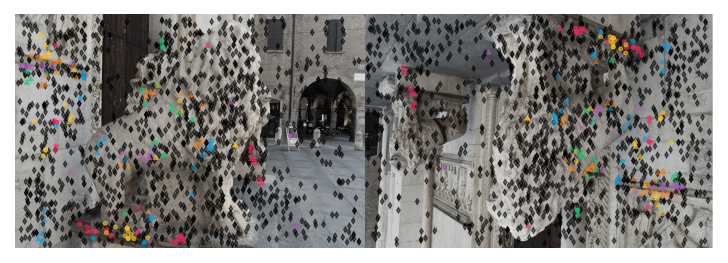}
	\hfill
	\includegraphics[width=0.47\textwidth]{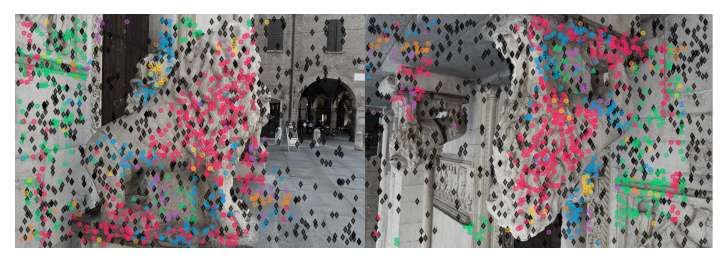}
	\hfill
	\\
	\hfill
	\includegraphics[width=0.47\textwidth]{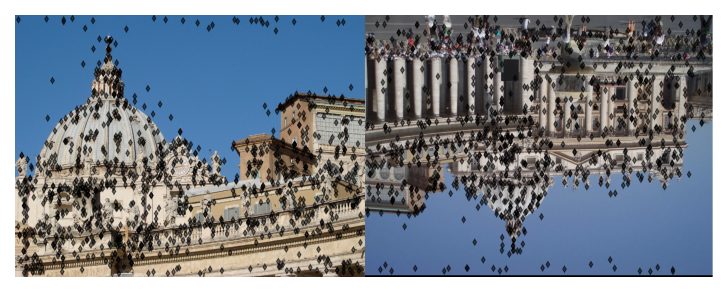}
	\hfill
	\includegraphics[width=0.47\textwidth]{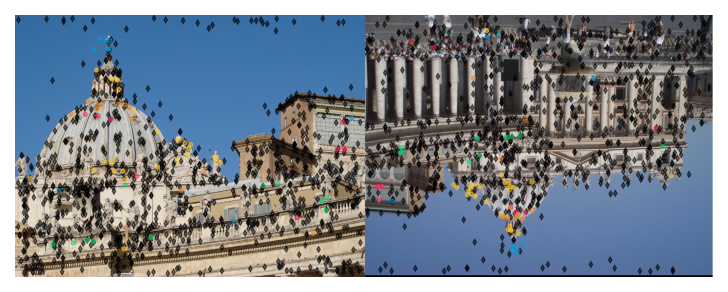}
	\hfill
	\\
	\hfill
	\subcaptionbox{MOP+MiHo without rotation fixing\label{q1}}{\includegraphics[width=0.47\textwidth]{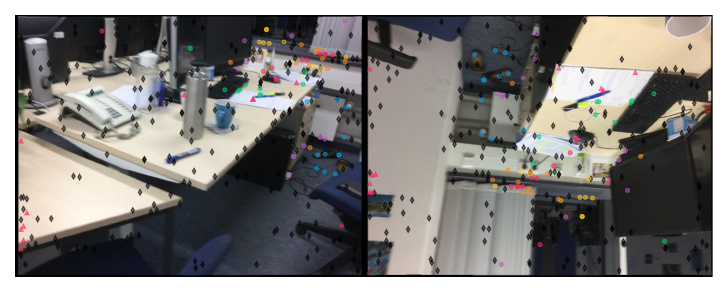}}
	\hfill
	\subcaptionbox{MOP+MiHo with rotation fixing\label{q2}}{\includegraphics[width=0.47\textwidth]{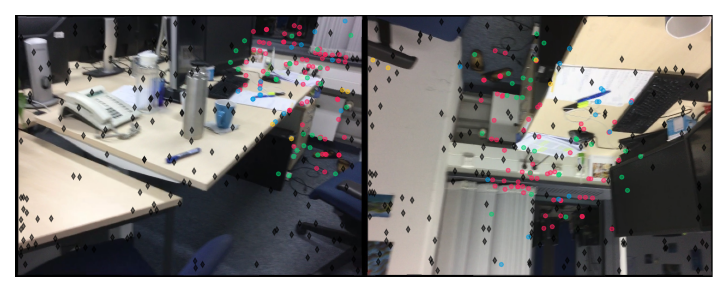}}
	\hfill
	\\
	\caption{MOP+MiHo visual clustering (\subref{q1}) without and (\subref{q2}) with rotation handling in the worst case of a $180^\circ$ relative rotation as described in Sec.~\ref{miho}. The same notation and image pairs of Fig.~\ref{miho_img} are used. Although the results with rotation handling are clearly better, these are not the same of Fig.~\ref{miho_img} since the input matches provided by Key.Net are different as OriNet~(\cite{affnet}) has been further included in the pipeline to compute patch orientations. Notice that the base MOP obtains results similar to (\subref{q2}) without special needs to handle image rotations. Best viewed in color and zoomed in.\label{miho_rot_img}}
\end{figure}

In detail, let us define a given input match $m_i=(\mathbf{x}_{1i},\mathbf{x}^\alpha_{2i})\in\mathcal{M}^\alpha$ and the corresponding midpoint $\mathbf{m}^\alpha_i=\frac{\mathbf{x}_{1i}+\mathbf{x}_{2i}^\alpha}{2}$, where $\mathbf{x}_{1i}\in I_1$ and $\mathbf{x}^\alpha_{2i}\in I^\alpha_2$ being the image $I_2$ rotated by $\alpha$. One has to choose $\alpha^\star\in\mathcal{A}=\{0,\tfrac{\pi}{2},\pi,\tfrac{3\pi}{2}\}$ to maximize the configuration for which midpoint inter-distances are between corresponding keypoint inter-distances 
\begin{equation}
	\alpha^\star=\argmax_{\substack{m_i,m_j\in\mathcal{M}^\alpha\\\alpha\in\mathcal{A}}} 
	\left\llbracket\parallel\mathbf{m}^\alpha_i-\mathbf{m}^\alpha_j\parallel \in \left[d_\downarrow^{ij},d^\uparrow_{ij}\right]\right\rrbracket
\end{equation}
where
\begin{align}\notag
	d_\downarrow^{ij}&=\min\left(\parallel\mathbf{x}_{1i}-\mathbf{x}_{1j}\parallel,\parallel\mathbf{x}^\alpha_{2i}-\mathbf{x}^\alpha_{2j}\parallel\right) \\        
	d^\uparrow_{ij}&=\max\left(\parallel\mathbf{x}_{1i}-\mathbf{x}_{1j}\parallel,\parallel\mathbf{x}^\alpha_{2i}-\mathbf{x}^\alpha_{2j}\parallel\right)     
\end{align}
and $\llbracket\cdot\rrbracket$ is the Iverson bracket. The global orientation estimation $\alpha^\star$ can be computed efficiently on the initial input matches $\mathcal{M}$ before running MOP, adjusting keypoints accordingly. Final homographies can be adjusted in turn by removing the rotation given by $\alpha^\star$. Figure~\ref{miho_rot_img} shows the MOP+MiHo result obtained on a set of matches without or with orientation pre-processing, highlighting the better matches obtained in the latter case. This approach can easily be extended to handle image reflections.

\section{Check for reflections}\label{mop_star}
As discussed in Sec.~\ref{miho}, a further check can be used directly inside the RANSAC loop in Appendix~\ref{mop_ransac} to further constraint the planar search when no image reflections are allowed as for the datasets used in the evaluation. This further variant of MOP is denoted by MOP$^\star$.

Given the minimum hypothesis sampled model $\mathcal{S}$ of Eq.~\ref{sample}, it is possible to define four triangles in the image $I_q$, $q=1,2$ according to the triplets
\begin{equation}
	\mathcal{T}^q_{ijk}=\left(\mathbf{s}_{iq},\mathbf{s}_{jq},\mathbf{s}_{kq}\right)
\end{equation} 
where $1\leq i <j< k\leq4$. There are no reflections if the corresponding triangles $\mathcal{T}^1_{ijk}$ and $\mathcal{T}^2_{ijk}$ considering the ordering provided by the triplet maintains the same clockwise or counterclockwise orientation. Hence it must holds that
\begin{equation}
	\det(\mathcal{T}^{'1}_{ijk})\det(\mathcal{T}^{'2}_{ijk}) > 1\;\;\forall1\leq i<j< k\leq4
\end{equation}
where
\begin{equation}
	\mathcal{T}^{'q}_{ijk}=\begin{bmatrix} \mathbf{s}_{iq} & \mathbf{s}_{jq} & \mathbf{s}_{kq}\\ 1 & 1 & 1\end{bmatrix}
\end{equation}

Notice that although similar in terms of approach, the reflection check is different from the twisted quadrilateral condition often used as homography constraint~(\cite{superransac}), implicitly taken into account in MOP RANSAC implementation by the quasi-affinity check described in Sec.~\ref{mop_ransac}.

Both MOP and MOP$^\star$ uses the same fixed random seed for comparative purposes.

\section{NCC}\label{ncc_appendix}
The keypoint refinement is achieved by obtaining the pixel offset which maximizes the NCC similarity of the corresponding patches among all the warping homography pairs contained in the set $\mathcal{H}_m$, defined in Eq.~\ref{local_image_warping}. Finally, parabolic interpolation is used to get a sub-pixel refinement.
\subsection{Implementation}
Let us denote a generic image by $I$, the intensity values of $I$ for a generic pixel $\mathbf{x}\in\mathbb{R}^2$ as $I(\mathbf{x})$, the window radius extension as
\begin{equation}
	\mathcal{W}_r=\left\{x\in [-r,r]\cap\mathbb{Z}\right\}
\end{equation}
such that the squared window set of offsets is 
\begin{equation}
	\mathcal{W}^2_r=\mathcal{W}_r\times\mathcal{W}_r   
\end{equation}
and has cardinality equal to
\begin{equation}
	\left|\mathcal{W}^2_r\right|=(2w+1)^2    
\end{equation}
The squared patch of $I$ centered in $\mathbf{x}$ with radius $w$ is the pixel set
\begin{equation}
	\mathcal{P}^I_{\mathbf{x},r}=\left\{I(\mathbf{x}+\mathbf{w}): \mathbf{w}\in\mathcal{W}^2_r\right\}
\end{equation}
while the mean intensity value of the patch $\mathcal{P}^I_{\mathbf{x},r}$ is
\begin{equation}
	\mu^I_{\mathbf{x},r}=\frac{1}{\left|\mathcal{W}^2_r\right|}\sum_{\mathbf{w}\in\mathcal{W}^2}I(\mathbf{x}+\mathbf{w}) 
\end{equation}
and the variance is
\begin{equation}
	\sigma^I_{\mathbf{x},r}=\frac{1}{\left|\mathcal{W}^2_r\right|}\sum_{\mathbf{w}\in\mathcal{W}^2}\left(I(\mathbf{x}+\mathbf{w})-\mu^I_{\mathbf{x},r}\right)^2
\end{equation} 
The normalized intensity value $\overline{I}_{\mathbf{x},r}(\mathbf{y})$ by the mean and standard deviation of the patch $\mathbf{P}^I_{\mathbf{x},r}$ for a pixel $\mathbf{y}$ in $I$ is 
\begin{equation}
	\overline{I}_{\mathbf{x},r}(\mathbf{y})=\frac{I(\mathbf{y})-\mu^I_{\mathbf{x},r}}{\sqrt{\sigma^I_{\mathbf{x},r}}}
\end{equation}
which is ideally robust to affine illumination changes~(\cite{gonzales_wood}). Lastly, the similarity between two patches $\mathcal{P}^A_{\mathbf{a},r}$ and $\mathcal{P}^B_{\mathbf{b},r}$ is
\begin{equation}
	S^{A,B}_{\mathbf{a},\mathbf{b},r}=\sum_{\mathbf{w}\in\mathcal{W}^2_r}\overline{A}_{\mathbf{a},r}(\mathbf{a}+\mathbf{w})\overline{B}_{\mathbf{b},r}(\mathbf{b}+\mathbf{w})    
\end{equation}
so that, for the match $m=(\mathbf{x}_1,\mathbf{x}_2)$ and a set of associated homography pairs $\mathcal{H}_m$ derived in Eq.~\ref{local_image_warping}, the refined keypoint offsets $(\mathbf{t}_1^\star,\mathbf{t}_2^\star)$ and the best aligning homography pair $(\mathrm{H}_1^\star,\mathrm{H}_2^\star)$ are given by
\begin{align}\label{best_ncc}\notag
	T^{I_1,I_2}_m=&((\mathbf{t}_1^\star,\mathbf{t}_2^\star),(\mathrm{H}_1^\star,\mathrm{H}_2^\star))\\=&\argmax_{\substack{(\mathbf{t}_1,\mathbf{t}_2)\in\overleftrightarrow{\mathcal{W}}^2_r\\(\mathrm{H}_1,\mathrm{H}_2)\in\mathcal{H}_m}}S^{\mathrm{H}_1\circ I_1,\mathrm{H}_2\circ I_2}_{\mathrm{H}_1\mathbf{x}_1+\mathbf{t_1},\mathrm{H}_2\mathbf{x}_2+\mathbf{t_2},r}
\end{align}
where $\mathrm{H}\circ I$ means the warp of the image $I$ by $\mathrm{H}$, $\mathrm{H}\mathbf{x}$ is the transformed pixel coordinates and
\begin{equation}\label{lr}
	\overleftrightarrow{\mathcal{W}}^2_r=\left(\mathbf{0}\times\mathcal{W}^2_r\right)\cup\left(\mathcal{W}^2_r\times\mathbf{0}\right)
\end{equation}
is the set of offset translation to check considering in turn one of the images as template since this is not a symmetric process. Notice that $\mathbf{t}^\star_1,\mathbf{t}^\star_2$ cannot be both different from $\mathbf{0}$ according to $\overleftrightarrow{\mathcal{W}}^2_r$. The final refined match
\begin{equation}
	m^\star=\left(\mathbf{x}_1^\star,\mathbf{x}_2^\star\right)
\end{equation}
is obtained by reprojecting back to the original images, i.e.
\begin{align}\label{ncc_translation} \notag
	&\mathbf{x}_1^\star={\mathrm{H}^\star_1}^{-1}\left(\mathrm{H}_1\mathbf{x}_1+\mathbf{t}_1^\star\right)\\
	&\mathbf{x}_2^\star={\mathrm{H}^\star_2}^{-1}\left(\mathrm{H}_2\mathbf{x}_2+\mathbf{t}_2^\star\right)
\end{align}
The normalized cross correlation can be computed efficiently by convolution, where bilinear interpolation is employed to warp images. The patch radius $r$ is also the offset search radius and has been experimentally set to $r=10\approx\sqrt{2}t_h$ px. According to preliminary experiments, a wider radius extension can break the planar neighborhood assumption and patch alignment.   

\subsection{Sub-pixel precision}
The refinement offset can be further enhanced by parabolic interpolation adding sub-pixel precision~(\cite{trucco_verri}). Notice that other forms of interpolation have been investigated but they did not provide any sub-pixel accuracy improvements according to previous experiments~(\cite{miho_base}). The NCC response map at $\mathbf{u}\in\mathbb{R}^2$ in the best warped space with the origin coinciding with the peak value can be written as  
\begin{equation}    
	S^\star_m(\mathbf{u})=S^{\mathrm{H}^\star_1\circ I_1,\mathrm{H}^\star_2\circ I_2}_{\mathrm{H}^\star_1\mathbf{x}_1+\mathbf{t}^\star_1+\mathbf{u},\mathrm{H}^\star_2\mathbf{x}_2+\mathbf{t}^\star_2+\mathbf{u},r}
\end{equation}
from $T_m^{I_1,I_2}$ computed by Eq.~\ref{best_ncc}. The sub-pixel refinement offsets in the horizontal direction are computed as the vertex $-\tfrac{b}{2a}$ of a parabola $ax^2+bx+c=y$ fitted on the horizontal neighborhood centered in the peak of the maximized NCC response map, i.e.
\begin{align}\notag
	a&=\frac{S^\star_m\left(\left[\begin{smallmatrix}1\\0\end{smallmatrix}\right]\right)-2S^\star_m\left(\left[\begin{smallmatrix}0\\0\end{smallmatrix}\right]\right)+S^\star_m\left(\left[\begin{smallmatrix}\text{-}1\\\;0\end{smallmatrix}\right]\right)}{2}\\
	b&=\frac{S^\star_m\left(\left[\begin{smallmatrix}1\\0\end{smallmatrix}\right]\right)-S^\star_m\left(\left[\begin{smallmatrix}\text{-}1\\\;0\end{smallmatrix}\right]\right)}{2}	
\end{align}
and analogously for the vertical offsets. Explicitly the sub-pixel refinement offset is $\mathbf{p}=\left[\begin{smallmatrix}p_x\\p_y\end{smallmatrix}\right]$ where
\begin{align}\notag
	p_x=\frac{S^\star_m\left(\left[\begin{smallmatrix}\text{-}1\\\;0\end{smallmatrix}\right]\right)-S^\star_m\left(\left[\begin{smallmatrix}1\\0\end{smallmatrix}\right]\right)}{2(S^\star_m\left(\left[\begin{smallmatrix}1\\0\end{smallmatrix}\right]\right)-2S^\star_m\left(\left[\begin{smallmatrix}0\\0\end{smallmatrix}\right]\right)+S^\star_m\left(\left[\begin{smallmatrix}\text{-}1\\\;0\end{smallmatrix}\right]\right))}\\
	p_y=\frac{S^\star_m\left(\left[\begin{smallmatrix}\;0\\\text{-}1\end{smallmatrix}\right]\right)-S^\star_m\left(\left[\begin{smallmatrix}0\\1\end{smallmatrix}\right]\right)}{2(S^\star_m\left(\left[\begin{smallmatrix}0\\1\end{smallmatrix}\right]\right)-2S^\star_m\left(\left[\begin{smallmatrix}0\\0\end{smallmatrix}\right]\right)+S^\star_m\left(\left[\begin{smallmatrix}\;0\\\text{-}1\end{smallmatrix}\right]\right))}
\end{align}
and from Eq.~\ref{ncc_translation} the final refined match is
\begin{equation}
	m^{\star\star}=(\mathbf{x}_1^{\star\star},\mathbf{x}_2^{\star\star})	
\end{equation}
with
\begin{align}\notag
	&\mathbf{x}_1^{\star\star}={\mathrm{H}^\star_1}^{-1}\left(\mathrm{H}_1\mathbf{x}_1+\mathbf{t}_1^\star+\left\llbracket\,\mathbf{t}_1^\star\ne\mathbf{0}\,\right\rrbracket\,\mathbf{p}\right)\\
	&\mathbf{x}_2^{\star\star}={\mathrm{H}^\star_2}^{-1}\left(\mathrm{H}_2\mathbf{x}_2+\mathbf{t}_2^\star+\left\llbracket\,\mathbf{t}_2^\star\ne\mathbf{0}\,\right\rrbracket\,\mathbf{p}\right)
\end{align}
where the Iverson bracket zeroes the sub-pixel offset increment $\mathbf{p}$ in the reference image according to Eq.~\ref{lr}.

\section{Evaluation details}
\subsection{Base pipelines}\label{pipeline_configs}
As introduced in Sec.~\ref{base_pipeline}, there are seven base pipelines employed for the main comparative analysis carried out in the paper. With the exception of LoFTR, all the evaluated methods are sparse image matching approaches.

SIFT+NNR is included as the standard and reference handcrafted pipeline. The OpenCV~\cite{opencv} implementation was used for SIFT, exploiting RootSIFT~(\cite{rootsift}) for descriptor normalization, while NNR implementation is provided by Kornia~(\cite{kornia}). To stress the match filtering robustness of the successive steps of the pipeline, which is the goal of this evaluation, the NNR ratio threshold was set rather high, i.e. to 0.95. Common adopted values range in $[0.7,0.9]$ depending on the scene complexity, with higher values yielding to less discriminative matches and then possible outliers. Moreover, upright patches are employed by zeroing the dominant orientation for a fair comparison with other recent deep approaches.

Key.Net+$\substack{\text{AffNet}\\\text{HardNet}}$+NNR, a modular pipeline that achieves good results in common benchmarks, is also taken into account. Excluding the NNR stage, it can be considered a modular deep pipeline. The Kornia implementation is used for the evaluation. As for SIFT, the NNR threshold is set very high, i.e. to 0.99, while more common values range in $[0.8,0.95]$. The deep orientation estimation module OriNet~(\cite{affnet}) generally siding AffNet is not included to provide upright matches.

\begin{figure}[h]
	\centering
	\subcaptionbox{\label{ia}}{\includegraphics[height=0.185\textwidth]{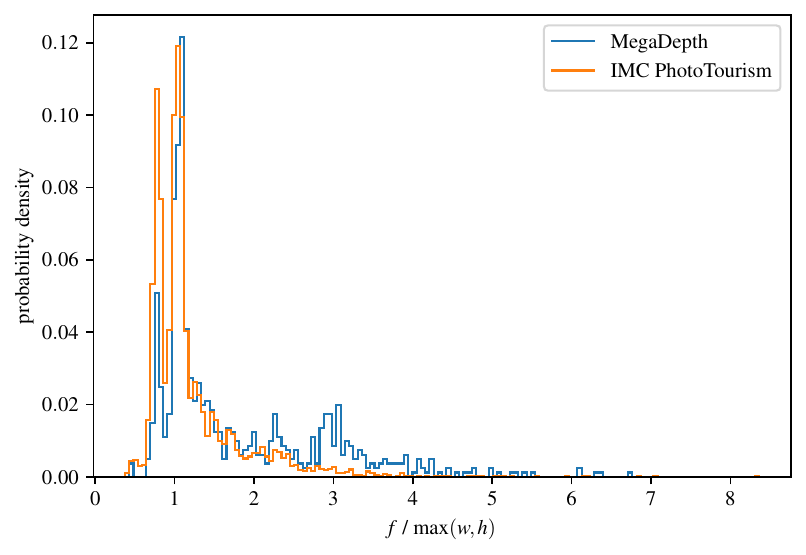}}
	\hfill
	\subcaptionbox{\label{ib}}{\includegraphics[height=0.185\textwidth]{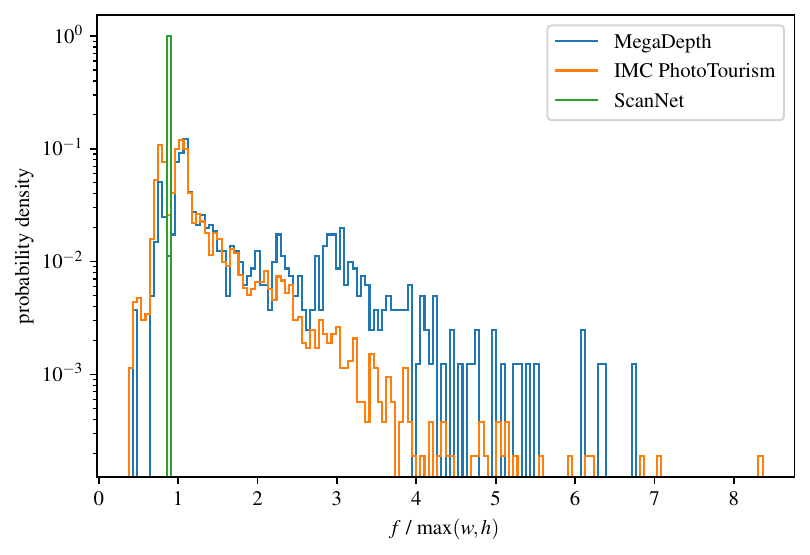}}
	\hfill
	\subcaptionbox{\label{ic}}{\includegraphics[height=0.185\textwidth]{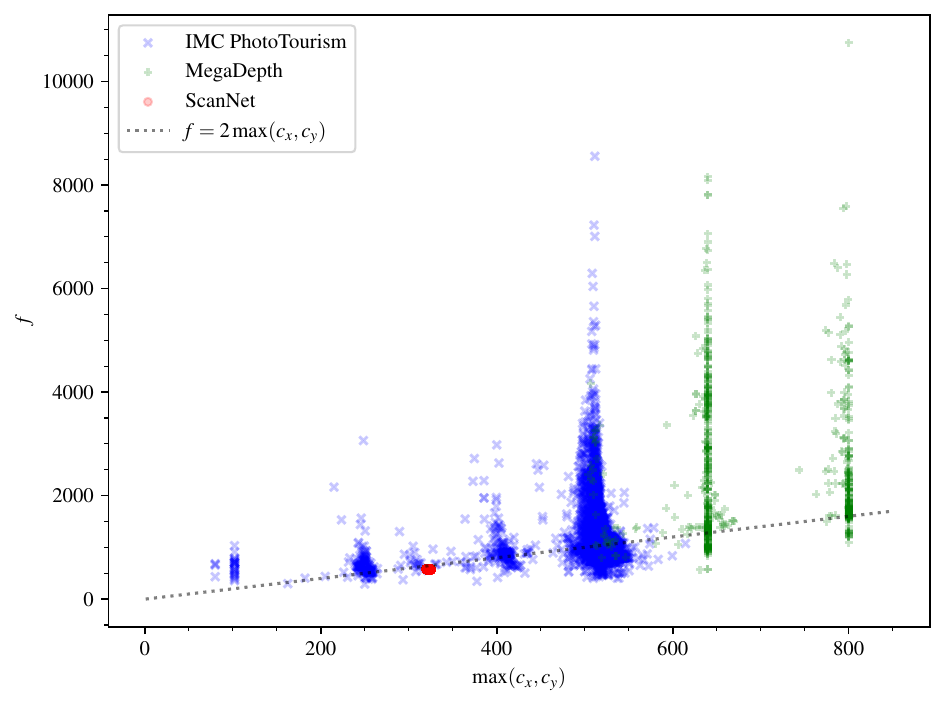}}
	\hfill
	\subcaptionbox{\label{id}}{\includegraphics[height=0.185\textwidth]{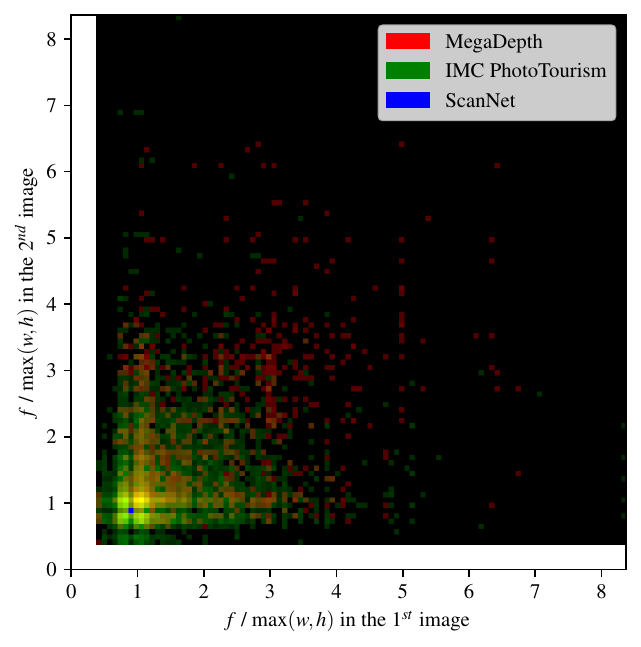}}	
	\caption{(\subref{ia}) Probability distribution of the ratio between the ground-truth focal length $f$ and the maximum image dimension $\max(h,w)$ in the non-planar outdoor MegaDepth and IMC PhotoTourism datasets. (\subref{ib}) The difference between the outdoor data with respect to the indoor ScanNet dataset is shown in a logarithmic scale for a more effective visual comparison. For the indoor scenes, the ratio is fixed to 0.9 as images have been acquired by the same iPad Air2 device~(\cite{scannet}). For the outdoor data, images have been acquired by different cameras and the ratio values cluster roughly around 1. As a heuristic derived by inspecting the images, the ratio is roughly equal to the zooming factor employed for acquiring the image. (\subref{ic}) Analogous representation as a scatter plot in terms of the ground-truth camera center $[c_x\,c_y]^T$ corroborates previous histograms and the intrinsic camera approximation heuristic used in the experiments described in Sec.~\ref{filtering}. (\subref{id}) The 2D distribution of the ratios for the image pairs employed in the evaluation as an overlay in RGB color channels. Notice that the distribution overlap of the outdoor datasets with the corresponding indoor dataset is almost null, as one can check in (\subref{ib}) and by inspecting the cross shape in the bottom left side of (\subref{id}). Best viewed in color and zoomed in.}\label{intrinsic_stats}
\end{figure}

Other base pipelines considered are SOTA full end-to-end matching networks. In details, these are SuperPoint+ LightGlue, DISK+LightGlue and ALIKED+LightGlue implemented by~(\cite{lightglue}). The input images are rescaled so the smaller size is 1024, following LightGlue default. For ALIKED the \texttt{aliked-n16rot} model weights are employed which according to~\cite{aliked} can handle slight image rotation better and are more stable on viewpoint changes.

The last base pipelines added in the evaluation are DeDoDe v2, which provides an alternative end-to-end deep architecture different from the above, and LoFTR, a semi-dense detector-free end-to-end network. These latter pipelines achieve SOTA results in popular benchmarks. The authors' implementation is employed for DeDoDe v2, setting to matching threshold to 0.01, while LoFTR implementation available through Kornia is used in the latter case.

Further experiments on a more selective evaluation configurations have been done using the recent RoMa and MAST3R dense architectures, leading to a total of nine evaluated pipelines. Both RoMa and MASt3R achieve impressive results, especially in complex scenarios. The implementations of their respective authors have been used. For RoMa the coarse and upsample resolutions were set respectively to 560 px and 864 px, while for MASt3R the input images were resized to 512 px. In both case the maximum number of matches was limited to 2048, considering computational resource constraints, and taking into account the low number of returned outliers.

Nowadays benchmarks and deep matching architecture designs assume upright image pairs, where the relative orientation between images is roughly no more than $\frac{\pi}{8}$. This additional constraint allows us to retrieve better initial matches in common user acquisition scenarios. Current general standard benchmark datasets, including those employed in this evaluation, are built so that the image pairs do not violate this constraint. Notice that many SOTA joint and end-to-end deep architectures do not tolerate strong rotations within images by design. 

\subsection{Datasets}\label{dataset_details}
As discussed in Sec.~\ref{base_pipeline}, evaluations have been carried out using four different datasets. The dataset statistics employed to extract the intrinsic camera matrix parameter statistics described in Sec.~\ref{filtering} are shown in Fig.~\ref{intrinsic_stats}.

MegaDepth and ScanNet, respectively outdoor and indoor datasets, are de-facto standard in nowadays image matching benchmarking; sample image pairs for each dataset are shown in Fig.~\ref{miho_img}. For these datasets, the same 1500 image pairs for each dataset and GT data indicated in the protocol employed by~\cite{loftr} are used. MegaDepth test image pairs belong to only 2 different scenes and are resized proportionally so that the maximum size is equal to 1200 px, while ScanNet image pairs belong to 100 different scenes and are resized to $640\times480$ px. Although according to previous evaluation~(\cite{lightglue}) LightGlue provides better results when the images are rescaled so that the maximum dimension is 1600 px, in this evaluation the original 1200 px resize was maintained since more outliers and minor keypoint precision are achieved in the latter case, providing a configuration in which the filtering and refinement of the matches can be better revealed. For deep methods, the best weights that fit the kind of scene, i.e. indoor or outdoor, are used when available.

The IMC-PT dataset is a curated collection of sixteen scenes derived by the  SfM models of the Yahoo Flickr Creative Commons 100 Million (YFCC100M) dataset~(\cite{yfcc100m}). For each scene 800 image pairs have been randomly chosen, resulting in a total of 12800 image pairs. These scenes also provide 3D model scale as GT data so that metric evaluations are possible. Note that MegaDepth data are roughly a subset of IMC-PT. Furthermore, YFCC100M has been often exploited to train deep image matching methods so it can be assumed that some of the compared deep filters are advantaged on this dataset, but this information cannot be retrieved or elaborated. Nevertheless, the proposed modules are handcrafted and not positively biased so the comparison is still valid for the main aim of the evaluation.

\begin{figure}
	\centering
	\includegraphics[width=0.45\textwidth]{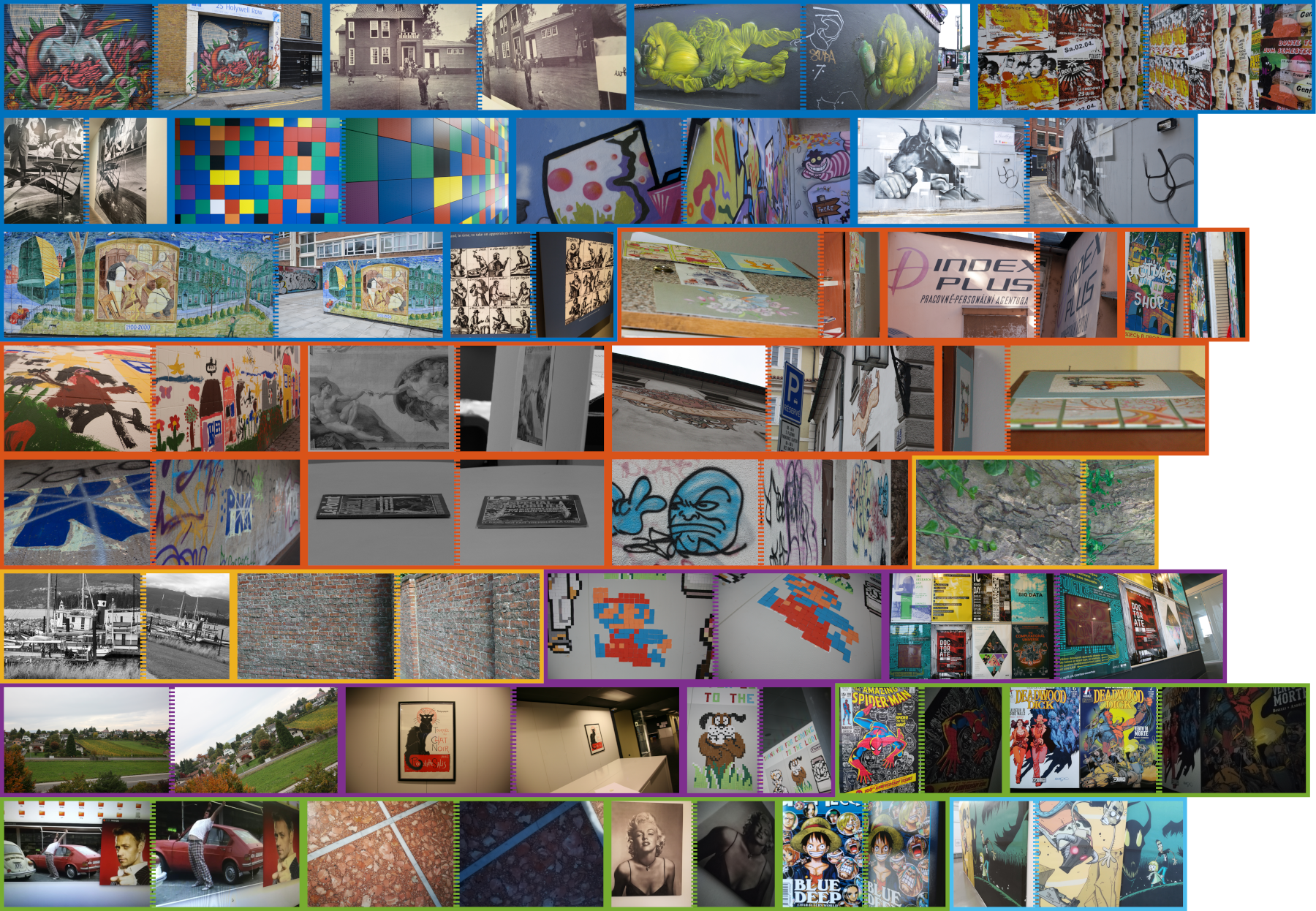}
	\vspace{0.5em}
	\caption{Examples of image pair for each scene in the planar dataset introduced in Sec.~\ref{setup}. Frame colors refer in order to the original dataset of the scene: HPatches~\cite{hpatches}, EVD~\cite{mods}, Oxford~\cite{hess_lapl_affine},~\cite{dtm} and~\cite{slime}. Best viewed in color and zoomed in.}\label{planar_thumb}
\end{figure}

The planar dataset contains 135 image pairs from 35 scenes collected from HPatches~(\cite{hpatches}), EVD~(\cite{mods}) and further datasets aggregated from~(\cite{hess_lapl_affine,dtm,slime}). Each scene usually includes five image pairs except those belonging to EVD consisting of a single image pair. The planar dataset includes scenes with challenging viewpoint changes possibly paired with strong illumination variations. All image pairs are adjusted to be roughly non-upright. Outdoor model weights are preferred for deep modules in the case of planar scenes. A thumbnail gallery for the scenes of the Planar dataset is shown in Fig.~\ref{planar_thumb}.

\subsection{Running times}\label{appendix_time}
Computational time evaluation has been carried out on half of the evaluation image pairs of MegaDepth and ScanNet. The system employed is an Intel Core i9-10900K CPU with 64 GB of RAM equipped with a NVIDIA GeForce RTX 2080 Ti, running Ubuntu 20.04. Average running times are detailed in the additional material\footref{github_results}.

The base matching pipelines run within 0.5-1 s in the testing environment employed. The overhead added by the filtering modules is almost irrelevant for all the compared filters except CC and the proposed ones. NCC is able to refine about 3000 patch pairs within 1 s, while MOP and MOP+MiHo computations depend not only on the size of the input but also on the outlier rate of the input images. In this latter case, running times can be mainly controlled by the number of the maximum iterations for RANSAC $c_{\max}$, defined in Sec.~\ref{mop_ransac} and set to 500, 1000, 1500 and 2000 for the analysis. It is not trivial to derive the running time trend for MOP and MOP+MiHo. In general, MOP goes roughly from 1 s to 2 s for SIFT and Key.Net moving $c_{\max}$ from 500 to 2000, and from 2.5 s to 7s with other pipelines. For MOP+MiHo, setting $c_{\max}$ from 500 to 2000 changes the running times from 1.5 s to 4 s for all pipelines except LoFTR and DeDoDe v2, for which the running time moves from 3 s to 6 s. 

According to the running time data, it turns out that the default value $c_{\max}$=2000 used for the main comparative evaluation in Tables~\ref{sift_res}-\ref{dedode_res} is optimal in terms of matching quality only for SIFT and Key.Net, while it slightly decreases the matching results for the other pipelines with respect to $c_{\max}$=500, which also improves the running times. In detail, moving from $c_{\max}$=500 to $c_{\max}$=2000 causes an increment of around 2\% of the recall and of about 1\% for the AUC pose estimation in the case of SIFT and Key.Net, which contain as initial configuration a higher number of outliers than other pipelines, for which both recall and pose AUC remain instead stable or even worsen. This can be motivated by observing that the next homography discovered in MOP-based approaches is more noisy or even not correct in the long run than the previous ones, and high $c_{\max}$ values force  to retrieve a plane and delay the iteration halting. According to these considerations, it is reasonable that forcing the homography retrieval by increasing $c_{\max}$ starts to degrade the found solutions as the initial outlier ratio decreases.

Notice also that MiHo can take part in the MOP early stopping since it introduces further constraints in the homography to be searched, justifying higher running times for MOP than those of MOP+MiHo in some configurations, although MiHo should ideally need more computational resources as two tied homographies are estimated at each step instead of only one. In particular, MOP+MiHo running time equals that of MOP for LoFTR and DeDoDe v2, is greater for SuperPoint, ALIKED, and DISK, and lower for SIFT and Key.Net. Excluding SIFT and Key.Net, MOP+MiHo finds more homographies than MOP, suggesting the reliability of the organization of input matches within the scene as almost-similarity transformations.

Compared with other filtering and refinement modules, MOP and MOP+MiHo running times are relatively high but still acceptable for off-line, medium-scale image matching tasks. The same holds for the running times of the NCC refinement module. The proposed modules have been implemented in PyTorch exploiting its base parallel support and both GPU workload and memory usage were around 25\% when processing data. Likely, hardware-related and coding optimization could speed up these modules. Notice also that deep-architecture are parallel by design while MOP-based computation inherits RANSAC serialization. Nevertheless, better optimizations to adaptively determine the stopping criteria and to avoid redundant computation dynamically or in a coarse-to-fine manner are planned as future work. 

\subsection{Error metrics}\label{error_metric_details}
As discussed in Sec.~\ref{error_definitions}, it is possible to define a fully metric AUC pose error for IMC-PT dataset and to derive an analogous of the pose error for the planar scenes. 

In the case of IMC-PT dataset, a fully metric estimation of the translation component of the pose is possible since the real GT reconstruction scale is available, i.e. $\left\|\mathbf{\tilde{t}}\right\|$ is the real metric measurement. Inspired by the IMC 2022 benchmark\footref{imc2022}, Eq.~\ref{translation_angle} can be replaced by
\begin{equation}\label{translation_metric}
	\rho^+(\mathbf{\tilde{t}},\mathbf{t})=z\left\|\mathbf{\tilde{t}}-\left\|\mathbf{\tilde{t}}\right\| \frac{\mathbf{t}}{\left\|\mathbf{t}\right\|}\right\|
\end{equation}
so that both the unit vectors are rescaled according to the GT baseline length $\parallel\mathbf{\tilde{t}}\parallel$. The fixed factor $z=10$ is used to translate errors from meters to angles when directly plugging $\rho^+$ into Eq.~\ref{angle_error} instead of $\rho$ in the AUC calculation. This corresponds to replace the angular thresholds $\theta$ of Eq.~\ref{angle_theta} implicitly by pairs of thresholds
\begin{equation}
	(\theta,t_+)\in\left\{(5^\circ,0.5\,\text{m}),(10^\circ,1\,\text{m}),(20^\circ,2\,\text{m})\right\}
\end{equation}  
The translation thresholds $t_+$ are reasonable since the 3D scene extensions are roughly of the order of 100 m. This metric pose accuracy estimation provides a reasonable choice able to discriminate when the real baseline is negligible with respect to the whole scene so that the viewpoint change appears as rotation only. In this case, the translation angular component can be quite high as noise but it does not affect the pose estimation effectively. Using Eq.~\ref{translation_angle} would stress this error without a real justification~\cite{heb}. Likewise above, the corresponding AUC values for the chosen pose estimation between Eq.~\ref{pose_fundamental} and Eq.~\ref{pose_essential} will be denoted by AUC$^F_{@(\theta,t_+)}$, AUC$^E_{@(\theta,t_+)}$ respectively, for the given threshold pair $(\theta,t_+)$. The final metric accuracy scores are given by the average over the $(\theta,t_+)$ value pairs and indicated \textbf{AUC$^F_\square$} and \textbf{AUC$^E_\square$}, respectively.

For the planar dataset, the homography accuracy is measured analogously by AUC. The pose error is replaced by the maximum of the average reprojection error in the common area when using one of the two images as a reference. In detail, assuming that both images have the same resolution of $w\times h$ px., the shared pixel sets for each image are respectively
\begin{align}
	\mathcal{X}_{1\rightarrow2}=&\left\{\mathbf{x}\in\mathbb{Z}^2:\mathrm{\tilde{H}}\mathbf{x} \in [1,w]\times[1,h]\right\}\\
	\mathcal{X}_{2\rightarrow1}=&\left\{\mathbf{x}\in\mathbb{Z}^2:\mathrm{\tilde{H}}^{-1}\mathbf{x} \in [1,w]\times[1,h]\right\}
\end{align}
where $\mathrm{\tilde{H}}$ is the GT homography matrix. The homography estimated by the OpenCV function \texttt{findHomography} from the pipeline matches is denoted as $\mathrm{H}_\mathcal{M}$
and the maximum average reprojection error is given by
\begin{equation}
	\pi(\mathcal{M})=\max\left(\pi_{\mathcal{X}_{1\rightarrow2}}^{\mathrm{H}_\mathcal{M},\mathrm{\tilde{H}}},\pi_{\mathcal{X}_{2\rightarrow1}}^{\mathrm{H}^{-1}_\mathcal{M},
		\mathrm{\tilde{H}}^{-1}}\right)
\end{equation}
where the average reprojection error in a single image is estimated as
\begin{equation}
	\pi_\mathcal{X}^{\mathrm{H}_1,\mathrm{H}_2}=\frac{1}{\left|\mathcal{X}\right|}\sum_{\mathbf{x}\in\mathcal{X}}\left\|\mathrm{H}_1\mathbf{x}-\mathrm{H}_2\mathbf{x}\right\|
\end{equation}
The AUC of the maximum average reprojection error in the common area for each image pair in the dataset and a fixed reprojection threshold
\begin{equation}
	t_\mathrm{H}\in\left\{5\,\text{px},\,10\,\text{px},\,15\,\text{px}\right\}
\end{equation}
is denoted by AUC$^H_{@t_\mathrm{H}}$. The final homography accuracy is provided as the mean AUC over the different thresholds and is indicated by \textbf{AUC$^H_\square$}. The accuracy estimation considering the common area is more reliable than an accuracy estimation by averaging the reprojection error only in the four image corners as done by~\cite{loftr}. This can be understood by observing that the GT planar homography is a pseudo GT estimated from correspondences, which lie inside the boundaries of both images. As an image point gets reprojected in the other image outside and far from the boundaries, which generally happens with the image corners as depicted in Fig.~\ref{miho_figa}, there is no way to check the real error, which also increases going away from the keypoint matches being employed to estimate the GT homography.

\section{More results}\label{more_results}
\subsection{Correlation between error metrics}\label{corr_appendix}
\begin{figure}[h!]
	\centering
	\begin{tabular}{c@{\hphantom{||}}c}
		\subcaptionbox{MegaDepth\label{ca}}{\includegraphics[height=0.11\textwidth]{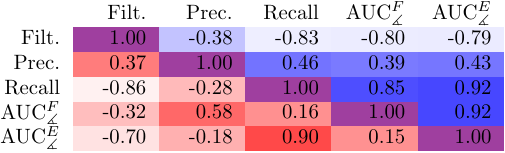}} & 	\subcaptionbox{ScanNet\label{cb}}{\includegraphics[height=0.11\textwidth]{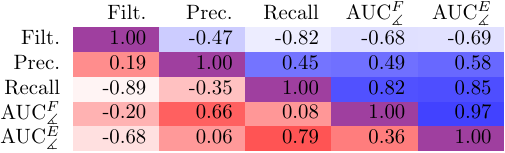}}\vspace{0.75em} \\
		\subcaptionbox{IMC-PT\label{cd}}{\includegraphics[height=0.15\textwidth]{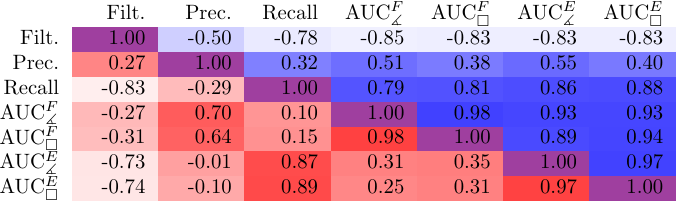}} &
		\subcaptionbox{Planar dataset\label{cc}}{\includegraphics[height=0.09\textwidth]{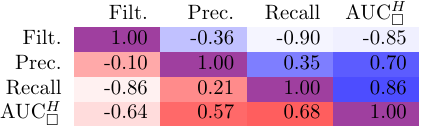}\vspace{1.25em}} \\	
	\end{tabular}
	\caption{Error measure correlation for the different datasets with and without MAGSAC, respectively in the upper (blue) and lower (red) triangular matrix parts, with color intensity highlighting the correlation level. Please refer to Sec.~\ref{prelude}. Best viewed in color.}\label{error_correlation}
\end{figure}

Figure~\ref{error_correlation} reports the correlation within the employed error measures for each dataset, assessing the analysis of Sec.~\ref{prelude}. On one hand, after the MAGSAC geometric constraints, for both planar and non-planar scenes there is a high correlation within recall, pose estimation accuracy as AUC, regardless of the estimation is derived from the fundamental or essential matrix, respectively indicated by the `$F$' and `$E$' superscripts, and these both are highly anti-correlated with the percentage of filtered matches. On the other hand, without MAGSAC in the non-planar sceness the recall and the pose accuracy estimated by the essential matrix are highly correlated and both strongly anti-correlated with the number of filtered matches, while precision has an almost moderate correlation with the pose accuracy estimated from the fundamental matrix only. Metric and angular AUC for the same kind of pose estimation have also high correlation levels. For the planar case, without MAGSAC the pose accuracy is only moderately correlated with both precision and recall, as well as anti-correlated with the percentages of the filtered matches within a similar degree.

\subsection{Number of matches for each method and dataset}\label{num_appendix}
For the sake of completeness, Table~\ref{avg_matches} reports the average number of input matches for each base method and dataset. This information can be used together with the percentage of filtered matches to get an absolute estimation of the filtered matches. According to the reported values, SuperPoint and ALIKED provide the lowest number of matches followed by SIFT and Key.Net, while DISK, LoFTR and DeDoDe v2 the highest number on matches.

\subsection{Further sparse pipelines}\label{other_sparse_appendix}
Table~\ref{disk_res}-\ref{dedode_res} report the results for DISK and DeDoDe v2, respectively, discussed in Sec.~\ref{other_sparse}.

\subsection{Dense pipelines}\label{dense_appendix}
Table~\ref{roma_res}-\ref{master_res} report additional results on RoMa and MASt3R, respectively, discussed in Sec.~\ref{dense}.

\begin{table}
	\caption{Average number of matches per image pair for the base matching pipelines, the highest value for each dataset is in bold. Please refer to Sec.~\ref{prelude}.}\label{avg_matches}
	\centering
	\includegraphics[width=0.6\textwidth]{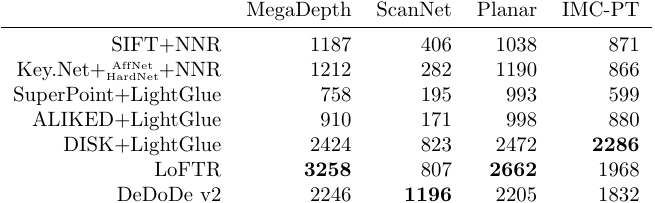}
\end{table}

\begin{table}
	\caption{DISK+LighGlue pipeline evaluation results. All values are percentages, please refer to Sec.~\ref{other_sparse}. Best viewed in color and zoomed in.}\label{disk_res}
	\includegraphics[width=1\textwidth]{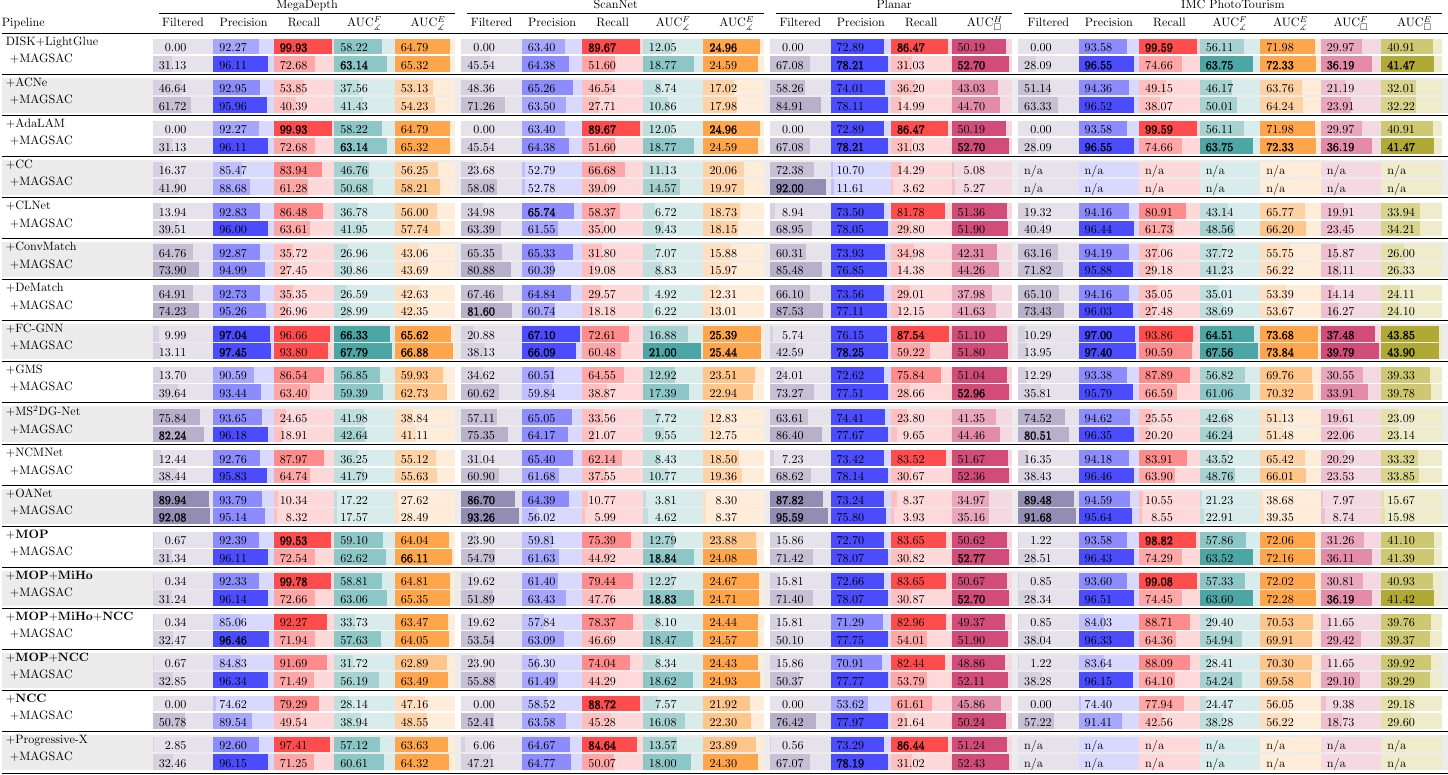}
\end{table}

\begin{table}
	\caption{DeDoDe v2 pipeline evaluation results. All values are percentages, please refer to Sec.~\ref{other_sparse}. Best viewed in color and zoomed in.}\label{dedode_res}
	\includegraphics[width=1\textwidth]{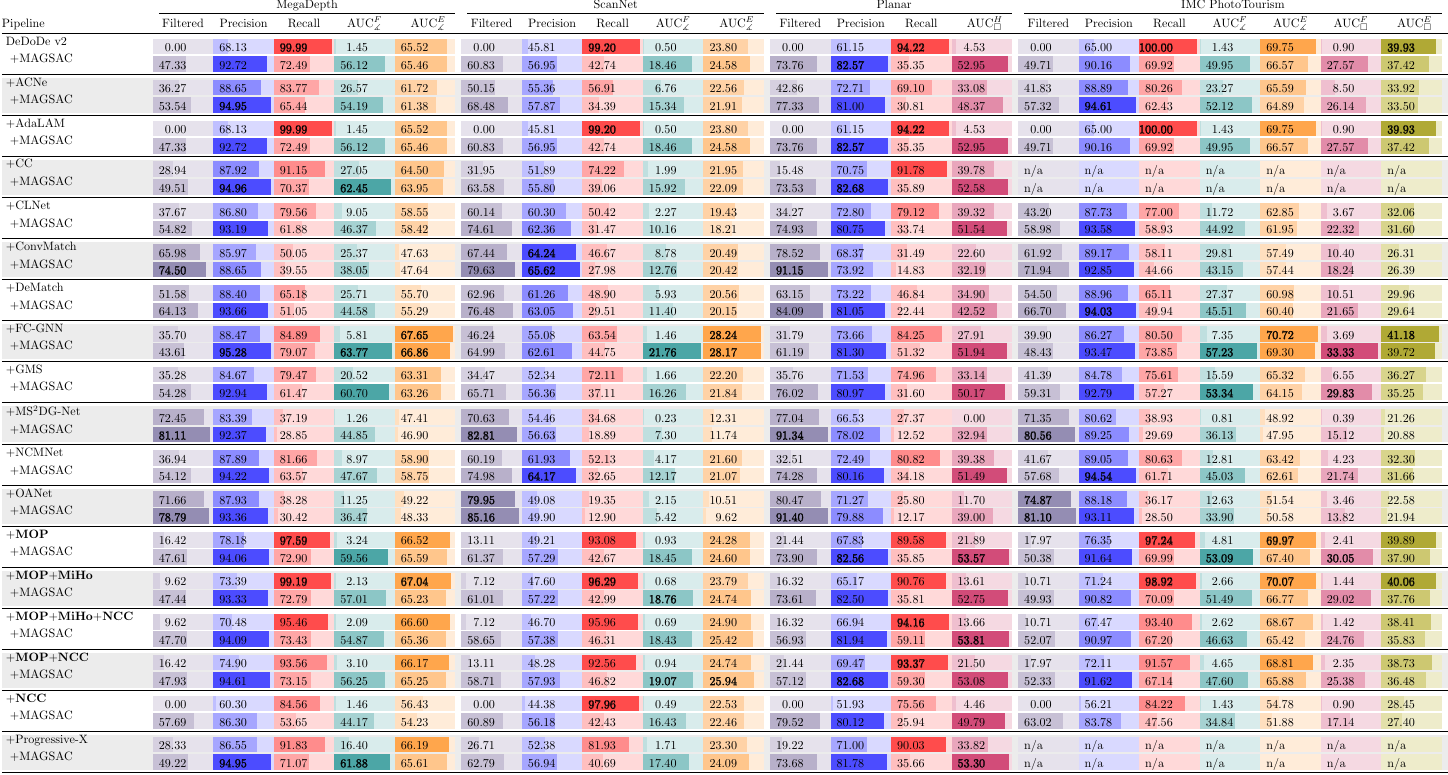}
\end{table}

\begin{table}
	\caption{RoMa pipeline evaluation results. All values are percentages, please refer to Sec.~\ref{dense}. Best viewed in color and zoomed in.}\label{roma_res}
	\centering
	\includegraphics[width=0.7\textwidth]{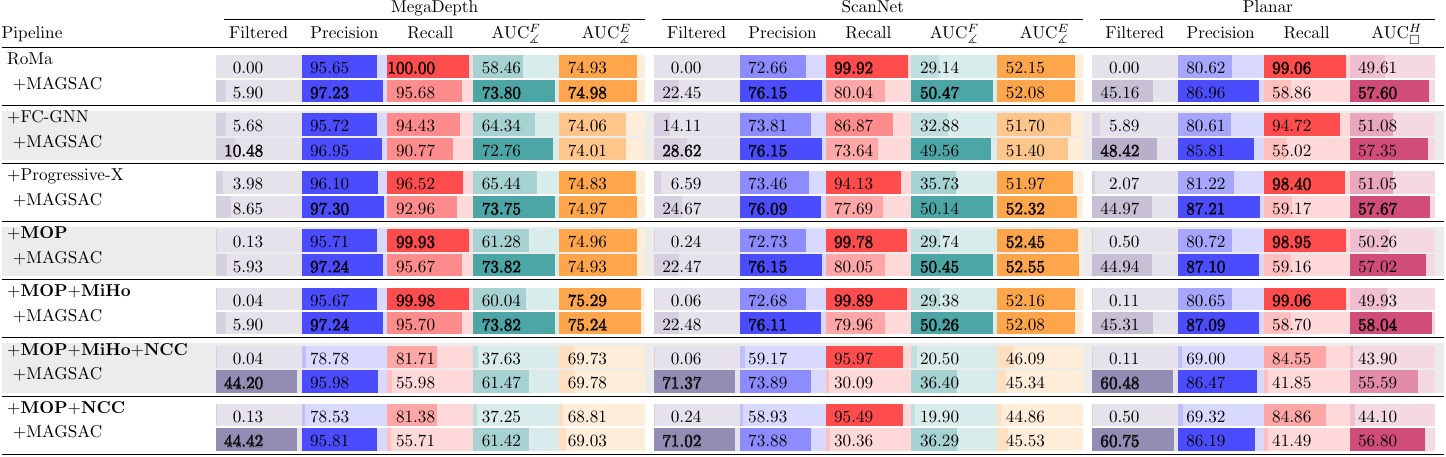}
\end{table}

\begin{table}
	\caption{MASt3R pipeline evaluation results. All values are percentages, please refer to Sec.~\ref{dense}. Best viewed in color and zoomed in.}\label{master_res}
	\centering
	\includegraphics[width=0.7\textwidth]{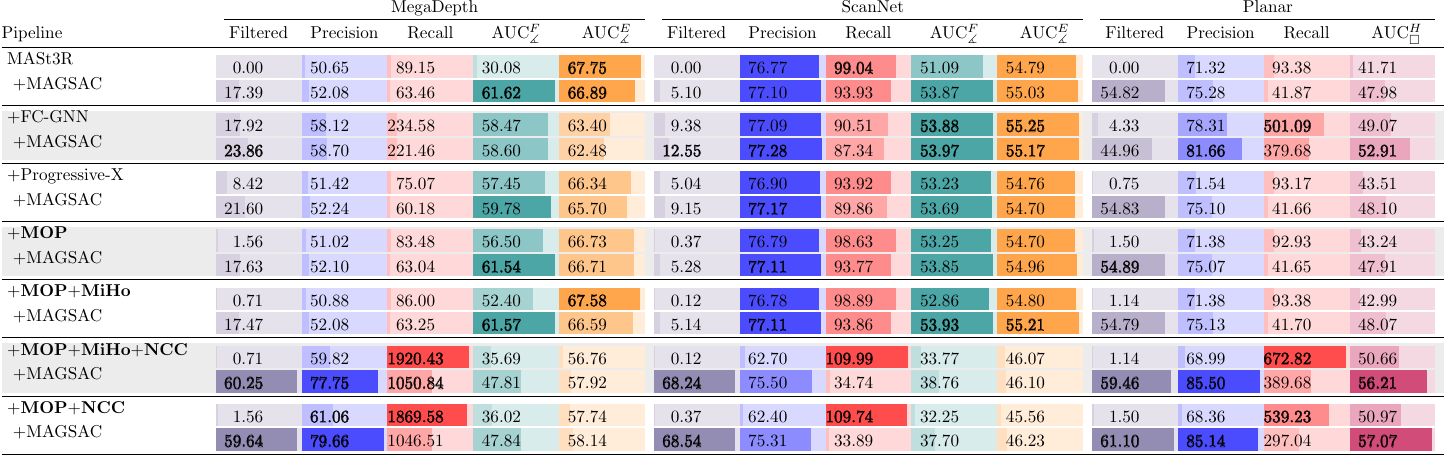}
\end{table}




\bibliographystyle{cas-model2-names}

\bibliography{miho_cviu}



\end{document}